%% file: main.tex
\documentclass[3p,authoryear,a4paper]{elsarticle}

\usepackage{newtxtext,newtxmath} 

\usepackage[colorlinks=true, linkcolor=blue, citecolor=blue, urlcolor=blue]{hyperref}
\usepackage{booktabs}
\usepackage{subfig}
\usepackage{array}
\usepackage{framed}
\usepackage{pgfplots}
\usepgfplotslibrary{fillbetween}
\usetikzlibrary{calc}
\usepackage{multimedia}
\usepackage{multirow}
\usepackage{verbatim}
\usepackage{ragged2e}
\usepackage{url}
\usepackage{booktabs, eqparbox, float, hyperref, nameref}
\usepackage{mathtools}
\usepackage{tikz}
\usepackage{dsfont}
\usepackage{amsmath}

\hypersetup{
  citebordercolor=blue,
  filebordercolor=red,
  linkbordercolor=blue
}
\usepackage{tcolorbox}
\usepackage{pgfplots}
\pgfplotsset{compat=1.7}
\definecolor{bubbles}{rgb}{0.91, 1.0, 1.0}
\usetikzlibrary{decorations.pathreplacing,calligraphy}
\newtheorem{definition}{Definition}

\newtheorem{remark}{Remark}[section]

\usepackage{algorithm}
\usepackage[noend]{algpseudocode}


\usepackage{natbib}

\pgfmathdeclarefunction{gammafcn}{1}{%
  \pgfmathparse{2.506628274631*sqrt(1/#1)+ 0.20888568*(1/#1)^1.5 + 0.00870357*(1/#1)^2.5 - (174.2106599*(1/#1)^3.5)/25920 - (715.6423511*(1/#1)^4.5)/1244160)*exp((-ln(1/#1)-1)*#1}%
}

\pgfmathdeclarefunction{gammadist}{2}{%
  \pgfmathparse{1/(gammafcn(#1)*#2^#1)*x^(#1-1)*exp(-x/#2)}%
}

\pgfmathdeclarefunction{gpd}{2}{%
  \pgfmathparse{(#1 > 0) * (1/#2)*(1+#1*x/#2)^(-1/#1-1)}%
}

\newcommand{\inputnum}{3}
\newcommand{\hiddennumone}{5}
\newcommand{\hiddennumtwo}{4}

\begin{document}

\begin{frontmatter}

\title{Distributional Refinement Network: \\ Distributional Forecasting via Deep Learning}

\author[UM]{Benjamin Avanzi}
\ead{b.avanzi@unimelb.edu.au}

\author[UNSW]{Eric Dong\corref{cor}}
\ead{tian.dong@unsw.edu.au}

\cortext[cor]{Corresponding author. }

\author[UNSW]{Patrick J. Laub}
\ead{p.laub@unsw.edu.au}

\author[UNSW]{Bernard Wong}
\ead{bernard.wong@unsw.edu.au}

\address[UM]{Centre for Actuarial Studies, Department of Economics, University of Melbourne VIC 3010, Australia}

\address[UNSW]{School of Risk and Actuarial Studies, UNSW Business School, UNSW Sydney NSW 2052, Australia}

\begin{abstract}

A key task in actuarial modelling involves modelling the distributional properties of losses. Classic (distributional) regression approaches like Generalized Linear Models (GLMs; \citealp{nelder1972glm}) are commonly used, but challenges remain in developing models that can (i) allow covariates to flexibly impact different aspects of the conditional distribution, (ii) integrate developments in machine learning and AI to maximise the predictive power while considering (i), and, (iii) maintain a level of interpretability in the model to enhance trust in the model and its outputs, which is often compromised in efforts pursuing (i) and (ii).

We tackle this problem by proposing a Distributional Refinement Network (DRN), which combines an inherently interpretable baseline model (such as GLMs) with a flexible neural network--a modified Deep Distribution Regression (DDR; \citealp{li2021deep}) method.
Inspired by the Combined Actuarial Neural Network (CANN; \citealp{schelldorfer2019nesting}), our approach flexibly refines the entire baseline distribution. 
As a result, the DRN captures varying effects of features across all quantiles, improving predictive performance while maintaining adequate interpretability.

Using both synthetic and real-world data, we demonstrate the DRN’s superior distributional forecasting capacity. 
The DRN has the potential to be a powerful distributional regression model in actuarial science and beyond.

\end{abstract}


\end{frontmatter}

\section{Introduction}

In regression problems within general insurance, an important objective is to model the distributional properties of losses, such as the mean, variance, and quantiles, given the risk features~\citep{frees2014predictive, embrechts2022challenges}. 
An associated significant challenge, also recognised in statistics and machine learning~\citep{kneib2021DRreview, klein2024distributional}, involves developing models that accurately capture the entire distribution of the response variable based on its features (or covariates, explanatory variables). 
This requires models with substantial distributional flexibility--precisely capturing the varying impacts of features across all quantiles of the response variable's conditional distribution~\citep{li2021deep, Fissler2023}. 
Overcoming this challenge improves forecasting precision concerning key distributional properties, which is vital for informed decision-making within the actuarial context~\citep{frees_2009}.
Deep learning techniques have gained attention for their potential to empower models with greater distributional flexibility~\citep{bishop1994mixture, taylor2000quantile, li2021deep}. 
However, the additional flexibility provided by neural network adaptations often comes at the cost of ``indispensable" interpretability~\citep{rugamer2023semi}; see also~\citet{richman2022mind} and~\citet{embrechts2022challenges}. 

Balancing flexibility and interpretability remains an ongoing challenge. 
Traditional parametric distributional regression models, such as Generalized Linear Models (GLMs; \citealp{nelder1972glm}), offer an interpretable and structured framework but are often limited by their rigid distributional assumptions. 
These models typically consider only Exponential Family Distributions (EFDs), where variance and quantiles are deterministic functions of the mean, influenced by the treatment of dispersion. 
This restricts their ability to capture the diverse impacts of features on key distributional properties, which is crucial in fields like insurance pricing~\citep{delong2021gamma, fung2022mixture} and loss reserving~\citep{al2022stochastic}.
More complex parametric models, such as deep feedforward neural networks with distributional assumptions and Mixture Density Networks (MDNs; \citealp{bishop1994mixture}), improve flexibility but often lack interpretability and are prone to overfitting~\citep{makansi2019overcoming}. 
Semiparametric methods, like quantile regression~\citep{koenker1978regression} and distribution regression~\citep{foresi1995conditional}, enhance distributional flexibility by relaxing distributional assumptions.
However, as argued by~\citet{zeng2007maximum}, they often lack efficient estimators and the rigorous theoretical frameworks found in parametric models.
They can also adopt unrealistic structural assumptions, leading to issues like quantile crossings in quantile regression~\citep{klein2024distributional}. 
Deep learning adaptations of the above methods, such as deep quantile regression~\citep{taylor2000quantile} and Deep Distribution Regression (DDR; \citealp{li2021deep}), further lack the inherent distributional interpretability.
For high-stakes decision-making, inherently interpretable models like GLMs are often preferred over post-hoc interpretability techniques such as Local Interpretable Model-agnostic Explanations (LIME;  \citealp{ribeiro2016should}) and Shapley Additive Explanations (SHAP; \citealp{lundberg2017unified}), which provide explanations only after predictions have been made. 
They often rely on the unrealistic assumption that features are independent of one another, leading to misleading interpretations of model behaviour~\citep{jethani2022fastshap, micheal2023SHAP}.
As a result, integrating traditional, inherently interpretable GLM frameworks within neural network structures has been proposed, as seen in models like DeepGLM~\citep{tran2020bayesian}, Combined Actuarial Neural Network (CANN; \citealp{schelldorfer2019nesting}), and LocalGLMnet~\citep{richman2022local}. 
However, these parametric frameworks still face the issue of limited distributional flexibility, focusing primarily on the mean and falling short of the ultimate goal of distributional regression~\citep{kneib2021DRreview, klein2024distributional}.

Therefore, a major gap in distributional regression methods lies in maximising distributional flexibility to accurately predict the entire conditional distribution while preserving sufficient distributional interpretability.
To address this gap, we introduce the Distributional Regression Network (DRN), a deep learning framework that provides exceptional distributional flexibility while maintaining a level of interpretability. 
Motivated by CANN, the DRN integrates an inherently interpretable parametric model, such as a GLM, within a flexible semiparametric deep learning framework inspired by the DDR. 
It can be regarded as a deep-learning-refined version of a GLM or other interpretable parametric distributional regression models.
The GLM component provides initial estimates and explanations for the features' impacts on different quantiles. 
The neural network component then refines these estimates in a regularised manner by utilising the available data to minimise a loss function focused on overall distributional forecasting performance.
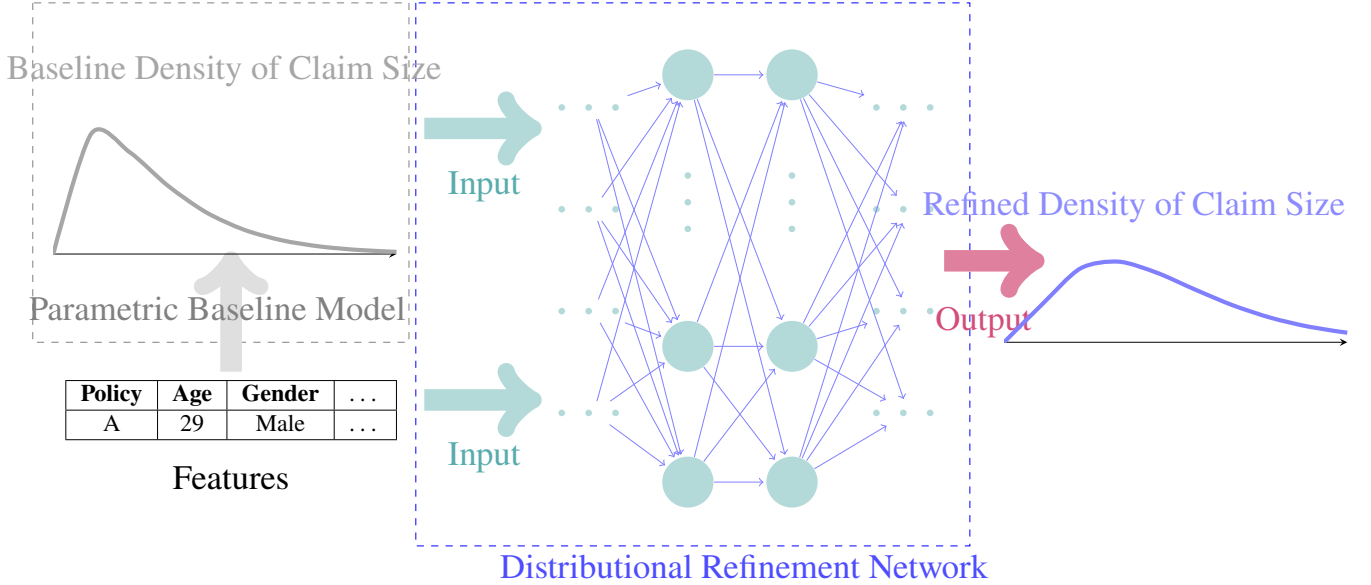
\begin{figure}[h]
    \centering
    \vspace{-1em}
    \scalebox{0.9}{
        \input{images/tikz/DRN_Overview.tex}
    }
    \caption{A schematic view of our proposed deep learning framework designed for distributional forecasting.
    The Distributional Refinement Network (DRN) inputs a feature set and a density estimate provided by a parametric baseline model, such as a GLM. 
    The DRN then outputs a refined version of the initial density estimate.}
    \label{fig: DRN Overview}
\end{figure}
Figure~\ref{fig: DRN Overview} provides a schematic representation of the DRN framework. The key benefits of our approach are as follows:
\begin{enumerate}
    \item The Distributional Refinement Network (DRN) leverages neural networks to flexibly impact various aspects of the response variable's conditional distribution. 
    This approach achieves excellent distributional flexibility through a method similar to DDR, enabling precise forecasting across all distributional properties by relaxing rigid distributional assumptions--a primary advantage over CANN.  
    Greater distributional flexibility allows for better predictive performance, addressing a primary goal in distributional regression within actuarial contexts and beyond.
    \item The DRN maintains a level of distributional interpretability by integrating an inherently interpretable baseline model inspired by the CANN approach. 
    By leveraging the baseline model's forecasting information and using tailored regularisation techniques, the framework combats overfitting and ensures transparency in predictions of key distributional properties--a significant improvement over flexible distributional regression approaches like MDN and DDR.
    In addition, the model’s predictions can be decomposed into contributions from both the transparent baseline model and the deep learning component, providing a practical method for suggesting improvements to the baseline model through analysing the prediction adjustments, if desired.
\end{enumerate}

The remainder of this paper is organised as follows.
Section~\ref{LitReview} reviews existing distributional forecasting models.
Section~\ref{DRN_Methodologies} introduces the Distributional Refinement Network.
Section~\ref{Synthetic Data Analysis} examines the performance of the proposed DRN compared to existing approaches using a synthetic dataset.
Section~\ref{Real Data Analysis} validates the DRN's ability to enhance distributional forecasts with a real-world claim severity dataset.

Section~\ref{Conclusion} concludes.

\section{General Setting and Notation}

Our general setting and associated notation can be summarised as follows:
\begin{itemize}
    \item $\boldsymbol{x}_i = (x_{i,1}, \ldots, x_{i, p})^{\top}\in \mathbb{R}^{p}$ corresponds to the features for the $i$th observation, $y_{i} \in \mathbb{R}$ represents the response variable for the $i$th observation, and $p$ represents the dimension of the features.
    The complete dataset is denoted by $\mathcal{D} = \{\left(\boldsymbol{x}_i,y_i \right)\}_{i=1}^{N}$, with the training, validation and test datasets represented as $\mathcal{D}_{\text{Train}}$, $\mathcal{D}_{\text{Val}}$, and $\mathcal{D}_{\text{Test}}$, respectively.
    \item The random variable $Y|\boldsymbol{X}$ represents the response variable $Y$, conditioned on the explanatory variable $\boldsymbol{X}$.
    The conditional distributions of $Y_i$'s given $\boldsymbol{x}_i$'s are independently distributed.
    The functions $f_{Y|\boldsymbol{X}}(y|\boldsymbol{x}; \boldsymbol{w})$ and $F_{Y|\boldsymbol{X}}(y|\boldsymbol{x}; \boldsymbol{w})$ represent the conditional probability density and cumulative distribution function, respectively, of the response variable $Y$ given the features $\boldsymbol{x}$ and model parameters/weights $\boldsymbol{w}$.
    The symbols $\mathbb{E}_{Y|\boldsymbol{X}}[Y|\boldsymbol{X}]$ and $\mathbb{V}_{Y|\boldsymbol{X}}[Y|\boldsymbol{X}]$ represent the conditional mean and variance, respectively, of $Y|\boldsymbol{X}$.
    Additionally, $Q_{Y|\boldsymbol{X}}(\alpha|\boldsymbol{X})$ signifies the $\alpha$ quantile of $Y|\boldsymbol{X}$.
    \item The quantile function, for a specific instance $\boldsymbol{x}$ and model parameters $\boldsymbol{w}$, computes the $\alpha$ quantile.
In the context of actuarial science and finance, it corresponds to the Value at Risk (VaR) at the $1-\alpha$ level, denoted as $\text{VaR}_{1-\alpha}(Y|\boldsymbol{x}; \boldsymbol{w})$.
We adopt analogous notation for the predicted conditional mean and variance for a given instance.
    \item For the implementation of a specific distributional regression technique referred to as DF, we define the model $\mathcal{M}_{\boldsymbol{w}_{\text{DF}}}$ by its trained parameters or weights represented as $\boldsymbol{w}_{\text{DF}}$.
When employing a GLM, we set $\boldsymbol{w}_{\text{GLM}}=\boldsymbol{\beta}_{\text{GLM}}=\boldsymbol{\beta}$, conforming to the widely accepted notations in existing literature.
Likewise, we can abbreviate the notation $\mathcal{M}_{\boldsymbol{w}_{\text{GLM}}}$ to $\mu_{\boldsymbol{\beta}}$ for simplicity.
Here, $\mu_{\boldsymbol{\beta}}$ represents the mean regression function that is parameterised by $\boldsymbol{\beta}$.
    \item The conditional distribution of $Y|\boldsymbol{X}$ is assumed to follow a specific distribution $\mathcal{F}$, characterised by $K$ parameters, denoted as $\boldsymbol{\theta}(\boldsymbol{X}) = (\theta_{1}(\boldsymbol{X}), \ldots,\theta_K(\boldsymbol{X}))^{\top}$.
When distributional parameters are estimated using a common set of weights, we use $\theta_k(\boldsymbol{X}; \boldsymbol{w}_{\text{DF}})$ to denote the $k$th distributional parameter estimated by the DF approach.
Conversely, when unique weights are assigned to predict each distributional parameter, $\boldsymbol{w}_{\text{DF}, k}$ represents the weights specific to the $k$th distributional parameter.
\end{itemize}

\section{Existing Distributional regression techniques} \label{LitReview}

This section offers a literature review of the current methods for distributional regression.
Sections~\ref{sec: parametric models} and \ref{sec: semi parametric models} elaborate the parametric and semiparametric models, respectively.

\subsection{Parametric Models} \label{sec: parametric models}

Parametric models such as the Generalised Linear Model (GLM), Generalised Additive Models for Location, Scale, and Shape (GAMLSS), Combined Actuarial Neural Network (CANN), and Mixture Density Network (MDN) embed specific distributional assumptions within their frameworks.
Traditional actuarial modelling often relies on the former two models, while the latter two have been extensively explored for their predictive capabilities.
This section provides a concise overview of the methodologies of these models.

\subsubsection{Generalised Linear Model} \label{GLM}

The GLM~\citep{nelder1972glm} stands as one of the most classical distributional regression models.
A GLM employs a regression function, $\mu_{\boldsymbol{\beta}}: \mathbb{R}^{p}\to \mathbb{R}$, to model the mean of the response variable, assuming it follows an exponential family distribution:
    \begin{align} \label{eq: GLM}
        \mu_{\boldsymbol{\beta}}(\boldsymbol{x}) = g^{-1}
    \bigg(\beta_0+ \sum_{j=1}^{p} \beta_j x_j \bigg).
    \end{align}
Here, $\boldsymbol{\beta} = (\beta_0, \beta_1, \ldots, \beta_p)^{\top} \in \mathbb{R}^{p+1}$ denotes the regression coefficient vector and $g^{-1}(\cdot)$ is the inverse link function, which maps the linear combination of the features $\boldsymbol{x}$ to the mean of the response variable.

\subsubsection{Generalised Additive Models for Location, Scale, and Shape} \label{GAMLSS}

GAMLSS~\citep{rigby2005generalized} is a general distributional regression framework.
It enhances distributional flexibility beyond traditional GLMs by supporting more flexible distributional assumptions that extend well beyond the standard exponential family distributions.
GAMLSS assumes that given a feature set $\boldsymbol{X}=\boldsymbol{x}$, the response variable $Y$ follows a distribution $\mathcal{F}$ characterised by parameters $(\theta_1(\boldsymbol{x}), \ldots, \theta_K(\boldsymbol{x}))$. 
Here, GAMLSS constructs $K$ separate regression functions, each modelling a different parameter of the underlying distribution.
The $k$th regression function $ \mathcal{M}_{\boldsymbol{w}_{\text{GAMLSS}, k}}: \mathbb{R}^{p}\to \mathbb{R}$ models the $k$th distributional parameter:
    \begin{align} \label{eq: GAMLSS}
       \mathcal{M}_{\boldsymbol{w}_{\text{GAMLSS}, k}}(\boldsymbol{x})= \theta_{k}(\boldsymbol{x}; \boldsymbol{\beta}_{k},\boldsymbol{\gamma}_k)  = g_k^{-1}\bigg(\beta_{k,0} +  \sum_{j=1}^{p} \beta_{k,j} x_j
                    + \sum_{j=1}^{J_k} h_{k,j}(\boldsymbol{x}; \boldsymbol{\gamma}_{k,j})\bigg),
    \end{align}
where $g_k^{-1}$ represents the $k$th inverse link function, $\boldsymbol{\beta}_k=(\beta_{k,0}, \beta_{k,1}, \ldots, \beta_{k,p})^{\top}\in \mathbb{R}^{p+1}$ denotes the linear vector of coefficients for the $k$th parameter, and $\boldsymbol{\gamma}_{k,j}$ represents the regression coefficient vector for the $j$th smooth non-parametric function predicting the $k$th parameter, i.e., $h_{k,j}$.

\subsubsection{Distributional Neural Network}

A simple feedforward neural network $\mathcal{M}_{\boldsymbol{w}}:\mathbb{R}^{p}\to \mathbb{R}^{K}$ follows
\begin{align} \label{FeedforwardNN}
   \mathcal{M}_{\boldsymbol{w}}(\boldsymbol{x})
   = a^{(L, L+1)}\circ \cdots
\circ a^{(1, 2)}  \circ a^{(0, 1)} (\boldsymbol{x}),
\end{align}
where $L$ represents the number of hidden layers, and $K$ is the number of neurons in the output layer.
For $l\in \{0, \ldots, L\}$, we consider the layer mapping $a^{(l, l+1)}: \mathbb{R}^{q^{(l)}} \to \mathbb{R}^{q^{(l+1)}}$ such that
\begin{align}
    a^{(l, l+1)}(\boldsymbol{z}^{(l)}) = \phi^{(l+1)} \Big( \big(b^{(l+1)}_{1} + \langle \boldsymbol{w}^{(l, l+1)}_1, \boldsymbol{z}^{(l)} \rangle,\ldots, b^{(l+1)}_{q^{(l+1)}} + \langle \boldsymbol{w}^{(l, l+1)}_{q^{(l+1)}}, \boldsymbol{z}^{(l)} \rangle \big)^{\top} \Big),
\end{align}
where $q^{(l)}$ is the number of neurons and $b^{(l)}_k$ is the bias term for the $k$th neuron in the $l$th layer for $k\in\{1, \ldots, q^{(l)}\}$.
The activation function for the $(l+1)$th layer is denoted as $\phi^{(l+1)}: \mathbb{R}^{q^{(l+1)}}\to  \mathbb{R}^{q^{(l+1)}}$.
$\boldsymbol{z}^{(l)} \in \mathbb{R}^{q^{(l)}}$ denotes the neuron values for the $l$th layer, where $\boldsymbol{z}^{(0)}=\boldsymbol{x}$ and $\boldsymbol{z}^{(L+1)}=  \mathcal{M}_{\boldsymbol{w}}(\boldsymbol{x})$.
Further, we define the weights connecting the neurons in the $l$th layer to the $k$th neuron in the $l+1$th layer:
\begin{align}
    \boldsymbol{w}^{(l, l+1)}_{k}=\left({w}^{(l, l+1)}_{1, k},\ldots,{w}^{(l, l+1)}_{q^{(l)}, k}\right)^{\top}\in \mathbb{R}^{q^{(l)}}
\end{align}
for all $k\in \{1, \ldots, q^{(l+1)}\}$.

For optimisation tasks, statisticians usually need to maximise a unique objective function tailored to the specific problem.
For neural networks, the common approach is to minimise a loss function with respect to the weights.
Hence, the goal of training is to find the neural network's weights $\boldsymbol{w}^{*}\in\boldsymbol{\mathcal{W}} $ such that
    \begin{align}
        \boldsymbol{w}^{*} = \underset{\boldsymbol{w}\in\boldsymbol{\mathcal{W}}}{\text{arg min}} \ \mathbb{E}_{(\boldsymbol{X}, Y)}[\mathcal{L}_{\text{FNN}}(\boldsymbol{w}, (\boldsymbol{X}, Y))]
     \approx
        \underset{\boldsymbol{w}\in\boldsymbol{\mathcal{W}}}{\text{arg min}} \ \frac{1}{|\mathcal{D}_{\text{Train}}|}\sum_{(\boldsymbol{x}_i, y_i)\in\mathcal{D}_{\text{Train}}}\mathcal{L}_{\text{FNN}}(\boldsymbol{w}, (\boldsymbol{x}_i, y_i))
    \end{align}
where $\mathcal{L}_{\text{FNN}}(\cdot, \cdot)$ is a user-defined loss function that would typically be chosen with consideration of the objectives of the modeller.
The model parameters are updated iteratively through mini-batch gradient descent, a process distinct from but complementary to the backpropagation algorithm detailed in \citet{backprop1998lecun}.
However, in practice, we usually minimise the loss in the validation dataset to avoid overfitting.
Early stopping will allow training the neural network for as long as the validation loss is still improving after a pre-specified number of epochs (threshold).

Researchers have also explored the combination of GLMs with simple feedforward neural networks, as highlighted by~\citet{tran2020bayesian}, to leverage both the favourable statistical properties of GLMs and the rich flexibility of neural networks.
In actuarial science, 
\citet{schelldorfer2019nesting} proposed the nesting of the GLM in a neural network known as the Combined Actuarial Neural Network (CANN) $\mathcal{M}_{\boldsymbol{w}_{\text{CANN}}}: \mathbb{R}^{p}\to \mathbb{R}$:
\begin{align}\label{eq:CANN}
 \mathcal{M}_{\boldsymbol{w}_{\text{CANN}}}(\boldsymbol{x}) &=g^{-1}\bigg(\alpha(\boldsymbol{x}; \boldsymbol{w}_{\text{CANN}}) \cdot \Big(\beta_0+ \sum_{j=1}^{p} \beta_j x_j \Big)  +(1-\alpha(\boldsymbol{x}; \boldsymbol{w}_{\text{CANN}}))\cdot  a^{(L, L+1)}( \boldsymbol{z}^{(L)})\bigg),
\end{align}
where  $\boldsymbol{\beta} = (\beta_0, \beta_1, \ldots, \beta_p)^{\top} \in \mathbb{R}^{p+1}$ is the vector of regression coefficients obtained through the GLM specified in Section~\ref{eq: GLM}.
The trainable credibility factor $\alpha(\boldsymbol{x}; \boldsymbol{w}_{\text{CANN}})$ balances the contributions between GLM and deep-learning components.
The loss function for training the CANN inherits the deviance loss function adopted for training the baseline GLM.

The MDN~\citep{bishop1994mixture} is another parametric deep-learning approach for distributional forecasting in actuarial science.
The MDN assumes the true underlying distribution to be a mixture of distributions with $K$ components.
The outputs of the MDN consist of parameters of the pre-defined mixture distribution.
Let $K$ denote the number of mixture components and $\boldsymbol{\theta}^{(k)}(\boldsymbol{x}; \boldsymbol{w}_{\text{MDN}})=\big(
\theta_1^{(k)}(\boldsymbol{x}; \boldsymbol{w}_{\text{MDN}})
,\ldots,
\theta_{|\boldsymbol{\theta}|}^{(k)}(\boldsymbol{x}; \boldsymbol{w}_{\text{MDN}})
\big)^{\top} \in \mathbb{R}^{|\boldsymbol{\theta}|}$ denote the distributional parameters of the $k$ th component for all $k\in\{1, \ldots, K\}$.
The MDN $\mathcal{M}_{\boldsymbol{w}_{\text{MDN}}}: \mathbb{R}^{p}\to \mathbb{R}^{K+K\cdot |\boldsymbol{\theta}|}$ outputs:
\begin{align}\label{MDN}
 \mathcal{M}_{\boldsymbol{w}_{\text{MDN}}}(\boldsymbol{x})= (\boldsymbol{\pi}(\boldsymbol{x}; \boldsymbol{w}_{\text{MDN}})^{\top}, \boldsymbol{\theta}^{(1)}(\boldsymbol{x}; \boldsymbol{w}_{\text{MDN}})^{\top}, \ldots,
 \boldsymbol{\theta}^{(K)}(\boldsymbol{x}; \boldsymbol{w}_{\text{MDN}})^{\top}
 )^{\top},
\end{align}
where $\boldsymbol{\pi}(\boldsymbol{x}; \boldsymbol{w}_{\text{MDN}})=(\pi_1(\boldsymbol{x}; \boldsymbol{w}_{\text{MDN}}), \ldots, \pi_K(\boldsymbol{x}; \boldsymbol{w}_{\text{MDN}}))^{\top}\in \mathbb{R}^{K}$ represents the distribution components' mixing weights.
To find the weights $\boldsymbol{w}_{\text{MDN}}$, we minimise the following negative log-likelihood (NLL) function
\begin{align} \label{MDNNLL}
    \mathcal{L}_{\text{MDN}}(\boldsymbol{w}, \mathcal{D}) = -\sum_{(\boldsymbol{x}_i, y_i)\in \mathcal{D}} \ln
     \bigg(
    \sum_{k=1}^{K} \pi_k(\boldsymbol{x}_i; \boldsymbol{w}) \cdot f_{Y|\boldsymbol{X}}(y_i; \boldsymbol{\theta}^{(k)}(\boldsymbol{x}_i; \boldsymbol{w}))
    \bigg)
\end{align}
with respect to the weights.
Mini-batch gradient descent is implemented to minimise the NLL given by Equation~\eqref{MDNNLL}. 
In actuarial science, \citet{al2022stochastic} employs the Gaussian MDN for loss reserving, and \citet{delong2021gamma} examines the effectiveness of a gamma MDN in modelling claim severity.

\subsection{Semiparametric Models} \label{sec: semi parametric models}

Quantile regression and distribution regression stand out as notable examples of semiparametric models.
These models offer enhanced adaptability and flexibility by avoiding imposing assumptions on the distributions of the response variable conditional on its features.
Nonetheless, the deep-learning extensions of these models, which we will discuss in detail in this section, are yet to be explored and effectively implemented for actuarial applications.

\subsubsection{(Deep) Quantile Regression}

Key risk measures can be quantified using quantile regression, which is a semi-parametric approach first introduced by~\citet{koenker1978regression}.
\citet{taylor2000quantile} proposed deep quantile regression, a deep learning variant of quantile regression models to estimate conditional densities.
The deep quantile regression for the $\alpha$ quantile $\mathcal{M}_{\boldsymbol{w}_{\alpha-\text{DQR}}}: \mathbb{R}^{p}\to \mathbb{R}$ follows
\begin{align}\label{DQR}
 \mathcal{M}_{\boldsymbol{w}_{\alpha-\text{DQR}}}(\boldsymbol{x}) = Q_{Y|\boldsymbol{X}}(\alpha| \boldsymbol{x}; \boldsymbol{w}_{\alpha-\text{DQR}}).
\end{align}
The loss function is the pinball loss, defined as follows:
\begin{align} \label{pinball}
    \mathcal{L}_{\alpha-\text{DQR}}(\boldsymbol{w}, \mathcal{D}) = \sum_{(\boldsymbol{x}_i, y_i)\in \mathcal{D}}
    (y_i-Q_{Y|\boldsymbol{X}}(\alpha| \boldsymbol{x}; \boldsymbol{w}))
    \big(\alpha - \mathds{1}_{\{y_i\le Q_{Y|\boldsymbol{X}}(\alpha|\boldsymbol{x} ;\boldsymbol{w})\}}\big).
\end{align}

\subsubsection{(Deep) Distribution Regression}\label{DDR}

An alternative semiparametric approach is adopting distribution regression, a statistical technique that models the conditional distribution of the response variable given features without making distributional assumptions.
\citet{foresi1995conditional} proposed using the generalised additive model to estimate the conditional distributions directly.

Recently, \citet{li2021deep} developed a deep-learning extension named Deep Distribution Regression (DDR).
The idea was to train a neural network to estimate the conditional density without making distributional assumptions.
One of the key steps was partitioning the support of the response variable.
This partitioning step is crucial for several reasons: i) tackling the challenges of specifying the architectures and nature of outputs; ii) the concept of fitting distributional parameters that determine the density levels for the entire range of the response variable is no longer valid due to the distribution-free nature regarding the response variable; iii) enhanced training tractability and efficiency since the conditional density estimation task is converted into a classification problem easier for neural networks to handle; iv) improved practicability with the implementation of loss functions, such as NLL and joint binary cross-entropy (JBCE) defined in Equation~\eqref{JBCE_raw}.
The method of partitioning is as follows:
\begin{enumerate}
    \item Identify an overall interval of focus $(c_{0}, c_{K})$ of the response variable $Y$ for adjustment, where $c_{K}=u\in \mathbb{R}$ is the upper bound and $c_{0}=l\in \mathbb{R}$ is the lower bound.
    \item Partition the interval into $K$ non-overlapping bins by specifying the cutpoints $c_1, c_2,\ldots,c_{K-1}$. The $k$th partitioned interval, denoted as $T_{k}=[c_{k}, c_{k+1})$, has equal width $|T_k|=c_{k+1}-c_{k}$ for all $k\in\{0, 1,\ldots,K-1\}$.
\end{enumerate}

\begin{figure}[H]
    \centering
    \scalebox{0.85}{
        \input{images/tikz/P1_Partition_Method_Default}
    }
    \caption{Demonstration of the DDR model proposed by \citet{li2021deep}.}
    \label{fig:F1}
\end{figure}
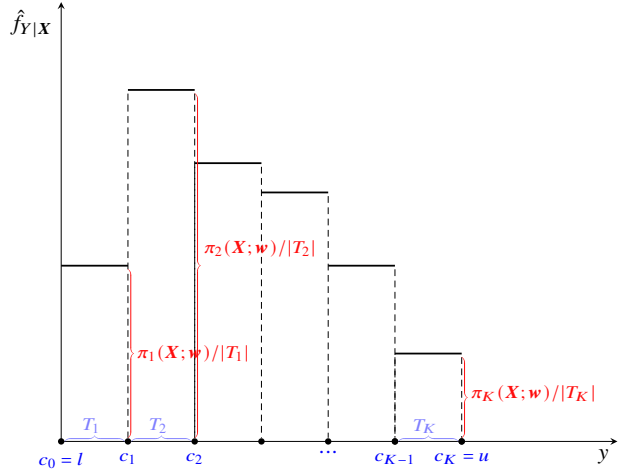

After specifying $K$ partitioned intervals, the DDR network $\mathcal{M}_{\boldsymbol{w}_{\text{DDR}}}: \mathbb{R}^{p} \to \mathbb{R}^{K}$ follows
    \begin{align}
        \mathcal{M}_{\boldsymbol{w}_{\text{DDR}}}(\boldsymbol{x})= \big(\pi_{1}\big(\boldsymbol{x}; \boldsymbol{w}_{\text{DDR}}\big), \ldots, \pi_{K}\big(\boldsymbol{x}; \boldsymbol{w}_{\text{DDR}}\big)\big)^{\top},
    \end{align}
where ${\pi}_{k}(\boldsymbol{x}; \boldsymbol{w}_{\text{DDR}})=\mathbb{P}(Y\in T_k|\boldsymbol{x}; \boldsymbol{w}_{\text{DDR}}) $.
The loss function is given by the joint binary cross-entropy (JBCE) loss function:
\begin{align} \label{JBCE_raw}
  \mathcal{L}_{\text{DDR}}(\boldsymbol{w}, \mathcal{D}) &=   \sum_{k=1}^{K}\sum_{(\boldsymbol{x}_i, y_i)\in \mathcal{D}} \mathds{1}_{\{y_i\le c_{k}\}} \ln F_{Y|\boldsymbol{X}}(c_k|\boldsymbol{x}_i; \boldsymbol{w})
   + \left(1- \mathds{1}_{\{y_i\le c_{k}\}}\right) \ln \left(1-  F_{Y|\boldsymbol{X}}(c_k|\boldsymbol{x}_i; \boldsymbol{w}) \right).
\end{align}
Intuitively, the JBCE loss penalises the discrepancies between the predicted cumulative distribution function (CDF) and actual distribution at various cutpoints, encouraging the model to produce accurate CDF estimates.
The DDR model estimates density as a piecewise constant, as illustrated in Figure~\ref{fig:F1}.

\section{Distributional Refinement Network} \label{DRN_Methodologies}

This section presents the Distributional Refinement Network (DRN) framework, designed for distributional forecasting within actuarial science and applicable to broader contexts.
The DRN is a feedforward neural network that refines an interpretable baseline model's predicted density functions. 
The underlying motivation is that the DRN offers sufficient flexibility to fine-tune baseline densities such that the refined densities better match the true data-generating density functions. 
Such an approach can relax the baseline model's rigid distributional or functional constraints, enhancing distributional flexibility by empowering the neural networks to learn the diverse impact of features on all quantiles.
Integrating an inherently interpretable baseline model ensures distributional interpretability, rendering it applicable to diverse applications.

Inputs to the DRN comprise both features and the transformed density functions predicted by the baseline distributional regression model. 
The related integration procedures and the DRN's architecture are detailed in Section~\ref{sec: DRN Architecture Design}. 
Section~\ref{sec: DRN Distributional Forecasting} explains how the DRN generates and evaluates its distributional forecasts, with details on tailored regularisation for generalisation and robustness considerations.

Without loss of generality, but for notational clarity and convenience, and to agree with a typical actuarial setting, we adopt $\boldsymbol{\beta}$ to denote the baseline model weights $\boldsymbol{w}_{\text{Baseline}}$, and $\boldsymbol{w}$ to denote the DRN weights $\boldsymbol{w}_{\text{DRN}}$.
Accordingly, the conditional density functions estimated by the baseline model and the DRN are represented by $f_{\boldsymbol{Y|\boldsymbol{X}}}(y|\boldsymbol{x}; \boldsymbol{\beta})$ and $f_{\boldsymbol{Y|\boldsymbol{X}}}(y|\boldsymbol{x}; \boldsymbol{w}, \boldsymbol{\beta})$, respectively.
This notation convention extends similarly to the estimated cumulative distribution functions (CDFs) $F_{Y|\boldsymbol{X}}$'s.

\subsection{Architecture Design} \label{sec: DRN Architecture Design}

This section delves into the DRN's comprehensive structural design and construction.
Section~\ref{Incorporating Baseline} covers the necessary steps to integrate a baseline model's distributional forecasting information in the DRN.
The specific architecture of the DRN and its method for generating outputs are elaborated in Section~\ref{sec: Model Architecture and Outputs}. 

\subsubsection{Baseline Model Integration} \label{Incorporating Baseline}

The distinctive characteristic of our proposed Distributional Refinement Network (DRN) is its ability to incorporate the forecasting insights from a baseline distributional regression model into its training process. 
The baseline model greatly impacts the DRN by providing essential ``prior" distributional forecasting information.
This setup enables the users to modulate the contributions between the baseline and the neural network components.
This balance is regulated by a Kullback--Leibler (KL) divergence~\citep{kullback1951information} penalty term, which is further elaborated in Section~\ref{sec: Loss Function and Regularisation Techniques}.
Essentially, the DRN's distributional forecasting performance relies on both the initialisation of the neural network's weights and the parameter estimates derived from the baseline model. 

Therefore, the selection of an appropriate baseline model becomes a nuanced decision pivotal to the DRN's performance. 
An ideal baseline model is expected to yield a reasonable distributional forecast across a given feature set $\boldsymbol{x}$. 
This involves generating both the PDF, $f_{Y|\boldsymbol{X}}(y|\boldsymbol{x};\boldsymbol{\beta})$, and the CDF, $F_{Y|\boldsymbol{X}}(y|\boldsymbol{x};\boldsymbol{\beta})$.
From a philosophical perspective, the baseline model should also be inherently interpretable. 
Regaining interpretability might be more challenging for a complex baseline model, and such a model may not be optimal for a straightforward problem or limited data.
To this end, two natural candidates are GLM and GAMLSS.
For distributional interpretability considerations, especially regarding the mean and occasionally the variance, GLM typically outperforms GAMLSS.
Nonetheless, a GAMLSS baseline might offer improved distributional forecasting performance due to higher flexibility.
Ultimately, the decision rests on the context of the problem and the required level of interpretability for the model. 
Throughout the paper, we select GLM as the baseline model because of its exceptional distributional interpretability and wide acceptance in actuarial applications.

Unfortunately, incorporating a baseline model into a feedforward neural network is challenging.
A feedforward neural network cannot directly process continuous density functions $f_{Y|\boldsymbol{X}}(y|\boldsymbol{x};\boldsymbol{\beta})$'s, which are defined through mathematical relationships predicted by the baseline model.
Transforming the baseline density function into a numerical format the DRN can process is essential.
One solution is to partition the density function into non-overlapping segments, converting information into a format compatible with the DRN.
Specifically, inspired by the DDR methodology, we introduce a ``Partitioned Piecewise Constant" (PPC) transformation that employs the partitioning strategy outlined in Section~\ref{DDR}.
This transformation is designed to modify the baseline density function, $f_{Y|\boldsymbol{X}}(y|\boldsymbol{x};\boldsymbol{\beta})$, into a piecewise constant function across specified intervals determined by a series of predefined cutpoints $c_0<c_1<\ldots<c_{K-1}<c_K$. 
The length of the  $k$th interval is given by $|T_k|=c_k-c_{k-1}$ for $k\in\{1,\ldots,K\}$.
The assumption is that the density function remains constant within each partitioned interval:
\begin{equation} \label{eq: PPC Transformation}
\text{PPC}(f_{Y|\boldsymbol{X}}(y|\boldsymbol{x};\boldsymbol{\beta})) = \begin{cases}
b_k(\boldsymbol{x};\boldsymbol{\beta})/|T_k|, & \text{if } y \in [c_{k-1}, c_k) \text{ for } k=1,\ldots,K, \\
f_{Y|\boldsymbol{X}} (y|\boldsymbol{x}; \boldsymbol{\beta}), & \text{otherwise},
\end{cases}
\end{equation}
where $b_k(\boldsymbol{x};\boldsymbol{\beta})$ calculates the predicted baseline probability mass for the $k$th partitioned interval. 
This is defined by the difference in the CDF values at the cutpoints, i.e., 
\begin{align} \label{eq: baseline prob masses}
    b_k(\boldsymbol{x};\boldsymbol{\beta}) = F_{Y|\boldsymbol{X}}(c_k|\boldsymbol{x};\boldsymbol{\beta})-F_{Y|\boldsymbol{X}}(c_{k-1}|\boldsymbol{x};\boldsymbol{\beta}).
\end{align}
The baseline probability masses $b_k(\boldsymbol{x};\boldsymbol{\beta})$'s are utilised as inputs into the DRN, a process which is thoroughly detailed in Section~\ref{sec: Model Architecture and Outputs}.
Hence, identifying suitable partitioning cutpoints $c_0,...,c_K$, as described in~\eqref{eq: PPC Transformation}, becomes a strategic consideration to transform baseline density for the DRN; see also Section~\ref{DDR}. 
A simple partitioning method involves selecting lower and upper bounds, denoted as $c_0^*$ and $c_K^*$, respectively. These are typically set to values less than the minimum and maximum observations of the response variable in the training dataset. Subsequently, we establish a uniform grid of cutpoints within these bounds.
This naive uniform partitioning method presents a straightforward initial approach; it is easily customised.
An improvement could involve recalibrating cutpoints to ensure that each interval contains at least $M\in \mathbb{Z}^{+}$ observations from the training dataset. 
Details on the algorithm used to merge cutpoints can be found in~\ref{appendix: partitioning algorithm}.

\subsubsection{Output Generation} \label{sec: Model Architecture and Outputs}

\begin{figure}[H]
    \centering
    \scalebox{0.9}{
        \input{images/tikz/P1_DRN_New.tex}
    }
    \caption{
    The schematic illustrates the architecture of the Distributional Refinement Network (DRN). 
    Notably, the baseline model depicted in the bottom left is not directly fed into the network.
    Instead, the DRN employs the summarised baseline information, $\boldsymbol{b}(\boldsymbol{x};\boldsymbol{\beta}) = (\hat{b}_1(\boldsymbol{x}),\ldots, \hat{b}_K(\boldsymbol{x}))^{\top}$, which comprises the predicted baseline probability masses as outlined in Equation~\eqref{eq: baseline prob masses}. 
    Represented by the brown neurons, this information is introduced into the DRN as a skip-connection, combined with the ``raw" outputs, $\boldsymbol{l}(\boldsymbol{x};\boldsymbol{w}) = (\hat{l}_1(\boldsymbol{x}),\ldots, \hat{l}_K(\boldsymbol{x}))^{\top}$. 
    These ``raw" outputs are obtained from the feature inputs $\boldsymbol{x} = (x_1,\ldots,x_{p})^{\top}$ (top left) after propagation through the hidden layers, depicted by the blue neurons.
    The red arrows in the figure signify the process of computing adjustment factors $\boldsymbol{a}(\boldsymbol{x};\boldsymbol{w},\boldsymbol{\beta}) = (\hat{a}_1(\boldsymbol{x}),\ldots,\hat{a}_K(\boldsymbol{x}))^{\top}$.
    They reflect the specific transformation and integration steps within the DRN, defined through Equation~\eqref{eq: adjustment factor}.
    This process remains unchanged during the backpropagation of weights.
    The trainable weights within the DRN are denoted by blue lines, while the grey arrows denote a series of transformations elaborated in Section~\ref{Incorporating Baseline}.
    The network employs \texttt{LeakyReLU} for hidden layers to prevent inactive neuron issues and \texttt{Linear} for the output layer.
    The detailed explanations of these choices are provided in~\ref{appendix: activation function choices}.
    }
    \label{fig: DRN}
\end{figure}
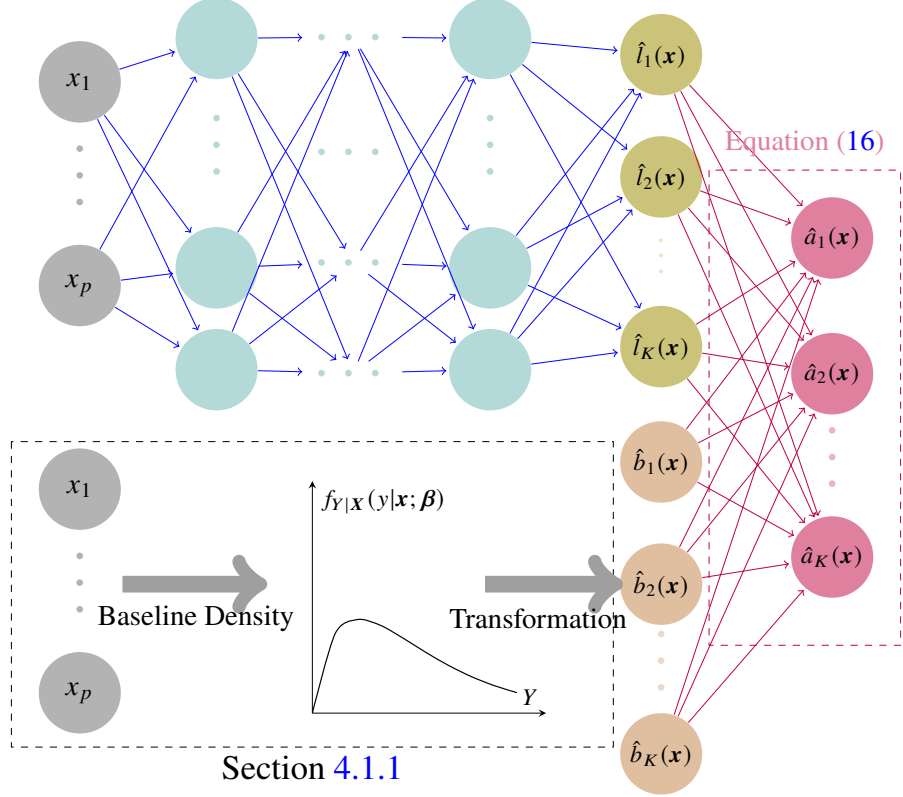

Figure~\ref{fig: DRN} illustrates the architecture of the DRN, structured as a feedforward neural network, with blue arrows representing the trainable weights. 
Inputs to the neural network include the feature set $\boldsymbol{x}$ along with the baseline model's probability masses $b_k(\boldsymbol{x}; \boldsymbol{\beta})$'s as defined in Equation~\eqref{eq: baseline prob masses}. 
These probability masses are depicted by neurons coloured brown. 
The DRN's ``raw" outputs $(l_{1}(\boldsymbol{x}; \boldsymbol{w}), \ldots, l_{K}( \boldsymbol{x};\boldsymbol{w}))^{\top}\in \mathbb{R}^{K}$, are represented by olive-coloured neurons. 
Each of these outputs, $l_{k}(\boldsymbol{x}; \boldsymbol{w})$, is a real-valued function crucial for the process of generating adjustment factors. 

The ``raw" outputs $l_{k}(\boldsymbol{x};\boldsymbol{w})$'s are combined with the baseline model's probability masses $b_k(\boldsymbol{x}; \boldsymbol{\beta})$'s to produce adjustment factors $a_k(\boldsymbol{x};\boldsymbol{w}, \boldsymbol{\beta})$'s for the transformed baseline density functions. 
The adjustment factor for each partitioned interval is defined in the formula below, with its procedural depiction in Figure~\ref{fig: DRN} represented by red arrows, illustrating the specific transformation step involved:
\begin{align} \label{eq: adjustment factor}
    \hat{a}_k(\boldsymbol{x})= a_k(\boldsymbol{x};\boldsymbol{w}, \boldsymbol{\beta})=  \big(F_{Y|\boldsymbol{X}}(c_K|\boldsymbol{x}; \boldsymbol{\beta})-F_{Y|\boldsymbol{X}} (c_0|\boldsymbol{x}; \boldsymbol{\beta})\big) \cdot \frac{\exp(l_{k}(\boldsymbol{x}; \boldsymbol{w}))}{\sum_{j=1}^{K} {b}_j(\boldsymbol{x};\boldsymbol{\beta})  \cdot \exp(l_{j}(\boldsymbol{x}; \boldsymbol{w}))}.
\end{align}
This calculation ensures that the product of the adjustment factor and the baseline probability mass equals the refined probability mass within a specific interval, i.e.,
\begin{align} \label{eq: adjusted prob masses}
    \mathbb{P}(Y\in T_k \,|\, \boldsymbol{x}; \boldsymbol{w}, \boldsymbol{\beta})  
    &= a_k(\boldsymbol{x};\boldsymbol{w}, \boldsymbol{\beta})\cdot b_k(\boldsymbol{x};\boldsymbol{\beta})
    \\ \nonumber
    &=
    \big(F_{Y|\boldsymbol{X}}(c_K|\boldsymbol{x}; \boldsymbol{\beta})-F_{Y|\boldsymbol{X}} (c_0|\boldsymbol{x}; \boldsymbol{\beta})\big)\cdot \text{Softmax}\bigl( \{
      \log(b_j(\boldsymbol{x};\boldsymbol{\beta}))
      + l_j(\boldsymbol{x}; \boldsymbol{w})\}_{j=1,\dots,K} \bigr)_k.
\end{align}

\subsection{Distributional Forecasting} \label{sec: DRN Distributional Forecasting}

This section details the methodologies used in the DRN's distributional forecasting and its evaluation.
Section~\ref{sec: Generating Distributions} describes the process by which DRN forecasts PDFs and CDFs. 
Candidates for the loss function and regularisation techniques to optimise the DRN training are elaborated in Section~\ref{sec: Loss Function and Regularisation Techniques}.
Section~\ref{sec: DRN Model Evaluation} offers a discussion of the evaluation metrics used.

\subsubsection{Distribution Generation}  \label{sec: Generating Distributions}

The DRN generates the density and cumulative distribution functions forecasts utilising the density adjustment factors from Section~\ref{sec: Model Architecture and Outputs} and the transformed baseline density function from Equation~\eqref{eq: PPC Transformation}.

\begin{figure}[H]
    \centering
    \includegraphics[width=0.45\textwidth]{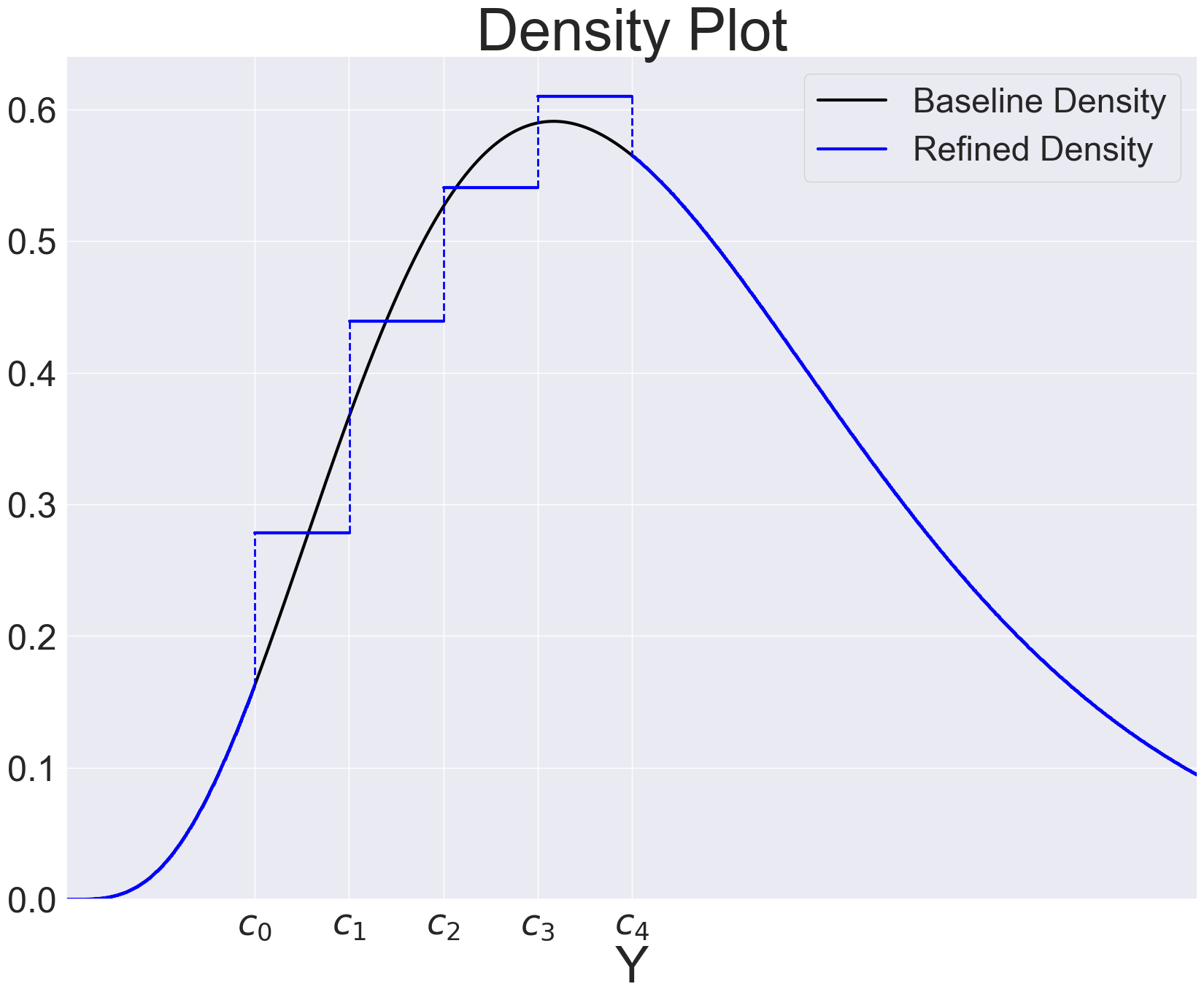}
    \caption{The figure showcases the adjustment factors $\hat{a}_k$'s applied to the density of $Y|\boldsymbol{X}=\boldsymbol{x}$, with dashed vertical lines marking the cutpoints, $c_k$'s.
    For illustration, five cutpoints $c_0<c_1<c_2<c_3<c_4$ form four intervals that require density modifications.
    The black line denotes the baseline model's density, while the blue line indicates the refined density as estimated by the DRN.}
    \label{fig: Density DEMO}
\end{figure}

The DRN's density is assumed to follow a uniform distribution within each partitioned interval.
Subsequently, we introduce the DRN density estimator as follows:
\begin{align} \label{eq: DRN density estimator}
    f_{Y|\boldsymbol{X}}(y|\boldsymbol{x}; \boldsymbol{w}, \boldsymbol{\beta}) &=\begin{cases}
     \sum_{k=1}^{K} \mathds{1}_{\{y\in T_k\}}
      a_k(\boldsymbol{x};\boldsymbol{w}, \boldsymbol{\beta}) \cdot b_k(\boldsymbol{x};\boldsymbol{\beta})/ |T_k|,
    &  y \in [c_0, c_K),\\
     f_{Y|\boldsymbol{X}} (y|\boldsymbol{x}; \boldsymbol{\beta}), & \text{otherwise}.
     \end{cases}
\end{align}
A key feature of the DRN is its ability to maintain practical density estimation beyond the distributional refinement region.
Specifically, for any $y$ outside the range $[c_0, c_K)$, the DRN estimator reverts to the estimate of the baseline model: $f_{Y|\boldsymbol{X}}(y|\boldsymbol{x}; \boldsymbol{w}, \boldsymbol{\beta})=f_{Y|\boldsymbol{X}} (y|\boldsymbol{x}; \boldsymbol{\beta})$.
As illustrated in Figure~\ref{fig: Density DEMO}, the DRN's refined density function mirrors the baseline model's density outside the refining region, highlighting the seamless baseline integration and transition.
Building on the DRN density estimator as defined in Equation~\eqref{eq: DRN density estimator}, we derive the DRN cumulative distribution function (CDF) estimator:
\begin{align} \label{eq: DRN CDF estimator}
      F_{Y|\boldsymbol{X}}(y|\boldsymbol{x}; \boldsymbol{w}, \boldsymbol{\beta})
      &=\begin{cases}
    F_{Y|\boldsymbol{X}} (y|\boldsymbol{x}; \boldsymbol{\beta}), & y\in (-\infty, c_0),\\
     F_{Y|\boldsymbol{X}} (c_0|\boldsymbol{x};\boldsymbol{\beta})+
     \int_{c_0}^{y} f_{Y|\boldsymbol{X}}(t|\boldsymbol{x}; \boldsymbol{w}, \boldsymbol{\beta}) \ \mathrm{d}t, &  y \in [c_0, c_K),\\
      F_{Y|\boldsymbol{X}}(y|\boldsymbol{x};\boldsymbol{\beta}), & y\in [c_K, \infty).
     \end{cases}
\end{align}
Specifically, the model directly adopts the baseline CDF for the response variable's values within $(-\infty, c_0)$ and beyond $[c_K, \infty)$.
For values within the interval $[c_0, c_K)$, it incorporates adjustments based on the DRN's output to refine the CDF estimation, enhancing the model's accuracy and adaptability.

\subsubsection{Loss Function and Regularisation Techniques}  \label{sec: Loss Function and Regularisation Techniques}

Selecting an appropriate loss function is critical for effectively training the DRN.
We consider two potential choices for the loss function.
One viable option is the Negative Log-Likelihood (NLL) loss function:
\begin{align}  \label{NLL}
  \mathcal{L}_{\text{NLL}}(\boldsymbol{w}, \mathcal{D}) &=   -\sum_{(\boldsymbol{x}_i, y_i)\in \mathcal{D}}  \ln\big({f}_{Y|\boldsymbol{X}}(y_i|\boldsymbol{x}_i; \boldsymbol{w}, \boldsymbol{\beta})\big),
\end{align}
An alternative option is the Joint Binary Cross-Entropy (JBCE) loss function:
\begin{align} \label{JBCE}
  \mathcal{L}_{\text{JBCE}}(\boldsymbol{w}, \mathcal{D}) =   \sum_{k=1}^{K}\sum_{(\boldsymbol{x}_i, y_i)\in \mathcal{D}} &\mathds{1}_{\{y_i\le c_{k}\}} \ln {F}_{Y|\boldsymbol{X}}(c_k|\boldsymbol{x}_i; \boldsymbol{w}, \boldsymbol{\beta}) \nonumber \\
  &+\left(1- \mathds{1}_{\{y_i\le c_{k}\}}\right) \ln \left(1-  {F}_{Y|\boldsymbol{X}}(c_k|\boldsymbol{x}_i; \boldsymbol{w}, \boldsymbol{\beta}) \right).
\end{align}

The JBCE loss function emerges as the superior choice for training the DRN due to several compelling reasons.
First, the JBCE loss ensures the model's performance remains stable across various partitioning configurations, meaning the distributional forecasting performance is less affected by the selection of cutpoints.
This loss more effectively mitigates instability in `zero-observation' intervals compared to the NLL loss~\citep{li2021deep}.
Secondly, building on the previous reason, the JBCE loss allows for the allocation of additional cutpoints~\citep{li2021deep}, enhancing the density refinement capability of the neural network due to increased distributional flexibility.
Finally, extending the classic distribution regression model \citep{foresi1995conditional} to predict the entire conditional distribution requires the sequential training of binary classifiers at each cutpoint \citep{kneib2021DRreview}.
This methodology intuitively necessitates using the JBCE loss, prompting the DRN to minimise the cumulative binary classification losses across all partitioned intervals. 

Building on the advantages of the JBCE loss function over NLL for training the DRN, we discuss the practical implementation of regularisation in this framework, driven by two primary motivations and desired properties.
First, we create a controlled distributional continuum with respect to the baseline model. 
Second, we reduce the roughness of the density estimation~\citep{good1971roughness}, enhancing the model's smoothness to ensure reliability and interpretability.
Define $\hat{f}_{\text{Baseline}}$ and $\hat{f}_{\text{DRN}}$ as the baseline and DRN density estimators, respectively, for an input instance $\boldsymbol{x}$.
The regularised loss function is defined as follows:
\begin{eqnarray}
  \mathcal{L}^*_{\text{JBCE}}(\mathcal{D}, \boldsymbol{w}) &=&   \mathcal{L}_{\text{JBCE}}(\mathcal{D}, \boldsymbol{w})
  \nonumber
  \\
  &&+  \alpha_1 \cdot \underbrace{\frac{1}{N}\sum_{i=1}^{N} D_{\text{KL}}\bigl[\hat{f}_{\text{Baseline}}(y|\boldsymbol{x}_i)||\hat{f}_{\text{DRN}}(y|\boldsymbol{x}_i)\bigr]}_
  {\text{KL Penalty: Baseline Distributional Resemblance}}
  \label{KL Divergence Regularisation}
  \\
  &&+ \alpha_2 \cdot \underbrace{\frac{1}{N}\sum_{i=1}^{N}  {\Phi}_{c_0}^{c_K}\bigl(\hat{f}_{\text{DRN}}(y|\boldsymbol{x}_i)\bigr)}_
  {\text{Roughness Penalty: Density Smoothness}}
  \label{Density Variation Regularisation}
  \\
  &&+  \alpha_3 \cdot \underbrace{\frac{1}{N}\sum_{i=1}^{N} \bigl(\mathbb{E}_{Y|\boldsymbol{X}}[Y|\boldsymbol{x}_i; \boldsymbol{w}, \boldsymbol{\beta}] - \mathbb{E}_{Y|\boldsymbol{X}}[Y|\boldsymbol{x}_i; \boldsymbol{\beta}]\bigr)^2}_
  {\text{Mean Penalty: Baseline Mean Resemblance}}
  \label{Mean Regularisation}
\end{eqnarray}
where $\alpha_1, \alpha_2, \alpha_3 \ge 0$ are some penalty coefficients.

\begin{figure}
    \centering
    \includegraphics[width=0.8\textwidth]{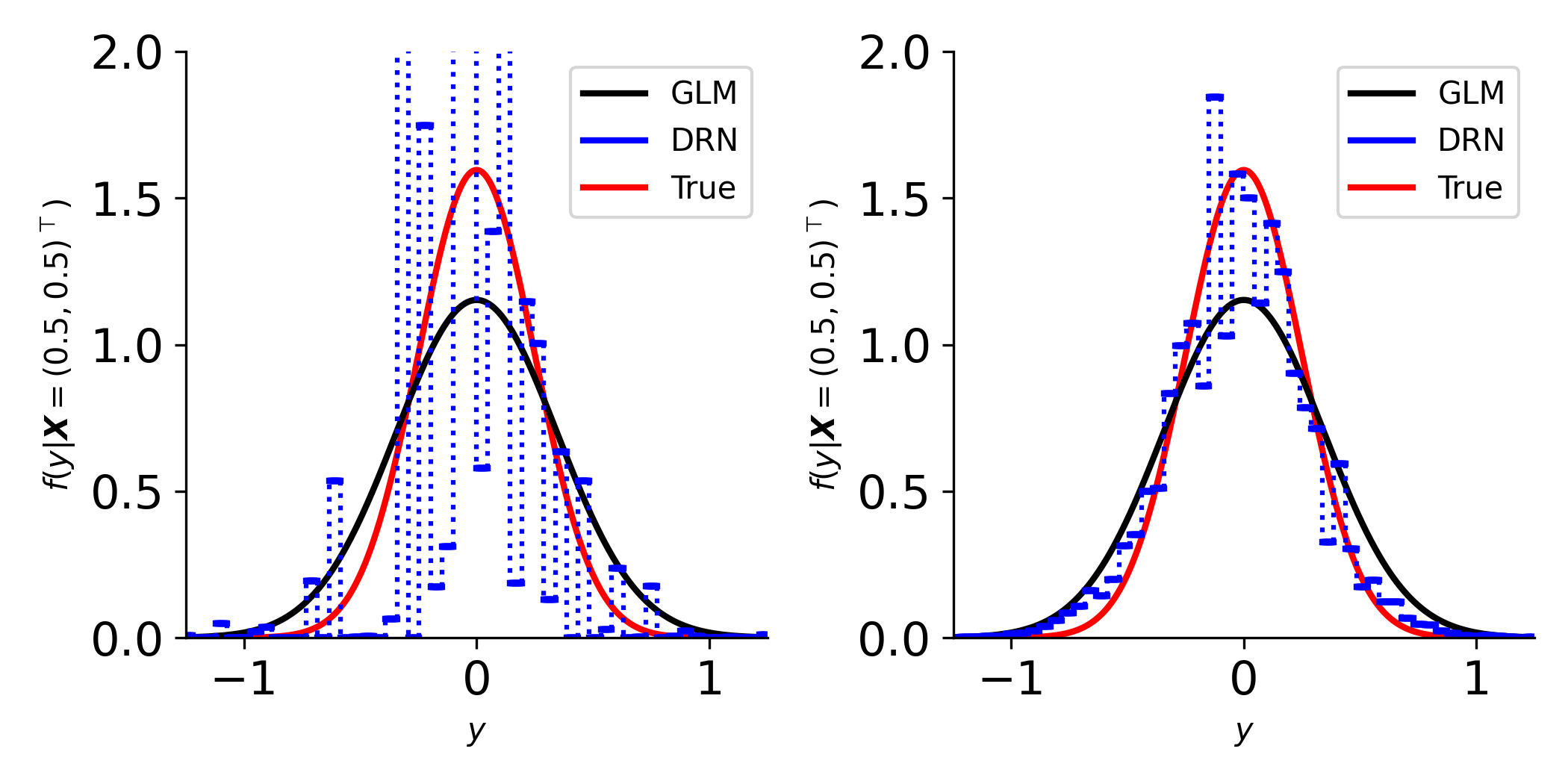}
    \caption{The conditional densities generated by the DRN are shown with and without KL regularisation applied.
    The left graph shows the unregularised DRN density estimate ($\alpha_1 = 0$) and the right graph shows KL regularisation applied ($\alpha_1 = 0.002$).
    The DRN density function (\textcolor{blue}{blue}) is compared against the baseline GLM (\textcolor{black}{black}) density and the true density (\textcolor{red}{red}) for the specific observation $\boldsymbol{X}=(0.5, 0.5)$.}
    \label{fig: KL Penalty}
\end{figure}

To rationalise the inclusion of these regularisation terms, we present both intuitive and theoretical justifications, complemented by empirical evidence from a study involving a simple synthetic dataset.
The procedures for generating the synthetic dataset are detailed in~\ref{appendix: regularisation synthetic dataset}.
The left panel of Figure~\ref{fig: KL Penalty} compares the baseline model's density function, the non-regularised DRN's density, and the true density function for instance $X=(0.5, 0.5)^{\top}$.
This comparison reveals significant fluctuations in the non-regularised DRN density function across ordinates, indicating a risk of overfitting.
Therefore, integrating a regularisation term that encourages the neural network to treat the baseline as ``prior" knowledge can significantly decrease the likelihood of the model capturing noise as part of the learning process.
More precisely, we apply a KL divergence penalty between the adjusted densities and baseline densities from~\eqref{KL Divergence Regularisation}:
\begin{align} \label{eq: KL divergence}
   D_{\text{KL}}\bigl[\hat{f}_{\text{Baseline}}(y|\boldsymbol{x}_i)||\hat{f}_{\text{DRN}}(y|\boldsymbol{x}_i)\bigr]
   &= \int_{c_0}^{c_K} \hat{f}_{\text{Baseline}}(y|\boldsymbol{x}_i) \cdot \log\bigg(\frac{\hat{f}_{\text{Baseline}}(y|\boldsymbol{x}_i)}{\hat{f}_{\text{DRN}}(y|\boldsymbol{x}_i)}\bigg) \ \mathrm{d}y  
   \nonumber \\
   &= - \sum_{k=1}^K \int_{c_{k-1}}^{c_k} \hat{f}_{\text{Baseline}}(y|\boldsymbol{x}_i) \log(\hat{f}_{\text{DRN}}(y|\boldsymbol{x}_i)) \ \mathrm{d}y + \mathcal{O}(1)  
   \nonumber \\
  &= - \sum_{k=1}^K \log(\hat{f}_{\text{DRN}}( c_{k-1}|\boldsymbol{x}_i)) \int_{y=c_{k-1}}^{c_k} \hat{f}_{\text{Baseline}}(y|\boldsymbol{x}_i) \ \mathrm{d}y + \mathcal{O}(1)  
  \nonumber \\
  &= - \sum_{k=1}^K \log(\hat{f}_{\text{DRN}}( c_{k-1}|\boldsymbol{x}_i)) \, \underbrace{(\hat{F}_{\text{Baseline}}(c_{k}|\boldsymbol{x}_i)-\hat{F}_{\text{Baseline}}(c_{k-1}|\boldsymbol{x}_i))}_{\hat{b}_k(\boldsymbol{x}_i) \text{ in~\eqref{eq: baseline prob masses}}} + \mathcal{O}(1)  
  \nonumber \\
\end{align}
This approach effectively endows the DRN with a well-structured baseline, mirroring a Bayesian approach that is particularly beneficial in scenarios characterised by sparse or limited data \citep{rossi2003bayesian}.
The right panel of Figure~\ref{fig: KL Penalty} illustrates how the adjusted density function considers the baseline density, aligning partially with the baseline estimation.
As the regularisation coefficient $\alpha_1$ increases towards infinity, the DRN's density estimation $\hat{f}_{\text{DRN}}(y|\boldsymbol{x})$ converges to the baseline model's estimation $\hat{f}_{\text{Baseline}}(y|\boldsymbol{x})$.

\begin{figure}[H]
    \centering
    \includegraphics[width=0.8\textwidth]{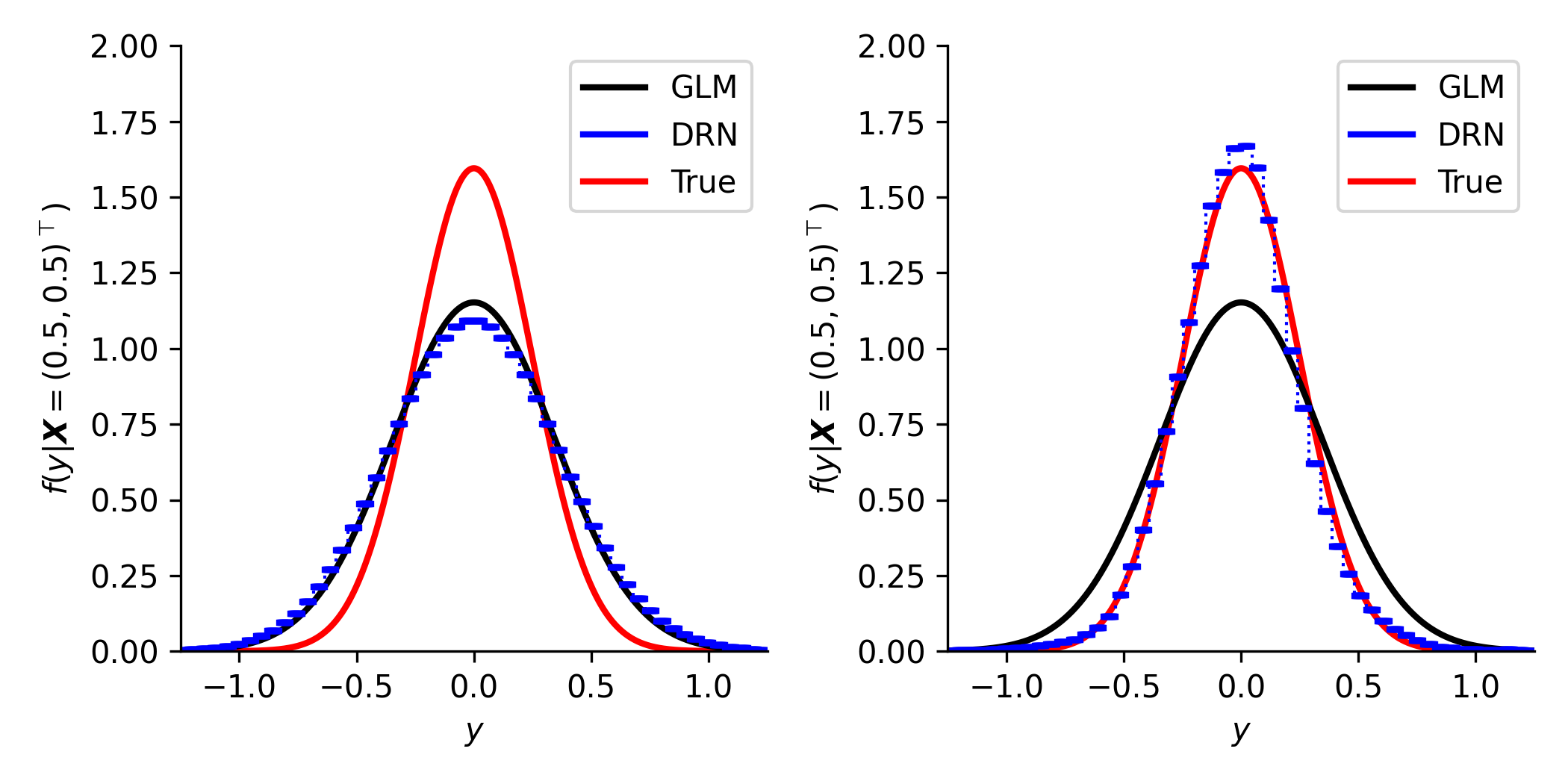}
    \caption{The comparison showcases DRN density functions under different roughness penalty intensities for the instance $X=(0.5, 0.5)^{\top}$.
    The left graph illustrates the outcome with extremely high regularisation ($\alpha_2 =1.0$), leading to a highly smooth but potentially less informative density estimate.
    The right graph, on the other hand, presents the impact of moderate regularisation ($\alpha_2 = 0.0005$), which offers a balanced density estimate that aligns more closely with the true distribution.
    In both graphs, the DRN density function (\textcolor{blue}{blue}) is compared against the baseline (\textcolor{black}{black}) and true density (\textcolor{red}{red}) functions.
    The KL divergence penalty coefficient is fixed at 0.005.}
    \label{fig: DV Penalty}
\end{figure}

A notable challenge with highly flexible deep learning models is their propensity to generate fluctuating estimated density functions, as shown in Figure~\ref{fig: KL Penalty}.
Ideally, a smoother density function is preferred for more effective interpretation and analysis, benefiting from the favourable statistical properties associated with either ordinary smooth or supersmooth distributions \citep{silverman1985some, fan1991optimal}.
Thus, we aim to enhance the ``overall smoothness" of the predicted density without neglecting the model's flexibility.
The concept of ``overall smoothness" here diverges from traditional definitions reliant on derivatives or characteristic functions.
Instead, it qualitatively addresses two main issues: significant jumps between adjacent density values and ``peaky" patterns.
To mitigate these abrupt fluctuations, we introduce a roughness penalty inspired by smoothing splines~\citep{craven1978smoothing}. This penalty is formulated as the square of the second-order difference in the density function, effectively reducing excessive and abrupt variations in the predicted densities:
\begin{align}
    {\Phi}_{c_0}^{c_K}(\hat{f}_{\text{DRN}}(y|\boldsymbol{x}_i))&=\sum_{k=1}^{K-2}\big((\hat{f}_{\text{DRN}}(c_{k+1}|\boldsymbol{x}_i)-\hat{f}_{\text{DRN}}(c_{k}|\boldsymbol{x}_i)) -(\hat{f}_{\text{DRN}}(c_k|\boldsymbol{x}_i)-\hat{f}_{\text{DRN}}(c_{k-1}|\boldsymbol{x}_i))\big)^2.
\end{align}
The impact of the substantial roughness penalty is showcased in the left panel of Figure~\ref{fig: DV Penalty}.
A large penalty value makes the estimated density overly ``smooth" and less informative, diminishing the model's precision for distributional forecasting.
Conversely, the right panel illustrates how a reasonable roughness penalty effectively mitigates excessive fluctuations and avoids unrealistic density jumps, thereby aligning the DRN's density estimate more closely with the actual density.
This approach yields a more stable and interpretable density estimate, enhancing the model's ability to accurately capture the underlying distribution without overfitting to noise.
Despite the advantages of predicting the full conditional distribution, the mean remains a critical distributional property for actuarial applications, including pricing (e.g. pure premium) and reserving (e.g. central estimate).
GLMs can provide unbiased estimators of the global population portfolio level mean, a feature known as the balance property.
However, this desirable attribute may not be preserved when training neural networks~\citep{wuthrich2021balance}.
In certain cases, a GLM can adequately capture the mean at individual levels.
Therefore, we integrate the component outlined in~\eqref{Mean Regularisation}, which refrains the DRN from significant deviations from the mean predictions generated by the baseline model.
This approach is identified as ``mean regularisation".
The implementation of mean regularisation can be practical when the baseline effectively models the mean.

\begin{remark}
    The KL divergence expression in~\eqref{eq: KL divergence} is effectively a weighted sum of differences. 
    In fact, two versions of the KL divergence exist, depending on the order with which the two distributions in~\eqref{KL Divergence Regularisation} are taken, because the first will become the weighting function found in~\eqref{eq: KL divergence}.
    When it is the distribution that is being fitted (in our case, the DRN density), one refers to ``reverse KL", whereas when it is the benchmark distribution (in our case, the GLM density), one refers to "forward KL". 
    The two formulations are not equivalent. 
    
    In traditional variational approximation~\citep{corduneanu2001variational}, reverse KL divergence is commonly employed for computational simplicity (because the weight is often Gaussian, simplifying calculations).
    Our method adopts the forward KL divergence, as demonstrated in Equation~\eqref{KL Divergence Regularisation}, effectively using a smooth baseline as the `weight' to emphasise the DRN's density deviations from the baseline. 
    Specifically, this weighting approach ensures logical consistency by prioritising alignment with the baseline forecasts, thus encouraging mean-seeking or mode-covering behaviours~\citep{murphy2022probabilistic}. 
    However, one could also employ reverse KL divergence in its approximate form, using the refined density as the weight. 
    This approach is feasible when zero-forcing or mode-seeking is ideal, i.e., when deviation from the baseline is desirable.
\end{remark}

\subsubsection{Model Evaluation} \label{sec: DRN Model Evaluation}

For assessing the DRN's distributional forecasting performance, we propose using the Continuous Ranked Probability Score (CRPS) \citep{gneiting2007strictly}, Negative Log-Likelihood (NLL), and Quantile Loss (QL).
The Root mean squared error (RMSE) is utilised to assess the accuracy of the central estimate.
The detailed formulas for these evaluation metrics are presented in~\ref{appendix: evaluation metrics}.
Note that both the CRPS and NLL or logarithmic score, are strictly proper scoring rules that encourage the predicted distribution to be equal to the actual underlying distribution so forecasts can be precise and optimal.

In addition to the numerical evaluation metrics, visual plots such as quantile residuals and calibration plots can be employed to assess model performance.
The quantile residuals plot \citep{dunn1996randomized} offers a diagnostic tool for evaluating the adequacy of fitting regression models.
Actuarial modelling widely uses it to measure a model's sufficiency in fitting.
To construct a quantile residuals plot, we first determine the quantiles of the observed response values, i.e., $\hat{F}^{-1}(y_i|\boldsymbol{x}_i)$'s.
Then we use the inverse of the standard normal distribution to transform quantiles into standard normal quantiles, i.e., compute $\Phi^{-1}(\hat{F}^{-1}(y_i|\boldsymbol{x}_i))$'s that are the quantile residuals.
Finally, we plot the obtained quantile residuals on a QQ-plot.
A well-fitted model's quantile residuals will closely align with the 45-degree line on a QQ-plot

Further, neural networks could be poorly calibrated: the $p\%$ prediction interval might not contain the accurate observation $p\%$ of the time.
A well-calibrated model is defined in Definition~\ref{def: probcalibration}, elaborated in~\ref{appendix: calibration}.
A calibration plot serves as a valuable tool for evaluating the calibration of a probabilistic model.
This plot contrasts the empirical cumulative probabilities with the predicted ones for all observations.
Ideally, a well-calibrated model should have the points closely aligned to the 45-degree line.
Deviations from this line can indicate an overestimation or underestimation of the predicted probabilities.
Strategies may include model refinement or applying recalibration techniques to enhance calibration, as \citet{Romano2019} and \citet{Kuleshov2018} discussed.

\section{Numerical Study using Synthetic Data} \label{Synthetic Data Analysis}

This section demonstrates the DDR’s capacity for distributional forecasting using a synthetic dataset. 
The process for data generation is detailed in Section~\ref{sec: Synthetic Data Generation}. 
Model specifications for competing distributional regression models, along with their hyperparameter tuning procedures, are detailed in Section~\ref{sec: Synthetic Dataset Model Specification}. 
The DDR's performance regarding distributional flexibility and interpretability is comprehensively examined in Section~\ref{sec: Distributional Flexibility and Interpretability}.

\subsection{Data Generation} \label{sec: Synthetic Data Generation}

This section describes the generation of a synthetic regression dataset, $\mathcal{D}={(\boldsymbol{x}_i, y_i)}_{i=1}^{n}$, representing the i.i.d.\ random vector $(\boldsymbol{X}, Y)$.
In this dataset, the feature vectors follow a multivariate Gaussian distribution, whereas the response variable combines gamma and lognormal distributions:
\begin{align}
    \boldsymbol{X} &= (X_1, X_2)^{\top} \sim \mathcal{N}(\boldsymbol{\mu}, \boldsymbol{\Sigma}), \nonumber \\
    Y &= \text{Gamma}(\mu(\boldsymbol{X}), \phi(\boldsymbol{X})) + \text{LogNormal}(\log(\mu(\boldsymbol{X})), \phi(\boldsymbol{X})),
\end{align}
where,
\begin{itemize}
    \item the features exhibit correlation, defined by parameters:
        \begin{align}
            \boldsymbol{\mu} = (0, 0)^{\top}, \boldsymbol{\Sigma} = \begin{bmatrix}
                0.25 ^2    & 0.25^3 \\
                0.25^3    &  0.25 ^2
            \end{bmatrix},
        \end{align}
        i.e., $\mu_1=\mu_2=0, \sigma_1 = 0.25, \sigma_2 = 0.25, \rho = 0.25$, and
    \item the dataset specifies mean and dispersion functions for $Y|\boldsymbol{X}$ as follows:
        \begin{align} \label{truemean}
            \mu(\boldsymbol{X}) &= \exp(-X_1 + X_2) \\
            \phi(\boldsymbol{X}) &= \exp(X_1)/(1+\exp(X_1\cdot X_2)).
        \end{align}
\end{itemize}
Drawing parallels to practical scenarios, such as motor insurance, gamma and lognormal random variables can be interpreted as administrative costs and actual incurred losses, respectively.
In this context, the feature $X_1$ can be conceptualised as the standardised age of the policyholder's license, while $X_2$ refers to the standardised value of the policyholder's vehicle.
Basing loss predictions exclusively on these two risk factors are insufficient.
Nonetheless, this simplification is deliberate, aiming to highlight two distinct behaviours: (i) the DRN makes reasonable adjustments to the mean predictions, mostly accurately but not perfectly forecasted by the baseline GLM; (ii) the DRN enhances the accuracy of predictions at various quantile levels, surpassing the GLM's performance, which falls short in modelling varying dispersion.

The dataset includes 12,000 observations for training, with both validation and test datasets comprising 4,000 observations each, all sampled randomly.

\subsection{Model Specification and Training} \label{sec: Synthetic Dataset Model Specification}

This section details the specifications and training procedures, including hyperparameter tuning, for the five models under comparison: GLM, CANN, MDN, DDR, and DRN.

The GLM is the standard in practice for actuarial regression problems.
For training the DRN within the proposed DRN framework, we select a gamma GLM with a log link function as the baseline. 
For all deep-learning models, we employ the \texttt{Adam} optimiser~\citep{kingma2014adam}.
Early stopping is set with a tolerance of $30$ epochs.
We employ the Leaky Rectified Linear Unit (\texttt{LeakyReLU}) as the activation function between the hidden layers, as explained in Appendix~\ref{appendix: activation function choices}.
The DDR's output layer utilises a \texttt{softmax} activation function to establish probabilities for each partitioned interval.
Both the CANN and our proposed DRN are utilising the gamma GLM as the baseline.
We pick the \texttt{exponential} activation function for the CANN's output layer to output the factors adjusting the means estimated by the gamma GLM.
The DRN's output layer uses a \texttt{Linear} activation function.
For the MDN’s output layer, we implement a \texttt{softmax} activation function to determine the mixing weights and two exponential activation functions to find each component's shape and scale parameters.

We employ the \texttt{skopt.gp\_minimize} function from \texttt{Scikit-Optimize} (\texttt{skopt}) for tuning hyperparameters. 
This method employs Gaussian process-based Bayesian optimisation, which efficiently navigates the trade-offs between exploration of the hyperparameter space and exploitation of known good regions. 
Table~\ref{tab:synthetic hyperparameters ranges} in~\ref{appendix: Hyperparameters Tables} presents a detailed comparison of the hyperparameter ranges employed across various advanced models.
The objective function for hyperparameter tuning is the average Continuous Ranked Probability Score (CRPS) loss. 
Each model undergoes a total of 125 evaluations with 25 random initialisations (known as \texttt{random\_starts}).
The final hyperparameters for the competing models are summarised in Table~\ref{tab: (Synthetic) Hyperparameters} in~\ref{appendix: Hyperparameters Tables}.

\subsection{Distributional Flexibility and Interpretability} \label{sec: Distributional Flexibility and Interpretability}

This section explores the DRN framework's distributional flexibility and interpretability, two major aspects we are concerned about in distributional regression.
Section~\ref{Synthetic Dataset Model Evaluation} displays its exceptional distributional flexibility by leveraging various evaluation metrics.
Section~\ref{SyntheticModelIntepretability} showcases how we integrate the SHAP framework into explaining our proposed model's predictions beyond the mean.

\subsubsection{Flexibility: Model Performance}\label{Synthetic Dataset Model Evaluation}

To assess both probabilistic and point forecasts, we employ four evaluation metrics: Continuous Ranked Probability Score (CRPS), Negative Log-Likelihood (NLL), Root Mean Squared Error (RMSE), and 90\% Quantile Loss (90\% QL).
Table~\ref{tab: (Synthetic) Evaluation Metrics} below demonstrates the results of different models.
The proposed DRN model, refining over the gamma GLM, has outperformed gamma MDN across all evaluation metrics for the test dataset.
It yields significant improvements for average NLL, CRPS and 90\%QL, respectively, compared to the GLM.
\begin{table}[H]
    \centering
    \caption{Model comparisons based on various evaluation metrics. To robustly assess if the DRN significantly enhances distributional forecasting compared to the baseline GLM, the Wilcoxon Signed-Rank Test, introduced by \citet{wilcoxon1945wilcoxontest}, is applied to NLL and CRPS scores.
    The blue highlighted metrics represent the best-performing model for each respective column.
    This non-parametric method evaluates the statistical significance of median differences between the performance metrics of the two models.
    The statement `$\text{Model1} < \text{Model2}$' implies testing the null hypothesis that the median metrics of both models are equal against the alternative that $\text{Model1}$'s median is lower. 
    A $p$-value less than 0.05 is indicated with one star (*), less than 0.01 with two stars (**), and less than 0.001 with three stars (***).}
    \label{tab: (Synthetic) Evaluation Metrics}
    \scalebox{0.95}{
    \begin{tabular}{l|cccc|cccc}
    \toprule
    \toprule
    &  \multicolumn{4}{c}{$\mathcal{D}_{\text{Validation}}$}& \multicolumn{4}{c}{ $\mathcal{D}_{\text{Test}}$}\\ 
     \cmidrule{2-5}  \cmidrule{6-9} $\text{Model}$ $\backslash$ $\text{Metrics}$ & NLL & CRPS & RMSE & 90\% QL & NLL & CRPS & RMSE & 90\% QL \\ \midrule
    GLM &  1.2825 &  0.5302  & 0.9985  & 0.2090  & 1.2696  & 0.5205  & 0.9896  & 0.2060 \\ 
    CANN &  1.2827 &  0.5299  & 0.9972  & 0.2093  & 1.2700  & 0.5206  & 0.9900  & 0.2063 \\ 
    MDN &  \textcolor{blue}{1.2497} &  \textcolor{blue}{0.5253}  & \textcolor{blue}{0.9960}  & \textcolor{blue}{0.2069}  & 1.2347  & 0.5177  & 0.9913  & 0.2040 \\
    DDR &  1.9666 &  0.5257 & 0.9965 & 0.2070  & 1.9318  & 0.5179  & 0.9919  & 0.2037 \\ 
    DRN &  \textcolor{black}{1.2532} &  0.5259  & 0.9971  & 0.2072  &  \textcolor{blue}{1.2344}  &  \textcolor{blue}{0.5166}  &  \textcolor{blue}{0.9893}  &  \textcolor{blue}{0.2034} \\ 
    \bottomrule
    \bottomrule
     &  \multicolumn{4}{c}{$\mathcal{D}_{\text{Validation}}$}& \multicolumn{4}{c}{$\mathcal{D}_{\text{Test}}$}\\
     \cmidrule{2-5}  \cmidrule{6-9} $\text{Model}$ $\backslash$ $\text{Metrics}$ & NLL & CRPS & RMSE & 90\% QL & NLL & CRPS & RMSE & 90\% QL \\ \midrule
    DRN $<$ GLM &  (***) &  (***) &  (***) &  (***)&  (***) &  (***)&  (***) &  (***)\\
    DRN $<$ CANN &  (***) &  (***) &  (***) &  (***)&  (***) &  (***)&  (**) &  (***)\\
    DRN $<$ MDN &  (***) &  (***) &  (**) &  (***)&  (***) &  (***)&  (***) &  (***)\\
    DRN $<$ DDR &  (***) &  (***) &  (***) &  (***)&  (***) &  (***)&  (***) &  (***)\\
    \bottomrule
    \bottomrule
    \end{tabular}
    }
\end{table}

\begin{figure}[H]
    \centering
    \includegraphics[width=0.46\textwidth]{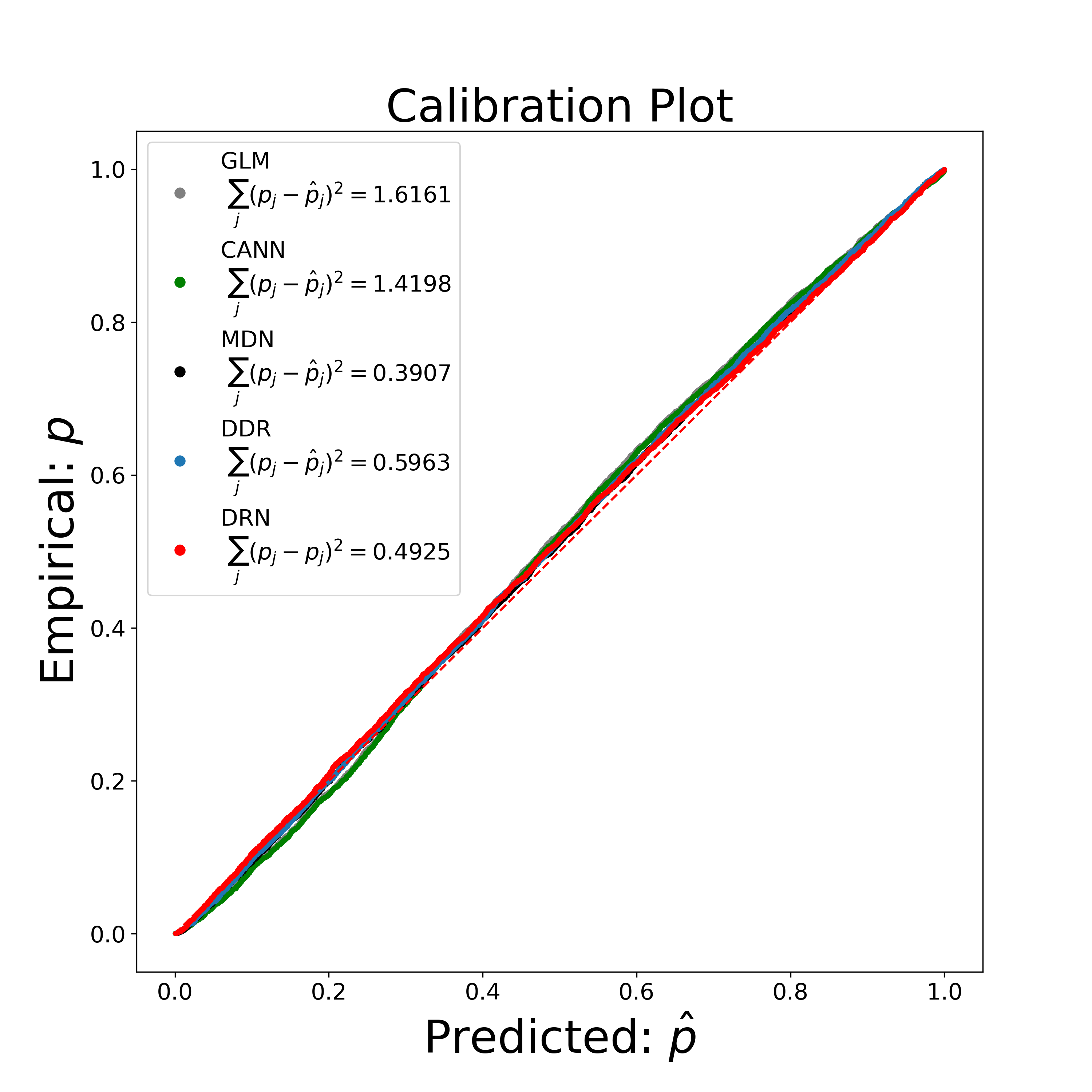}
    \includegraphics[width=0.5\textwidth]{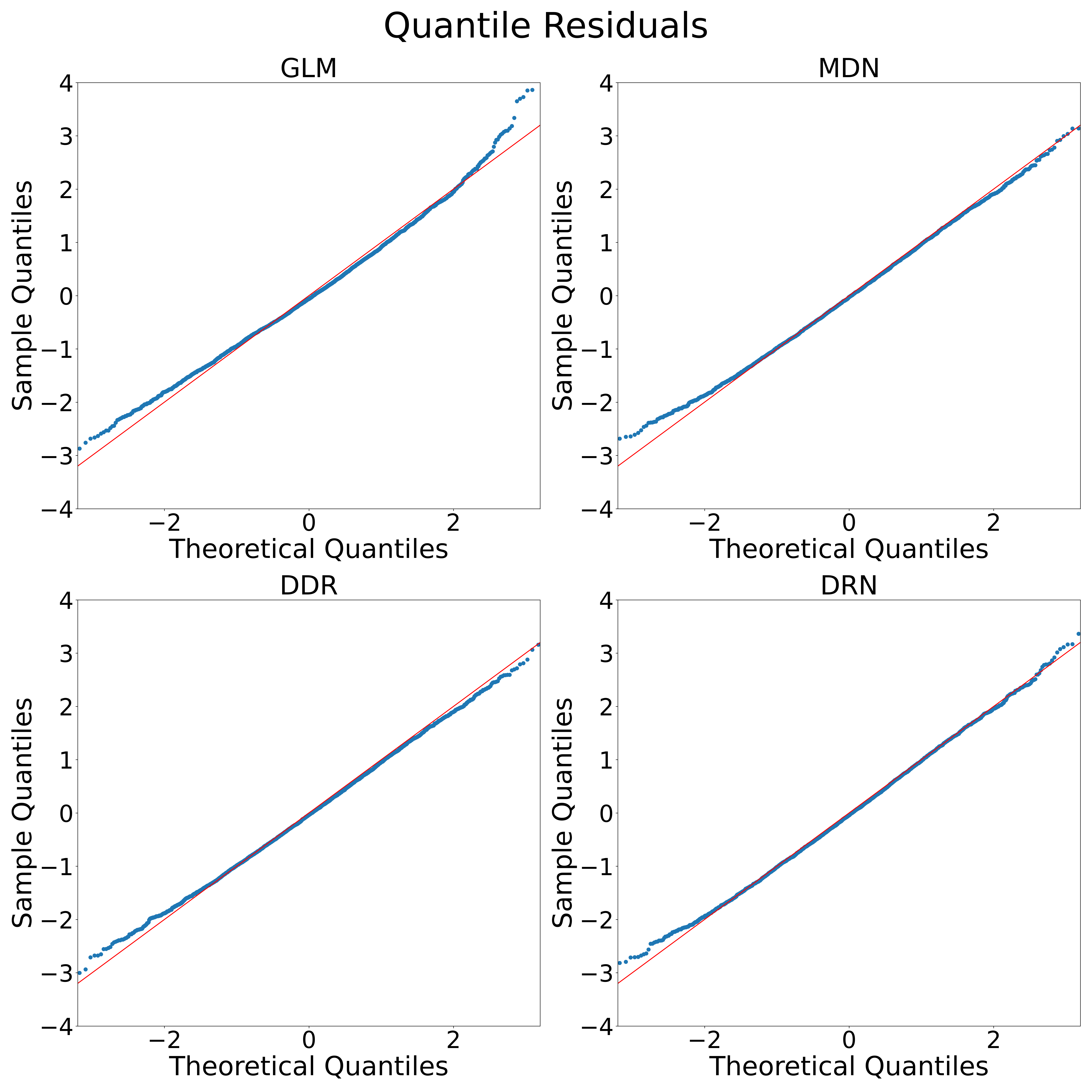}
    \caption{Calibration (left) and quantile residuals (right) plots generated on the test dataset.}
    \label{fig: (Synthetic) Quantile Residuals and Calibration Plots}
\end{figure}

The left panel of Figure~\ref{fig: (Synthetic) Quantile Residuals and Calibration Plots} presents the calibration plots, further highlighting the superiority of the DRN, both qualitatively and quantitatively.
The points are closely aligned to the 45-degree line, and the calibration score for the DRN is lower than the GLM's calibration score.
The right panel displays the quantile residual plots.
The points are relatively aligned to the 45-degree line for the DRN, highlighting the superiority of the DRN in distributional forecasting as it exhibits the most optimal Q-Q plot.

\subsubsection{Interpretability: Forecast Transparency}\label{SyntheticModelIntepretability}

The scope of interpretability also comprises two primary categories: local and global; see~\citep{molnar2020interpretable, burkart2021survey}. 
Local interpretability concentrates on clarifying the rationale behind specific model predictions, enabling detailed analysis of particular decisions. 
Meanwhile, global interpretability aims to reveal the model's holistic logic, offering insights into its behaviour across a wide range of scenarios. 
The GLM and GAMLSS are regarded as interpretable for two major reasons: i) the modelling ``simplicity" is guaranteed due to the small number of distributional parameters required to define the distribution; ii) distributional properties are relatively straightforward transformations of the features, facilitating a clear understanding of feature contributions to predictions of mean, variance, etc.

\begin{figure}[H]
    \centering
    \includegraphics[width=0.55\textwidth]{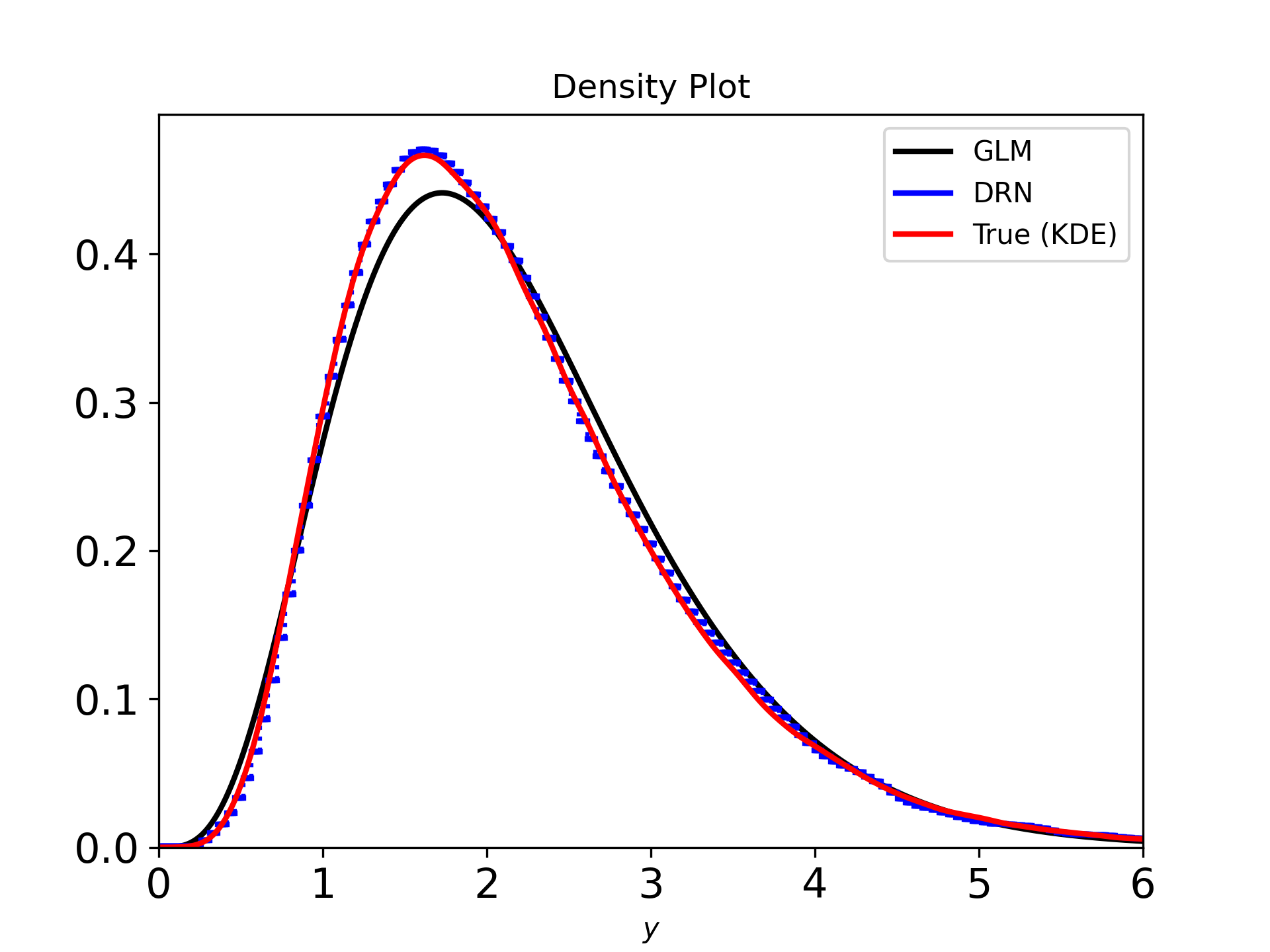}
    \caption{Comparison of predicted conditional densities of $Y|\boldsymbol{x}^{*}$ by the baseline model (\textcolor{black}{black}) and the DRN (\textcolor{blue}{blue}) with the true density (\textcolor{red}{red}), where $\boldsymbol{x}^*=(0.1, 0.1)^{\top}$.}
    \label{fig: (Synthetic) Example Density}
\end{figure}

\begin{figure}[H]
    \centering
    \includegraphics[width=0.49\textwidth]{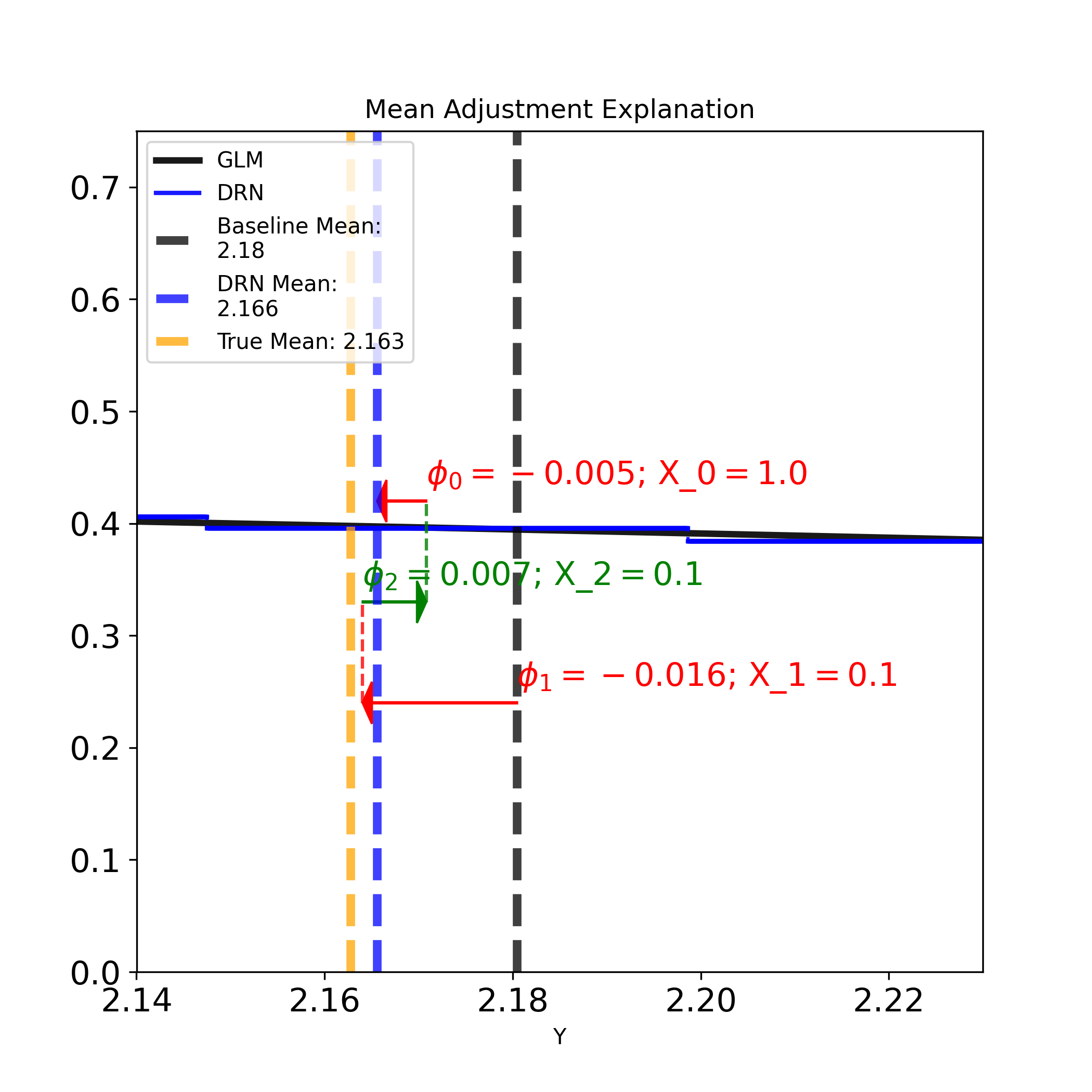}
    \includegraphics[width=0.49\textwidth]{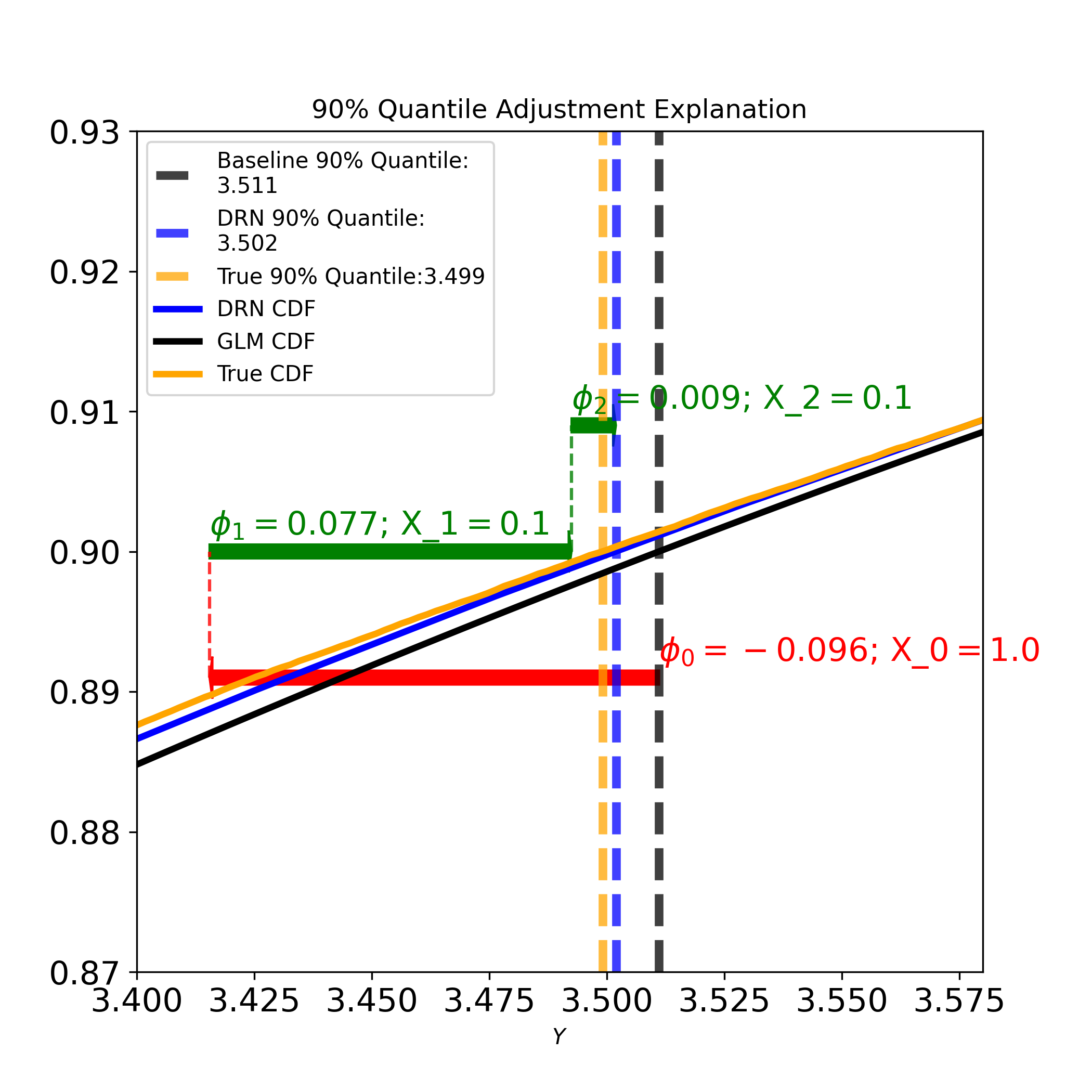}
    \caption{The instance of investigation is $\boldsymbol{x}^{*}=(0.1, 0.1)^{\top}$.
    The left panel utilises the Kernel SHAP values to explain the mean adjustment made by the DRN.
    The right panel utilises the Kernel SHAP values to explain the 90\% quantile adjustment made by the DRN.
    The DRN is decreasing the 90\% quantile prediction.
    The first feature's Kernel SHAP value is $\phi_1=0.077$, which implies that it contributes to the 90\% quantile prediction $0.077$ more in the DRN than within the baseline model.}
    \label{fig: (Synthetic) Mean and Quantile Adjustment SHAP}
\end{figure}

\begin{figure}[H]
    \centering
    \includegraphics[width=0.47\textwidth]{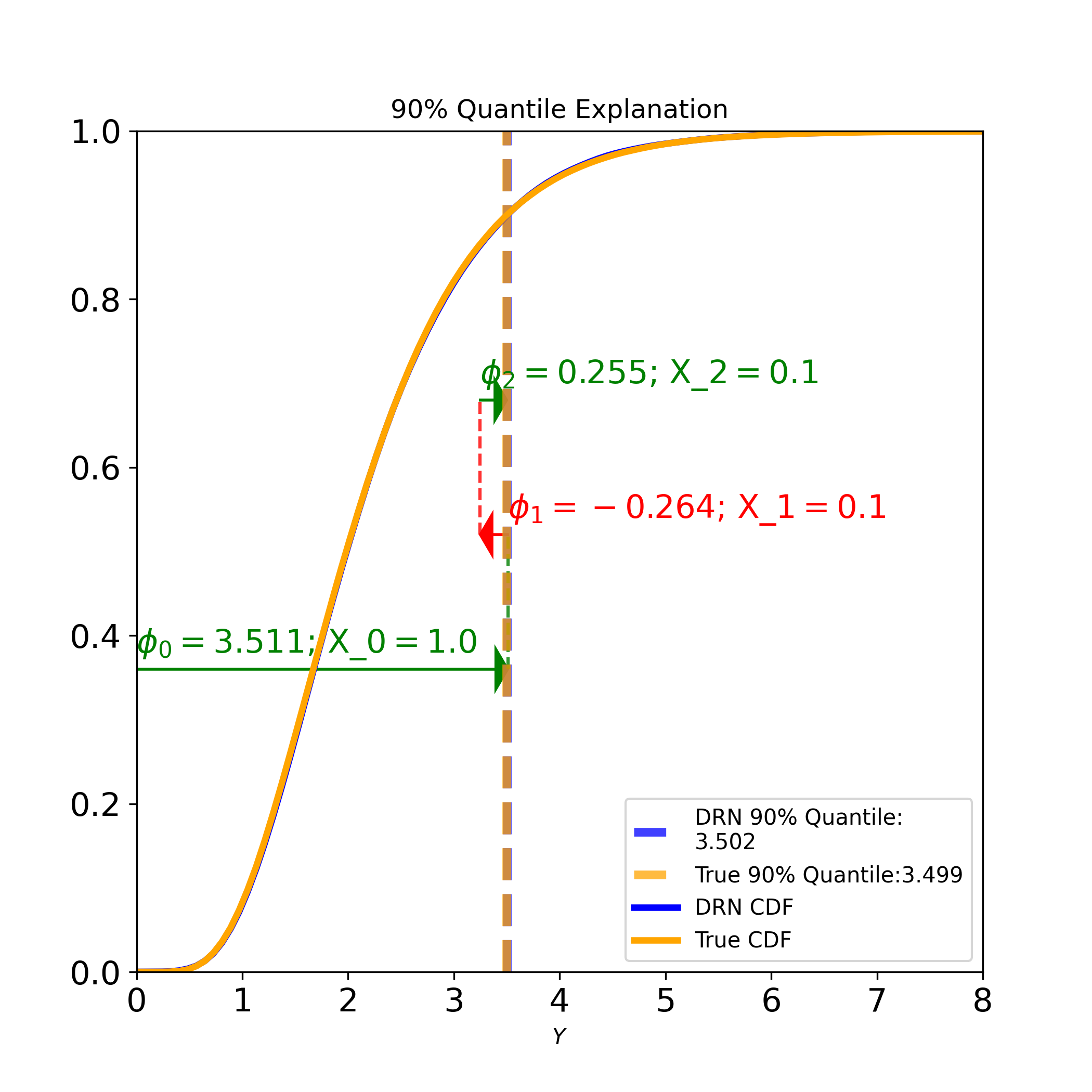}
    \caption{This plot utilises the Kernel SHAP values to explain how the DRN predicts the 90\% quantile of $Y|\boldsymbol{x}^*$, where $\boldsymbol{x}^*=(0.1, 0.1)^{\top}$.
    For example, $\phi_2=0.255$ implies that $X_2$ being 0.1 leads to a $0.255$ increment over the average 90\% quantile.}
    \label{fig: (Synthetic) Quantile Explanation SHAP}
\end{figure}

\begin{figure}[H]
    \centering
    \includegraphics[width=0.81\textwidth]{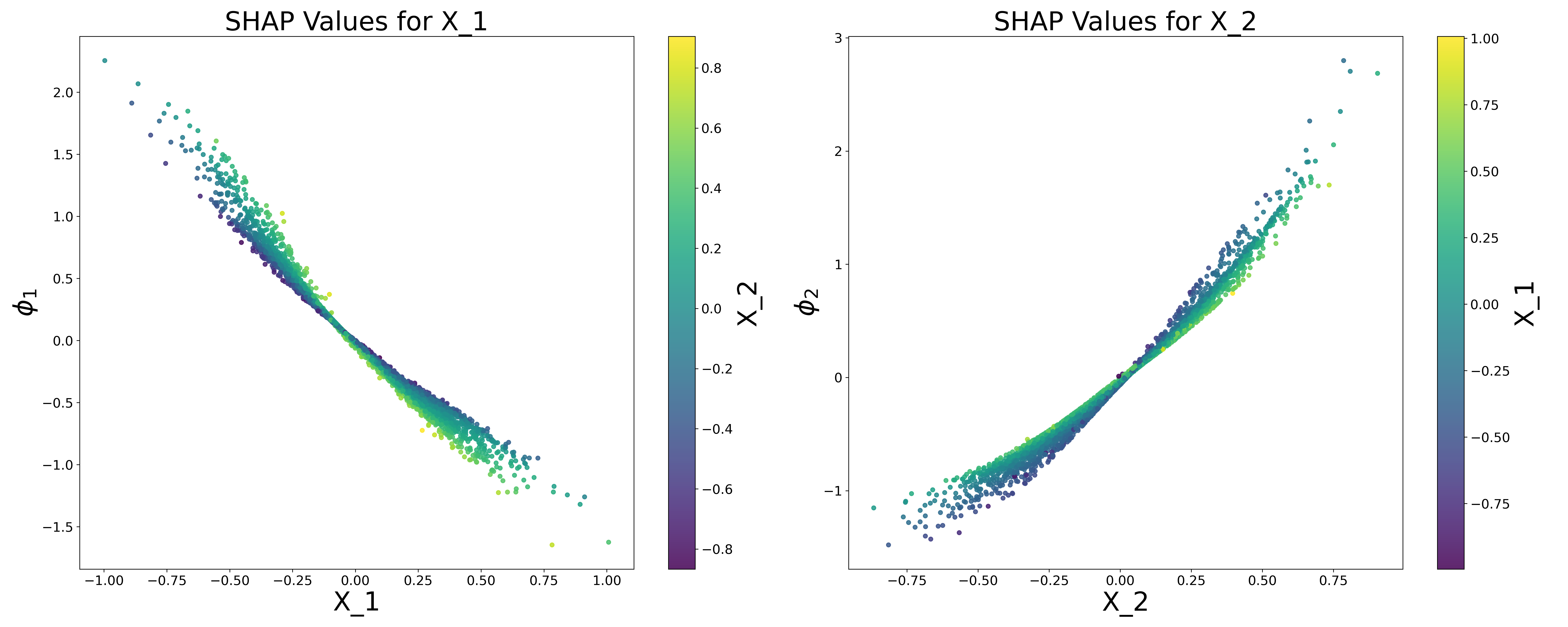}
    \caption{SHAP dependence plot for mean predictions made by the DRN.
    (Left): The graph illustrates how, with an increase in $X_1$, the DRN model's prediction for the mean decreases relative to the average mean prediction.
    (Right): Demonstrates that as $X_2$ decreases, there is a corresponding decrease in the DRN's mean prediction compared to the average, albeit at a diminishing rate.}
    \label{fig: (Synthetic) Quantile Explanation SHAP Dependence}
\end{figure}

To address the first aspect, the DRN framework enables users to visualise how the distribution deviates from the baseline, enhancing transparency.
Such a functionality effectively mitigates the concern of the DRN's reliance on a significantly larger set of distributional parameters.
 Through this visualisation, users gain clarity on the model's predictions despite the complexity introduced by the increased parameter count.
For illustration, we examine an instance  $\boldsymbol{x}^*=(x
^*_1, x^*_2)^{\top}=(0.1, 0.1)^{\top}$, characterised by feature values near their means.
Given our control over the synthetic dataset generation, we can directly compare the true distribution with those generated by the baseline and DRN models.
Figure~\ref{fig: (Synthetic) Example Density} showcases this comparison: the true distribution is represented in red, the baseline in black, and the DRN's prediction in blue.
Qualitatively, the DRN's distribution aligns more closely with the true distribution than the baseline's.
Such a visualisation allows us to trust the model's distributional flexibility.
Concerning the second aspect, we integrate Kernel SHAP \citep{lundberg2017unified}, detailed in~\ref{appendix: Kernel SHAP}, to augment transparency in predicting the key distributional properties.
This leads to more informed decisions on whether to trust the less inherently interpretable DRN.
Furthermore, the inherent ``refining" characteristic of our framework allows for a detailed examination of the neural network's adjustments, both at individual (local) and overall (global) levels.
Building on the preceding example, we expect the DRN to (i) perform subtle but correct modifications to the mean predictions and (ii) refine the baseline model's quantile predictions for enhanced accuracy.
The first anticipation is exemplified in the left panel of Figure~\ref{fig: (Synthetic) Mean and Quantile Adjustment SHAP}, which showcases the mean adjustment decomposition utilising the Kernel SHAP values.
The mean prediction is closer to the true mean, confirming our initial expectation.
The interpretation of how the DRN predicts the mean of $Y|\boldsymbol{x}^*$ is relatively straightforward:
\begin{align} \label{eq: (Synthetic) Mean Interpretation}
    \mathbb{E}_{Y|\boldsymbol{X}}[Y|\boldsymbol{x}^*; \boldsymbol{w}, \boldsymbol{\beta}] &\approx \mathbb{E}_{Y|\boldsymbol{X}}[Y|\boldsymbol{x}^*; \boldsymbol{\beta}] + \underbrace{(- 0.005)}_{\phi_0 \equiv \text{Difference in } \mathbb{E}_{\boldsymbol{X}}[\mathbb{E}[Y|\boldsymbol{X}]]} + \underbrace{(-0.016)}_{\phi_1 \equiv \text{Additional impact of }X_1} + \underbrace{(+0.007)}_{\phi_2 \equiv \text{Additional impact of }X_2} \nonumber \\
    &=
\exp(\beta_0 + \beta_1 \cdot (0.1) + \beta_2 \cdot (0.1)) - 0.005 - 0.016 + 0.007.
\end{align}
To verify our second hypothesis, we generate the right panel of Figure~\ref{fig: (Synthetic) Mean and Quantile Adjustment SHAP}, focusing on adjustments at the 90\% quantile level.
For this specific instance ($X_1=0.1, X_2 =0.1$), it is observed that the DRN is correcting the GLM's underestimation of the impacts of $X_1$ and $X_2$ ($\phi_1, \phi_2>0$).
On a global level, discrepancies arise between the GLM and the DRN regarding predictions at the 90\% quantile level:
\begin{align}
\phi_0\Big(v^{\text{90\% Quantile}}_{\mathcal{M}_{\boldsymbol{w}},\boldsymbol{\beta}} -v^{\text{90\% Quantile}}_{\mathcal{M}_{\boldsymbol{\beta}}}, \boldsymbol{X}\Big)&=
\mathbb{E}_{\boldsymbol{X}}[Q_{Y|\boldsymbol{X}}(0.9|\boldsymbol{X}; \boldsymbol{w}, \boldsymbol{\beta})]
-
\mathbb{E}_{\boldsymbol{X}}[Q_{Y|\boldsymbol{X}}(0.9|\boldsymbol{X}; \boldsymbol{\beta})]
\nonumber
\\
&\approx \frac{1}{n}\Bigg(
\sum_{i=1}^{n} Q_{Y|\boldsymbol{X}}(0.9| \boldsymbol{x}_i; \boldsymbol{w}, \boldsymbol{\beta})
-
Q_{Y|\boldsymbol{X}}(0.9| \boldsymbol{x}_i; \boldsymbol{\beta})
\Bigg)
\approx -0.096.\nonumber
\end{align}
Such a difference in the ``baseline" value implies that the GLM struggles to capture quantile levels accurately, evidenced by the discrepancy in the 90\% quantile loss demonstrated in Table~\ref{tab: (Synthetic) Evaluation Metrics}.
The improvement in loss reduction achieved by the DRN emphasises its capacity to overcome the limitations inherent in the baseline GLM, specifically its inability to model non-constant and complex dispersion effectively.
Deriving a human-interpretable expression, as illustrated in Equation~\eqref{eq: (Synthetic) Mean Interpretation}, proves challenging due to the inherent complexity involved in expressing the quantiles of a gamma distribution, even when we know the mean and dispersion.
Therefore, we further employ Kernel SHAP decomposition to achieve an additive explanation for the DRN’s prediction of the 90\% quantile (Figure~\ref{fig: (Synthetic) Quantile Explanation SHAP}):
\begin{align}
    Q_{Y|\boldsymbol{X}}(0.9|\boldsymbol{x}^{*}; \boldsymbol{w}, \boldsymbol{\beta})&\approx Q_{Y|\boldsymbol{X}}(0.9|\boldsymbol{x}^{*}; \boldsymbol{\beta})
    + \underbrace{(-0.096)}_{\phi_0 \equiv \text{Difference in } \mathbb{E}_{\boldsymbol{X}}[Q_{Y|\boldsymbol{X}}(0.9|\boldsymbol{X})] (\text{Figure~\ref{fig: (Synthetic) Mean and Quantile Adjustment SHAP}})} + \underbrace{(+0.077)}_{\phi_1(\text{Figure~\ref{fig: (Synthetic) Mean and Quantile Adjustment SHAP}})} + \underbrace{(+0.009)}_{\phi_2 (\text{Figure~\ref{fig: (Synthetic) Mean and Quantile Adjustment SHAP}})} \nonumber \\
    &\approx \underbrace{(+3.511)}_{\mathbb{E}_{\boldsymbol{X}}[Q_{Y|\boldsymbol{X}}(0.9|\boldsymbol{X}; \boldsymbol{w}, \boldsymbol{\beta})](\text{Figure~\ref{fig: (Synthetic) Quantile Explanation SHAP}})}
    +\underbrace{(-0.264)}_{X_1 \text{'s Contribution}
    (\text{Figure~\ref{fig: (Synthetic) Quantile Explanation SHAP}})}
    +\underbrace{(0.255)}_{
    X_2 \text{'s Contribution} (\text{Figure~\ref{fig: (Synthetic) Quantile Explanation SHAP}})}.
\end{align}
The close alignment between the estimated CDF by the DRN model and the true CDF, as showcased in Figure~\ref{fig: (Synthetic) Quantile Explanation SHAP}, highlights the model's effectiveness in accurately capturing various quantiles.
To address the global interpretability of the DRN, it's crucial to examine the contributions of $X_1$ and $X_2$ to the distributional properties across all instances.
Mathematically speaking, a SHAP dependence plot represents the set of points
$
    \Big\{\big(x_{i,j}, \phi_{j}\big(v^{\text{Mean}}_{\mathcal{M}}, \boldsymbol{x}_i\big)\big)\Big\}_{i=1}^{n}
$
for $j\in \{1, 2\}$.
Figure~\ref{fig: (Synthetic) Quantile Explanation SHAP Dependence} reveals that both $\phi_{1}(v^{\text{Mean}}_{\mathcal{M}}, \boldsymbol{X})$ and $\phi_{2}(v^{\text{Mean}}_{\mathcal{M}}, \boldsymbol{X})$ exhibit approximately linear dependencies on $X_1$ and $X_2$, respectively.
Notably, $\phi_{2}(v^{\text{Mean}}_{\mathcal{M}}, \boldsymbol{X})$ does show a slight quadratic relationship as well.
However, it is essential to note that these observations are based on the assumption of feature independence.
Kernel SHAP approximates the value functions using the marginal distributions of $X_1$ and $X_2$ rather than their conditional distributions $X_1|X_2$ and $X_2|X_1$. 
This means that the interaction between $X_1$ and $X_2$ is not fully captured, potentially leading to deviations from the expected relationships.
Hence, one should interpret and leverage the created plots carefully and objectively.
Thus, this paper profoundly acknowledges the limitations of employing Kernel SHAP for analysing the features' impact on distributional forecasting.
It recognises the potential benefits of exploring conditional value function approaches for a more nuanced understanding of the model's interpretability.

\section{Illustrating DRN's Practical Application with Real Data}
\label{Real Data Analysis}

To showcase the practical applicability of our proposed DRN framework in an insurance context, we evaluate its performance in modelling claim severity.
For this purpose, we utilise an actuarial dataset with the response variable as claim amounts (positive random variables) and high-dimensional features consisting of both categorical and numerical features.
Specifically, we employ the \texttt{freMPL1} dataset from the \texttt{R} package \texttt{CASdatasets}, focusing on modelling claim amounts given the features.

We selected this particular dataset for three main reasons.
First, the empirical conditional density of claim amounts, given the features, exhibits characteristics such as multimodality and heavy-tailedness.
These features represent the frequent challenges in actuarial modelling.
Second, the dataset possesses a moderate number of observations and a mix of numerical and categorical features.
It comprises 3,265 observations, three continuous features, four binary features, and eleven categorical features.
Lastly, this dataset was employed by \citet{delong2021gamma} to investigate the Gamma MDN, one of our benchmark models.

Section~\ref{EDA} delves into Data Preprocessing and Exploratory Data Analysis.
In Section~\ref{ModelTrainingReal}, we detail the procedure for model training, leading into Section~\ref{ModelPerformanceReal}, which examines the performance of these models.
Section~\ref{ModelInterpretabilityReal} focuses on the model's distributional interpretability.

\subsection{Data Preprocessing and Exploratory Data Analysis} \label{EDA}

\begin{figure}[H]
    \centering
    \includegraphics[width=0.43\textwidth]{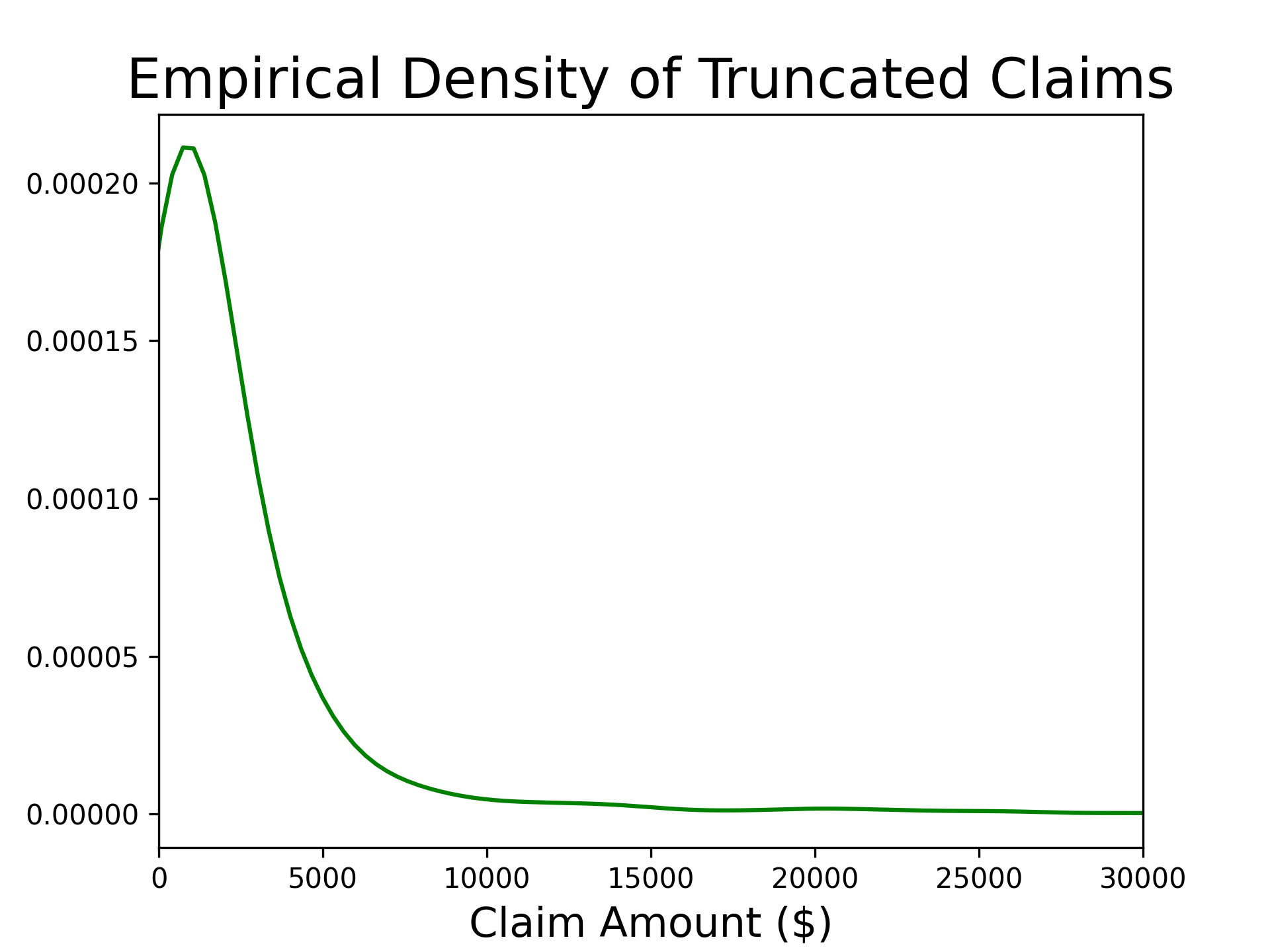}
    \vspace{-0.6em}
    \caption{The empirical density of claim amounts truncated at \$30,000.}
    \label{fig: (Real) Empirical Density}
\end{figure}

Figure~\ref{fig: (Real) Empirical Density} presents the empirical density of claim amounts censored at a maximum value of \$30,000.
It's important to note that the applied truncation and density smoothing technique somewhat conceal the distribution's notable characteristics, such as heavy-tailedness and multimodality.
Comprehensive data cleaning and transformation steps are employed.
First, we scale the response variable, \texttt{ClaimAmount}, which is reduced by a factor of 1,000.
Additionally, we discard several columns, namely \texttt{RecordBeg}, \texttt{RecordEnd}, \texttt{ClaimInd}, and \texttt{Garage}, with the exclusion of \texttt{Garage} due to many missing values.
Regarding the feature \texttt{VehAge} (vehicle age), which was originally provided in categories, it is mapped to specific numerical values.
Notably, `6-7' is assigned with a value of 6 and `10+' is assigned with a value of 11.
We convert categorical speed ranges into a numeric sequence for the \texttt{VehMaxSpeed} (maximum speed range of the vehicle) feature.
For example, `1-130 km/h' is converted to an integer value of 1, `130-140 km/h' has an integer value of 2, and so on.

For feature preprocessing, we apply one-hot encoding to categorical variables such as \texttt{HasKmLimit}, \texttt{Gender}, \texttt{SocioCateg} and \texttt{MariStat}, among others.
The dataset is subsequently divided into training and testing sets, with 60\%, 20\%, and 20\% for training, validation, and testing, respectively.

\subsection{Model Training and Hyperparameter Tuning}\label{ModelTrainingReal}

We select a gamma GLM with a log link function as the baseline for the DRN.
The early stopping patience parameter is fixed at $30$ for all neural networks.
We employ the first-order \texttt{Adam} optimiser.
The activation functions for the hidden and output layers resemble the ones highlighted in Section~\ref{sec: Synthetic Dataset Model Specification}.
The comparison of the hyperparameter ranges used across various advanced models is presented in Table~\ref{tab:real hyperparameters ranges} within Appendix~\ref{appendix: Hyperparameters Tables}.
Each model is subject to 200 evaluations.
The DRN model undergoes 30 random initialisations, while the other models undergo 25.
This difference is deliberate and recommended because the DRN model has the highest number of hyperparameters.
The tuned hyperparameters are summarised in Table~\ref{tab: (Real) Hyperparameters} in~\ref{appendix: Hyperparameters Tables}.

\subsection{Distributional Flexibility} \label{ModelPerformanceReal}

Table~\ref{tab: (Real) Evaluation Metrics} demonstrates the results of different models regarding a few evaluation metrics.
The proposed DRN, when refined over the gamma GLM, has the lowest values of NLL, CRPS, RMSE and 90\% QL NLL, indicating superior predictive performance based on the test set.

\begin{table}[H]
    \centering
    \caption{
    Model comparisons based on various evaluation metrics. 
    The blue highlighted metrics represent the best-performing model for each respective column.
    The Wilcoxon Signed-Rank Test \citep{wilcoxon1945wilcoxontest} is applied to NLL and CRPS scores. 
    The statement `$\text{Model1} < \text{Model2}$' implies testing the null hypothesis that the median metrics of both models are equal against the alternative that $\text{Model1}$'s median is lower. 
    A $p$-value less than 0.05 is indicated with one star (*), less than 0.01 with two stars (**), and less than 0.001 with three stars (***).}
    \label{tab: (Real) Evaluation Metrics}
    \scalebox{0.95}{
        \begin{tabular}{l|cccc|cccc}
        \toprule
        \toprule
        &  \multicolumn{4}{c}{$\mathcal{D}_{\text{Validation}}$}& \multicolumn{4}{c}{$\mathcal{D}_{\text{Test}}$}\\
         \cmidrule{2-5}  \cmidrule{6-9} $\text{Model}$ $\backslash$ $\text{Metrics}$ & NLL & CRPS & RMSE & 90\% QL & NLL & CRPS & RMSE & 90\% QL \\ \midrule
        GLM &  1.9359 &  1.5115  & 6.8326  & 1.0083  & 1.9601  & 1.7953  & 8.8029  & 1.1950 \\ 
        CANN &  2.2808 &  1.2365  & 3.4851  & 0.6533  & 3.4421  & 1.7045  & 8.7166  & 1.0650 \\
        MDN &  1.5901 &  1.1507  & 11.5481  & 0.6865  & 1.8006  & 1.6494  & 8.7711  & 1.0490 \\ 
        DDR &  1.8072 &  1.1372  & \textcolor{blue}{3.1037}  & 0.6616  & $\infty$  & 1.6103  & 8.6573  & 1.0222 \\ 
        DRN &  \textcolor{blue}{1.0829} &  \textcolor{blue}{0.8883}  & \textcolor{black}{3.3641}  & \textcolor{blue}{0.5139}  & \textcolor{blue}{1.1219}  & \textcolor{blue}{1.3199}  & \textcolor{blue}{8.6046}  & \textcolor{blue}{0.9408} \\ 
    \bottomrule
    \bottomrule
     &  \multicolumn{4}{c}{$\mathcal{D}_{\text{Validation}}$}& \multicolumn{4}{c}{$\mathcal{D}_{\text{Test}}$}\\
     \cmidrule{2-5}  \cmidrule{6-9} $\text{Model}$ $\backslash$ $\text{Metrics}$ & NLL & CRPS & RMSE & 90\% QL & NLL & CRPS & RMSE & 90\% QL \\ \midrule
    DRN $<$ GLM 
    &  (***) &  (***) &  (***) &  (***)
    &  (***) &  (***) &  (***) &  (***)\\
    DRN $<$ CANN 
    &  (***) &  (***) &  (***) &  (***)
    &  (***) &  (***) &  (***) &  (***)\\
    DRN $<$ MDN 
    &  (***) &  (***) &  (***) &  (***)
    &  (***) &  (***) &  (***) &  (***)\\
    DRN $<$ DDR 
    &  (***) &  (***) &  (***) &  (***)
    &  (***) &  (***) &  (***) &  (***)\\
    \bottomrule
    \bottomrule
    \end{tabular}
    }
\end{table}

\begin{figure}[H]
    \centering
    \includegraphics[width=0.46\textwidth]{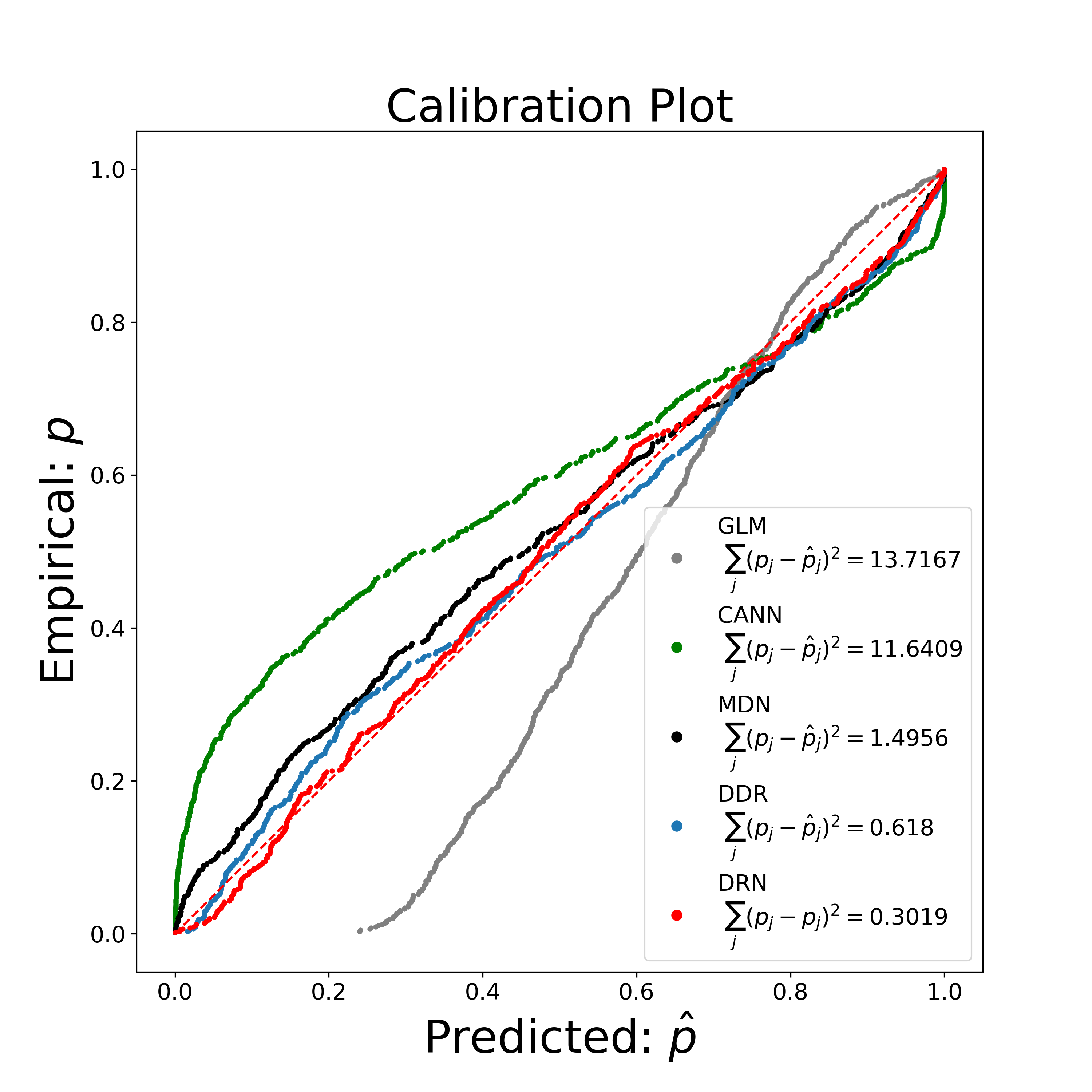}
    \includegraphics[width=0.5\textwidth]{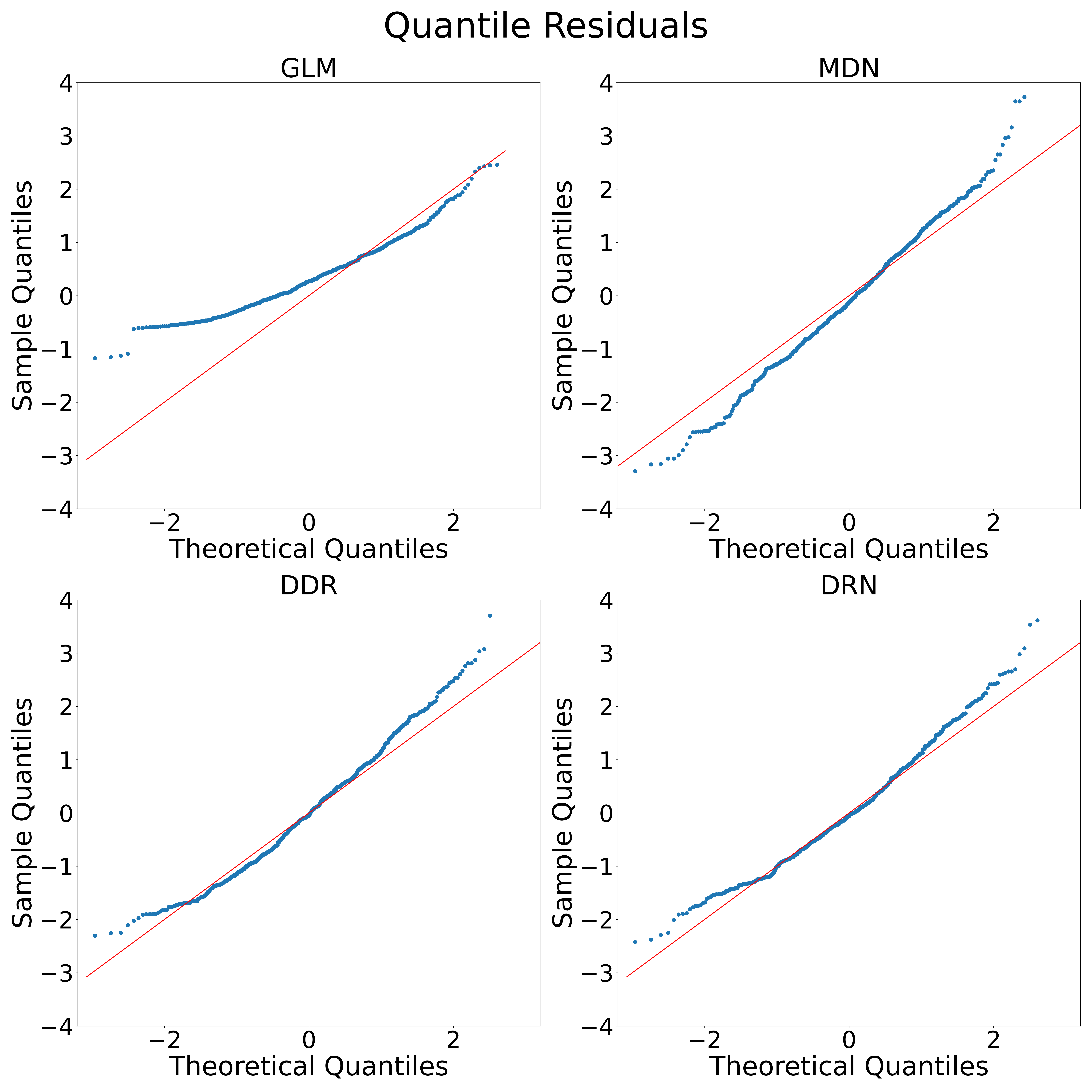}
    \caption{Calibration (left) and quantile residuals (right) plots generated on the test dataset.}
    \label{fig: (Real) Calibration and Quantile Residuals Plots}
\end{figure}

Figure~\ref{fig: (Real) Calibration and Quantile Residuals Plots} displays the calibration and quantile residual plots.
The DRN model's calibration points, shown in the left panel, are close to the 45-degree line on average, and the calibration score for the DRN is the lowest.
This shows that the DRN is a well-calibrated model.
The DRN model also performs competitively in the quantile residual plot, showcased in the right panel.

\subsection{Distributional Interpretability} \label{ModelInterpretabilityReal}

This section examines the DRN's distributional interpretability from two key perspectives: local and global. These are demonstrated in Sections~\ref{Real Dataset Local Interpretability} and~\ref{Real Dataset Global Interpretability}, respectively.

\subsubsection{Local Interpretability}
\label{Real Dataset Local Interpretability}

\begin{figure}[H]
    \centering
    \includegraphics[width=0.49\textwidth]{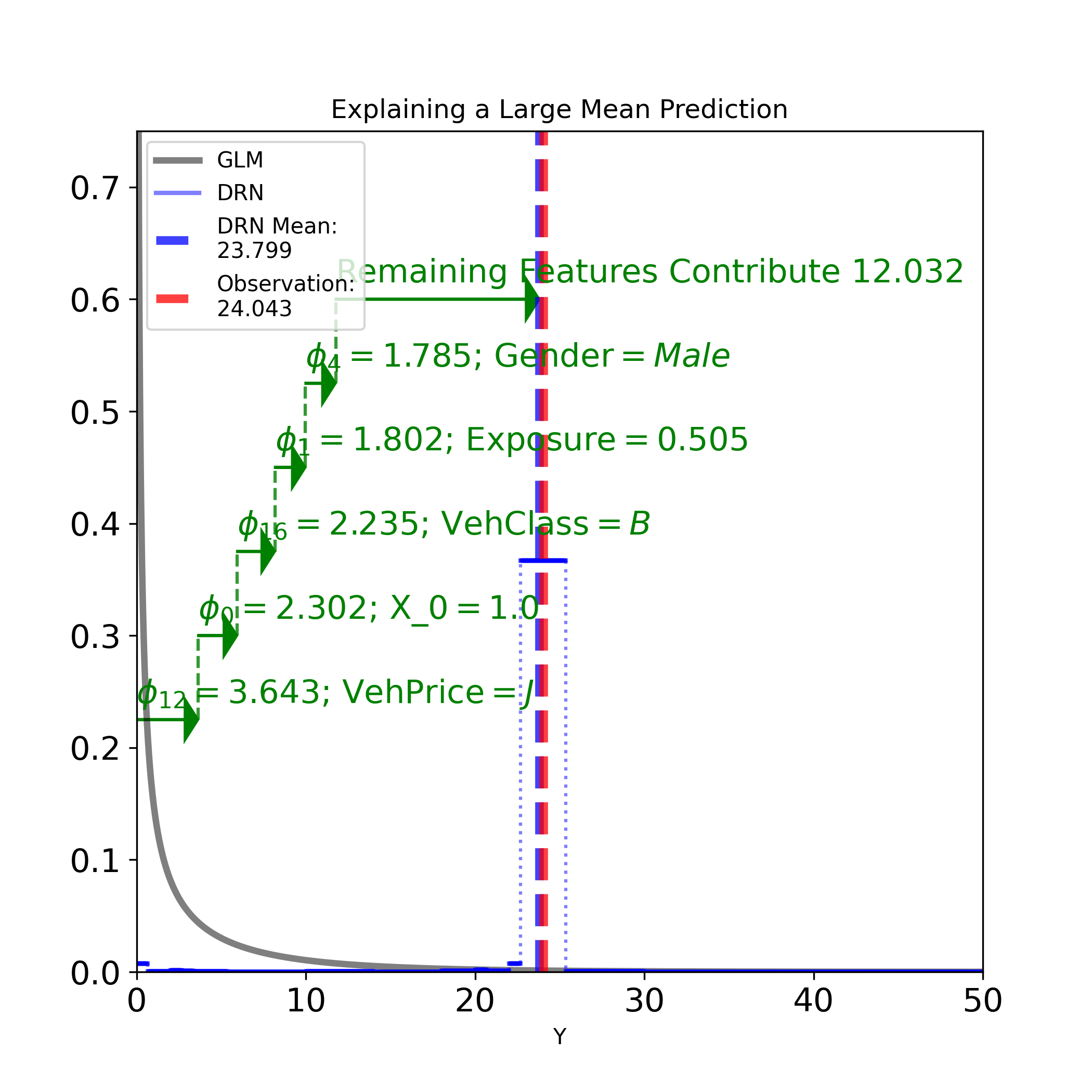}
    \includegraphics[width=0.5\textwidth]{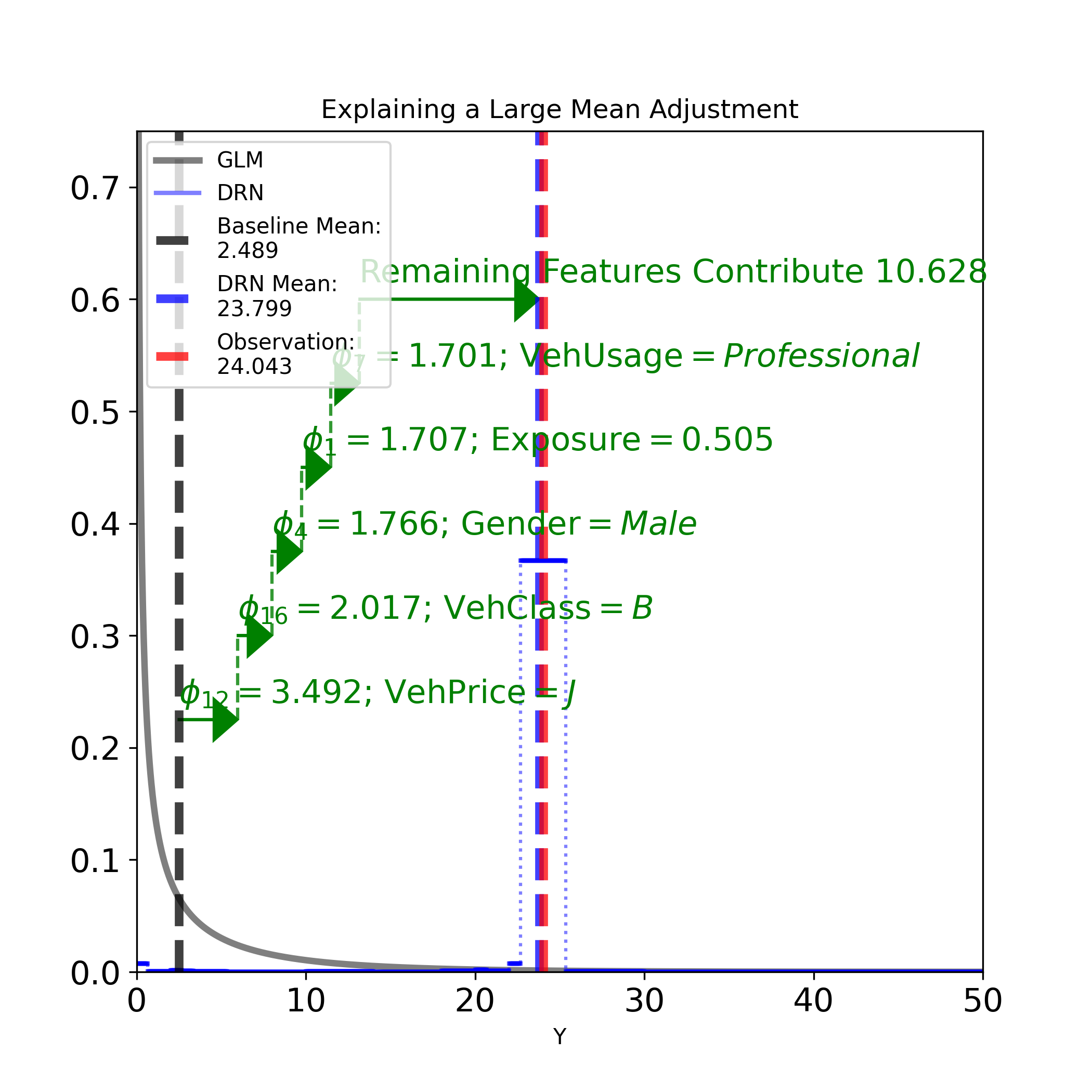}
    \vspace{-0.6em}
    \caption{
    The left panel breaks down the DRN's second-largest mean prediction in the test set by decomposing it from the ground up using SHAP values. 
    The right panel illustrates how the GLM mean is adjusted upwards to match the DRN predicted mean, also using SHAP values.}
    \label{fig: (Real) Mean Adjustment and Explanation SHAP}
\end{figure}

For illustrative purposes, we focus on the policyholder with the second-largest mean adjustment in the test dataset.
The left panel of Figure~\ref{fig: (Real) Mean Adjustment and Explanation SHAP} shows the density functions predicted by the baseline and the DRN.
We can see that the DRN forecasts a distribution with a heavier tail.
Meanwhile, we decompose the DRN's mean prediction utilising Kernel SHAP values.
In order of impact, this policyholder's top risk factors are \texttt{VehPrice}, \texttt{VehClass}, \texttt{Exposure}, and \texttt{Gender}.
The magnitude of the SHAP values determines this ranking.
The policyholder's vehicle price, class, exposure, and gender significantly influence the mean prediction.
Trust in the DRN may be enhanced by verifying that the explanations align with intuitive and expert expectations.

The right panel of Figure~\ref{fig: (Real) Mean Adjustment and Explanation SHAP} demonstrates the corresponding mean adjustment decomposition.
Comparing the left panel to the right panel, the identical significance of features and signs of the SHAP values reveals the baseline model's inability to fully capture the features' impacts on the mean prediction for this particular instance.
Furthermore, the positive effects on the mean prediction attributed to vehicle price, class and exposure appear to have been underestimated or wrongly attributed by the baseline model.

In other cases, the refinements can be less severe, as demonstrated in Figure~\ref{fig: (Real) More Mean Adjustments SHAP}, where we plot the adjustments to two other examples:

\begin{figure}[H]
    \centering
    \includegraphics[width=0.48\textwidth]{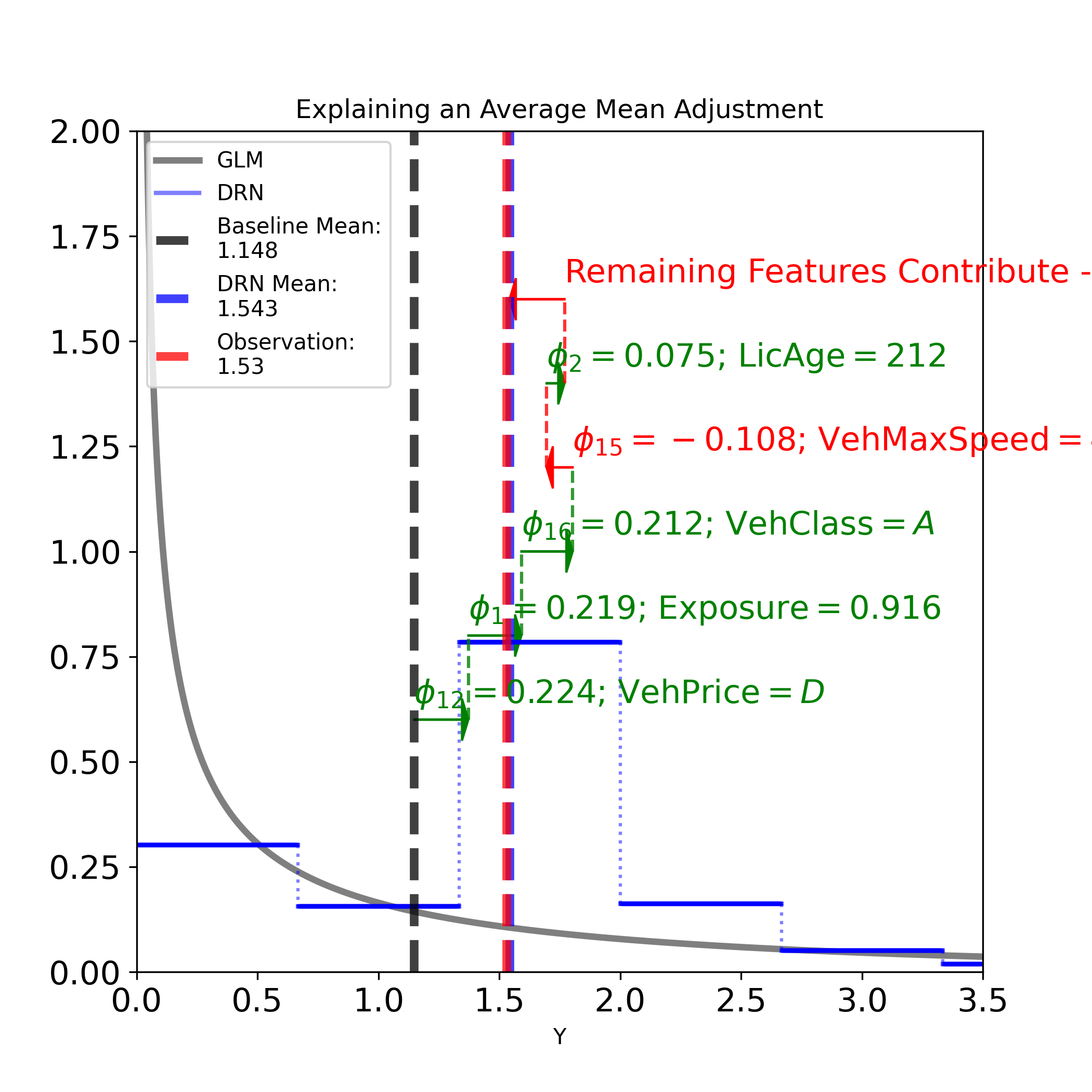}
    \hspace{0.3em}
    \includegraphics[width=0.49\textwidth]{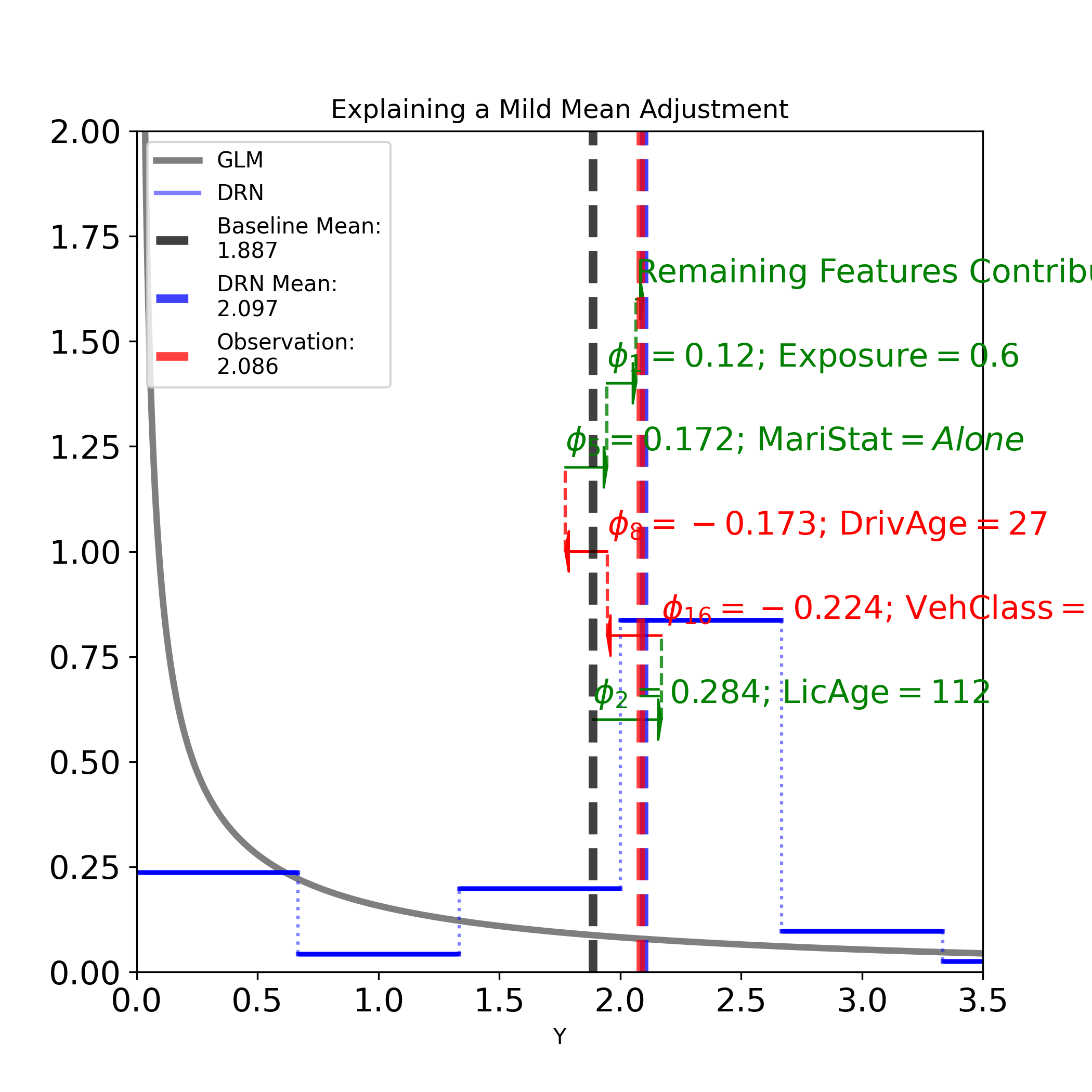}
    \vspace{-0.6em}
    \caption{
    Two (different) examples of the refinements.} 
    \label{fig: (Real) More Mean Adjustments SHAP}
\end{figure}

\subsubsection{Global Interpretability} \label{Real Dataset Global Interpretability}

\begin{figure}[H]
    \centering
    \includegraphics[width=0.52\textwidth]{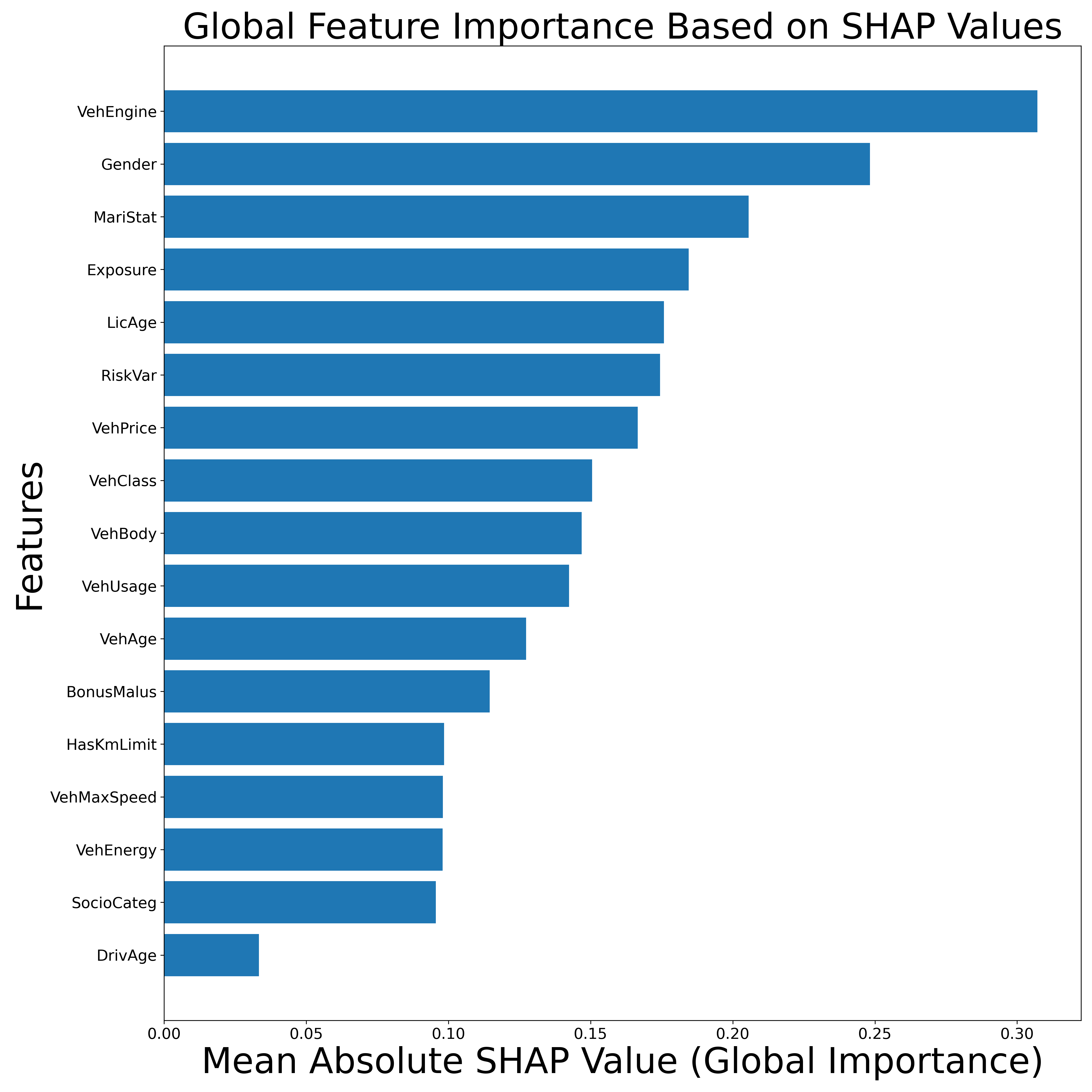}
    \includegraphics[width=0.46\textwidth]{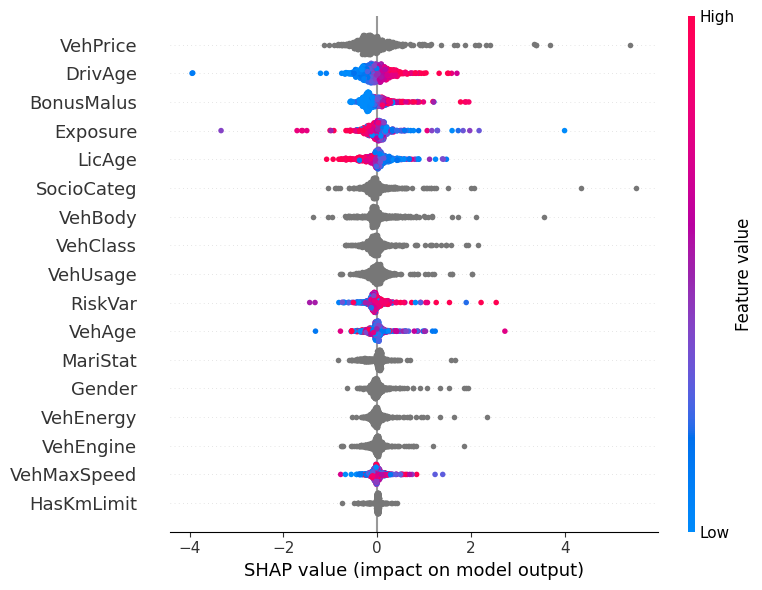}
    \vspace{-0.6em}
    \caption{Left Panel: This displays the global importance of features in the DRN's mean prediction, ranked by their mean absolute SHAP values. Right Panel: A summary plot illustrating how these features impact the DRN's mean prediction.}
    \label{fig: (Real) Global Importance and Beeswarm Plots}
\end{figure}

\begin{figure}[H]
    \centering
    \includegraphics[width=0.85\textwidth]{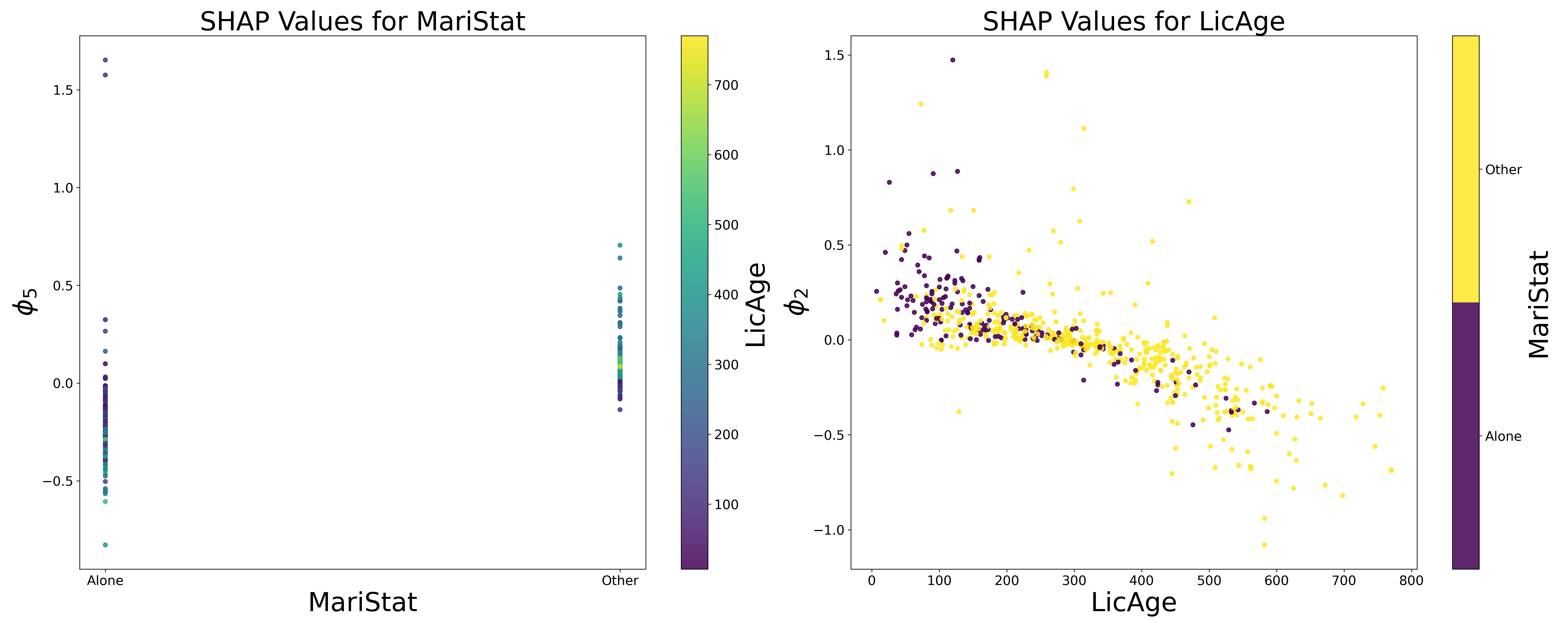}
    \caption{DRN Mean Prediction: SHAP dependence plot between \texttt{MariStat} and \texttt{LicAge}.}
    \label{fig: DRN Beeswarm and SHAP Dependence}
\end{figure}

The left panel of Figure~\ref{fig: (Real) Global Importance and Beeswarm Plots} highlights the global importance of features identified by the DRN while predicting the mean.
It recognises \texttt{VehEngine}, \texttt{Gender}, \texttt{MaritStat}, \texttt{Exposure} and \texttt{LicAge} as the top five features for mean prediction in terms of the mean absolute SHAP values.
The higher the mean absolute SHAP value, the more important the feature is.
The right panel provides insight into how the features impact the DRN's mean prediction outcome.
It is a beeswarm summary plot displaying the distributions of feature values and their global impacts.
It reveals a positive association of the mean prediction with increasing \texttt{DrivAge}, \texttt{BonusMalus} and \texttt{RiskVar}.
On the other hand, \texttt{LicAge}, \texttt{VehAge} and \texttt{Exposure} exhibit dual impacts: increasing the mean prediction at smaller values and decreasing it at larger values.
Figure~\ref{fig: DRN Beeswarm and SHAP Dependence} displays clear associations between \texttt{MariStat} and \texttt{LicAge}.
The plots from Figures~\ref{fig: (Real) Global Importance and Beeswarm Plots} and~\ref{fig: DRN Beeswarm and SHAP Dependence} comprehensively show how feature values individually and potentially interactively contribute to the mean prediction.

\begin{figure}[H]
    \centering
    \includegraphics[width=0.85\textwidth]{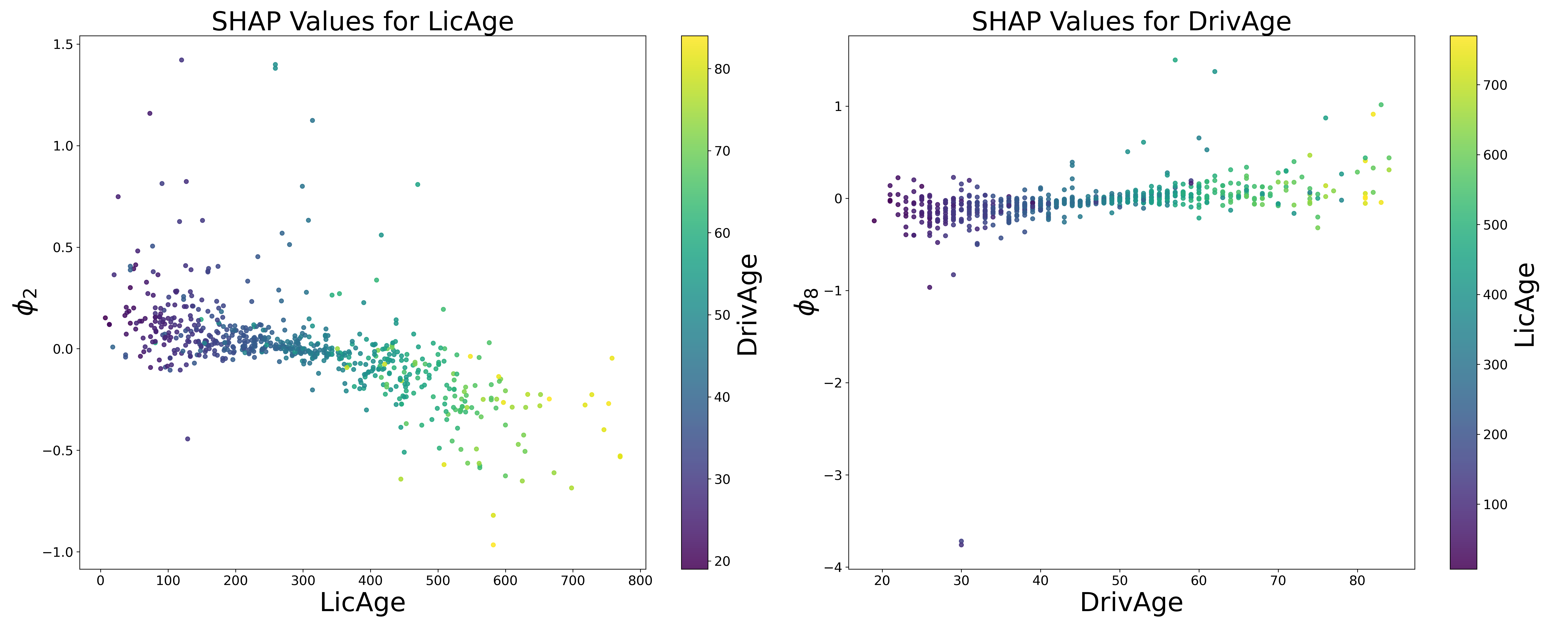}
    \caption{Mean adjustment SHAP dependence plot for \texttt{LicAge} and \texttt{DrivAge}.}
    \label{fig: (Real) SHAP Dependence Mean Adjustment}
\end{figure}

To explain the adjustments made by the DRN on a global level, we generate the so-called mean adjustment SHAP dependence plot, as demonstrated in Figure~\ref{fig: (Real) SHAP Dependence Mean Adjustment}.
Mathematically, we plot the points $\{(x_{i,j}, {\phi}_j(v_{\mathcal{M}_{\boldsymbol{w}}}^{\text{Mean}} - v_{\mathcal{M}_{\boldsymbol{\beta}}}^{\text{Mean}}))\}_{i=1}^{n}$ for $j\in\{\text{\texttt{LicAge}}, \text{\texttt{DrivAge}}\}$.
A detailed analysis of the SHAP value distributions yields several interpretations and suggestions: i) incorporate a polynomial term in the GLM for \texttt{LicAge} or \texttt{DrivAge}; ii) explore interaction terms and possible dependencies between the features; iii) reevaluate feature importance based on mean absolute SHAP values, i.e., we can drop unimportant features during refinement to improve interpretability and reduce feature dimensionality if feasible.
We refrain from implementing these possibilities in this paper while acknowledging the potential of these approaches to create a positive feedback loop between the baseline and neural network components.

\section{Conclusions} \label{Conclusion}

In this paper, we introduce the Distributional Refinement Network (DRN), a novel distributional regression model that integrates inherently interpretable parametric models with flexible semiparametric deep learning techniques. 
The DRN framework demonstrates potential in optimising distributional flexibility while preserving a level of interpretability, effectively navigating a main challenge in distributional regression.
The DRN refines distributional forecasts from a baseline regression model (a GLM in this paper) to accurately capture the varying impacts of features across all quantiles.
We detail our methodological approach for training the DRN, covering the specialised partitioning algorithm, the selection rationale for the loss function, and the integration of regularisation techniques.

While this paper's primary motivation was the need for improved distributional forecasting methods relative to traditional GLM-based models, our method can alternatively also be seen as an improvement on (Deep) Distribution Regression. 
Specifically, the DRN offers three significant improvements over DDR through baseline model integration: (i) it optimises efficiency and the capability to accurately forecast all crucial distributional properties by using reasonable initial values from the baseline model and applying effective regularisation techniques; (ii) the integration defines the predicted distribution beyond the refining region, overcoming the DDR model's limitations in analysing tail behaviour; and (iii) the predictions of various distributional properties can be divided between the transparent baseline and the deep learning component, offering a viable pathway to enhance the baseline by analysing these refinements.

\section*{Acknowledgements}

An earlier version of this paper was presented at the 2023 Australasian Actuarial Education and Research Symposium in Wellington, New Zealand.
The authors are grateful for constructive comments received from colleagues who attended those events. 

Eric Dong acknowledges financial support by the Australian Government Research Training program, as well the UNSW Business School through supplementary scholarships. 
The views expressed herein are those of the authors and are not necessarily those of the supporting organisations.

\section*{Data and Code}

The synthetic datasets were generated in Python. The real dataset \texttt{freMPL1} was obtained from the \texttt{R} package \texttt{CASdatasets}. Results presented in this paper can be replicated using the Python code available at \url{https://github.com/agi-lab/drn}.

\section*{Python package \texttt{drn}}

The GitHub repository can be found at \url{https://github.com/EricTianDong/drn}. 
This package includes different distributional regression models, such as GLM, CANN, MDN, and DDR, as well as the proposed DRN model from this paper.
A Python package named \texttt{drn} will be available on PIP soon.

\bibliographystyle{elsarticle-harv}

\newpage

\appendix

\section{Partitioning Algorithm} \label{appendix: partitioning algorithm}

\begin{algorithm}[htb]
    \caption{Partitioning Algorithm: Merge Cutpoints}
    \label{alg: MergeCutpointsMinObs}
    \begin{algorithmic}[1]
    \State \textbf{Input:} Initial list of cutpoints $L_{\text{Raw}} = \{c_0^*, \ldots, c^*_{\Tilde{K}}\}$, training dataset $\mathcal{D}_{\text{Train}}$, minimum observations $M$.
    \State \textbf{Output:} Refined list of cutpoints $L_{\text{Merged}}$ ensuring at least $M$ observations per interval.
    \State Initialise $L_{\text{Merged}} \gets \{c_0^*\}$ and $\texttt{left} \gets 0$.
    \For{$\texttt{right} \gets 1$ to $\Tilde{K}-1$}
        \State Count the number of observations $y_i \in \mathcal{D}_{\text{Train}}$ within $[c_\texttt{left}^*, c_\texttt{right}^*)$.
        \State Count the number of observations $y_i \in \mathcal{D}_{\text{Train}}$ within $[c_\texttt{right}^*, c_{\tilde{K}})$.
        \If{both counts are $\geq M$}
            \State Append $c_\texttt{right}^*$ to $L_{\text{Merged}}$ and update $\texttt{left} \gets \texttt{right}$.
        \EndIf
    \EndFor
    \State Append $c_{\Tilde{K}}^*$ to $L_{\text{Merged}}$.
    \State \Return $L_{\text{Merged}}$
    \end{algorithmic}
\end{algorithm}

\section{Regularisation Dataset Generation}
\label{appendix: regularisation synthetic dataset}

The synthetic dataset for the regularisation study in Section~\ref{sec: Loss Function and Regularisation Techniques} consists of $40,\!000$ independent and identically distributed (i.i.d.) samples from the random vector $(\boldsymbol{X}, Y)$, with the feature vector $\boldsymbol{X}=(X_1, X_2)^{\top} \in \mathbb{R}^{2}$.
Each feature $X_j$ is independently distributed according to a Gaussian distribution $\mathcal{N}(0, 0.5^2)$ for $j\in\{1, 2\}$.
The conditional distribution of $Y$ given $\boldsymbol{X}$ is also Gaussian, defined as $Y|\boldsymbol{X}\sim\mathcal{N}(-X_1+X_2, (0.5\cdot(X_1^2+X_2^2))^2)$.
For this study, the employed baseline model is linear regression, while the DRN adopts a two-hidden-layer architecture with 128 neurons each, a dropout rate of 0.2, a learning rate of 0.002, and a batch size of 200.
The cutpoint ratio is 0.05, with 5 minimum observations required in each partitioned interval.
The data is split into 60\%, 20\%, and 20\% for training, validation, and testing, respectively.
An early stopping mechanism, with a patience parameter of $30$ epochs, is implemented to prevent overfitting.
This means the training ceases if validation loss has not improved within $30$ epochs.

\section{Activation Functions} \label{appendix: activation function choices}

The Leaky Rectified Linear Unit (\texttt{LeakyReLU}) is recommended as the activation function of choice between the input, the first hidden layer, and between hidden layers.
Other choices, such as the Recitified Linear Unit (\texttt{ReLU}) and the hyperbolic tangent function (\texttt{tanh}) function, can also be viable.
The preference for \texttt{LeakyReLU} serves as a remedy for the ``Dying ReLU" issue encountered with the standard \texttt{ReLU} activation function \citep{alzubaidi2021review}.
In scenarios where \texttt{ReLU} is employed, neurons may cease learning altogether due to an absence of gradient descent updates when their outputs are consistently zero.
\texttt{LeakyReLU} addresses this by introducing a small, positive gradient for negative inputs.

The \texttt{Linear} activation function is selected for the output layer, promoting model flexibility and reducing numerical instability.
This is highly recommended to maximise model flexibility and minimise numerical instability.
While alternatives like the exponential function (\texttt{exp}) and the hyperbolic tangent function (\texttt{tanh}) are theoretically viable, they are generally not recommended. 
The use of the exponential function can potentially result in exploding gradients, and the hyperbolic tangent may fail to achieve optimal refinement.

\section{Hyperparameters for the Numerical Examples}
\label{appendix: Hyperparameters Tables}

\begin{table}[H]
\centering
\caption{
(Synthetic Dataset) Hyperparameter Ranges Across Models.
 $Real(a, b)$ means the hyperparameter space is continuous from $a$ to $b$ (inclusive). $Int(a, b)$ means the hyperparameter space consists of discrete integer values from $a$ to $b$ (inclusive).
}
\label{tab:synthetic hyperparameters ranges}
\begin{tabular}{lcccc}
\toprule
\toprule
Hyperparameter       & CANN                 & MDN                  & DDR                  & DRN                  \\ 
\midrule
Learning Rate       & $Real(0.0002, 0.01)$   & $Real(0.0002, 0.01)$    & $Real(0.0002, 0.01)$    & $Real(0.0002, 0.01)$   \\
Batch Size           & [$128, 256, 512$]   & $[128, 256, 512] $  & $[128, 256, 512] $    & $[128, 256, 512]$   \\
Dropout Rate         & $Real(0.0, 0.5)$      & $Real(0.0, 0.5)$      & $Real(0.0, 0.5)$      & $Real(0.0, 0.5)$     \\
Number of Layers     & $Int(1, 4)$        & $Int(1, 4)$        & $Int(1, 4)$        & $Int(1, 4)$        \\
\multirow{2}{*}{Neurons per Layer}    & $[16, 32, 64,$    &  $[16, 32, 64,$    &  $[16, 32, 64,$     &  $[16, 32, 64,$ \\
& $128, 256, 512]$    &   $128, 256, 512]$    &   $128, 256, 512]$    &   $128, 256, 512]$ 
\\
Mixture Components   & -                    & $Int(1, 10)$       & -                    & -                    \\
Proportion           & -                    & -                    & $\boldsymbol{p}_{\text{DDR}}$\footnotemark[2] & $\boldsymbol{p}_{\text{DRN}}$\footnotemark[3]   \\
Min.\footnotemark[1]         & -                    & -                    & - & $[1, 3, 5]$  \\
KL Penalty             & -                    & -                    & -                    & $Real(10^{-5}, 0.1)$     \\
Roughness Penalty            & -                    & -                    & -                    & $\boldsymbol{\alpha}_{\text{Roughness}}$\footnotemark[4]    \\
Mean Penalty           & -                    & -                    & -                    & $\boldsymbol{\alpha}_{\text{Mean}}$\footnotemark[5]      \\
\bottomrule
\bottomrule
\end{tabular}
\begin{flushleft}
    \footnotemark[1]{
        The minimum number of training observations required in each partitioned interval.
    }
\\
        \footnotemark[2]{
       $\boldsymbol{p}_{\text{DDR}}=[0.01, 0.125, 0.015, 0.175, 0.02, 0.0225, 0.025, 0.0275, 0.03]$.
    }
\\
        \footnotemark[3]{
        $\boldsymbol{p}_{\text{DRN}}=[0.02, 0.0225, 0.025, 0.0275, 0.03]$.
    }
\\
        \footnotemark[4]{
        $\boldsymbol{\alpha}_{\text{Roughness}}=[10^{-3}, 10^{-2}, 10^{-1}]$.
    }
\\
        \footnotemark[5]{
        $\boldsymbol{\alpha}_{\text{Mean}}=[10^{-5}, 10^{-4}, 10^{-3}, 10^{-2}, 10^{-1}]$.
    }
\end{flushleft}
\end{table}

\begin{table}[H]
    \centering
    \caption{(Synthetic Dataset) Hyperparameters for the competing models.
    The distributional assumption is gamma for CANN and MDN.}
    \label{tab: (Synthetic) Hyperparameters}
    \scalebox{0.95}{
        \begin{tabular}{l|cccc}
        \toprule
        \toprule
        \text{Hyperparameters} & \text{CANN} & \text{MDN} & \text{DDR} & \text{DRN} \\
        \midrule
        Learning Rate  
        & 0.00638
        & 0.00451
        & 0.00642
        & 0.00081 \\
        Batch Size 
        & 256 
        & 128 
        & 256 
        & 256 \\
        Dropout Rate  
        & 0.100 & 0.5 & 0.0192 & 0.140 \\
        Number of Layers 
        & 3 & 1 & 3 & 3 \\
        Neurons per Layer 
        & 512 & 256 & 32 & 128 \\
        Mixture Components
        & - & 10 & - & - \\
        Cutpoints Ratio (Min.\footnotemark[1]) 
        & - & - & 0.03 (0) & 0.025 (5) \\
        KL Penalty
        & - & - & - & 0.00047\\
        Roughness Penalty
        & - & - & - & 0.1 \\
        Mean Penalty
        & - & - & - & 0.01\\
        Activation Function 
        & \texttt{LeakyReLU} & \texttt{LeakyReLU} & \texttt{LeakyReLU} & \texttt{LeakyReLU}  \\
        Output Activation 
        & \texttt{exponential} & \texttt{Softmax}, \texttt{Softplus} & \texttt{Softmax} & \texttt{Linear}  \\
        \bottomrule
        \bottomrule
        \end{tabular}
        \vspace{0.2em}
    }
\begin{flushleft}
    \footnotemark[1]{
        The minimum number of training observations required in each partitioned interval.
    }
\end{flushleft}
\end{table}

\begin{table}[H]
\centering
\caption{(\texttt{freMPL1} Dataset) Hyperparameter Ranges Across Models. $Real(a, b)$ means the hyperparameter space is continuous from $a$ to $b$ (inclusive). $Int(a, b)$ means the hyperparameter space consists of discrete integer values from $a$ to $b$ (inclusive).}
\label{tab:real hyperparameters ranges}
\begin{tabular}{lcccc}
\toprule
\toprule
Hyperparameter       & CANN                 & MDN                  & DDR                  & DRN                  \\ 
\midrule
Learning Rate       & $Real(0.0001, 0.01)$   &  $Real(0.0001, 0.01)$ &  $Real(0.0002, 0.01)$   &  $Real(0.0002, 0.1)$ \\
Batch Size           &  $[64, 128, 256, 512]$ &$ [64, 128, 256, 512]$     & $[64, 128, 256, 512]$  & $[64, 128, 256, 512]$  \\
Dropout Rate         & $Real(0.0, 0.5)$      & $Real(0.0, 0.5)$      & $Real(0.0, 0.5)$      & $Real(0.0, 0.5)$      \\
Number of Layers     & $Int(1, 6)$           & $Int(1, 6)$          & $Int(1, 6)$       & $Int(1, 6)$           \\
Neurons per Layer    & $[32, 64, 128, 256, 512]$ 
& $[32, 64, 128, 256, 512]$  &$ [32, 64, 128, 256, 512]$ & $[32, 64, 128, 256, 512]$  \\
Mixture Components   & -                    & $Int(2, 10)$    & -                    & -                    \\
Proportion           & -                    & -                    & $\boldsymbol{p}_{\text{DDR}}$\footnotemark[2]         & $\boldsymbol{p}_{\text{DRN}}$  \footnotemark[3]          \\
Min.\footnotemark[1]         & -                    & -                    & - & [1, 3, 5]  \\
KL Penalty             & -                    & -                    & -                    & $Real(10^{-6}, 10^{-1}) $    \\
Roughness Penalty            & -                    & -                    & -                    & $\boldsymbol{\alpha}_{\text{Roughness}}$\footnotemark[4]    \\
Mean Penalty           & -                    & -                    & -                    & $\boldsymbol{\alpha}_{\text{Mean}}$\footnotemark[5]      \\
\bottomrule
\bottomrule
\end{tabular}
\begin{flushleft}
    \footnotemark[1]{
        The minimum number of training observations required in each partitioned interval.
    }
\\
        \footnotemark[2]{
       $\boldsymbol{p}_{\text{DDR}}=[0.05, 0.075, 0.1, 0.125, 0.15]$.
    }
\\
        \footnotemark[3]{
        $\boldsymbol{p}_{\text{DRN}}=[0.1, 0.125, 0.15]$.
    }
\\ 
        \footnotemark[4]{
        $\boldsymbol{\alpha}_{\text{Roughness}}=[0, 10^{-6}, 10^{-5}, 10^{-4}, 10^{-3}, 10^{-2}, 10^{-1}]$.
    }
\\
        \footnotemark[5]{
        $\boldsymbol{\alpha}_{\text{Mean}}=[0, 10^{-6}, 10^{-5}, 10^{-4}, 10^{-3}, 10^{-2}, 10^{-1}]$.
    }
\end{flushleft}
\end{table}

\begin{table}[H]
    \centering
    \caption{(\texttt{freMPL1} Dataset) Hyperparameters for the competing models.
    The distributional assumption is gamma for the GLM, CANN and MDN.}
    \label{tab: (Real) Hyperparameters}
    \scalebox{0.95}{
        \begin{tabular}{l|cccc}
        \toprule
        \toprule
        \text{Hyperparameters} & \text{CANN} & \text{MDN} & \text{DDR} & \text{DRN} \\
        \midrule
        Learning Rate  
        & 0.00890
        & 0.00845
        & 0.00578
        & 0.00291\\
        Batch Size 
        & 256 
        & 256 
        & 256 
        & 512 \\
        Dropout Rate  
        & 0.43674 & 0.43747 & 0.5 & 0.26987 \\
        Number of Layers 
        & 4 & 3 & 1 & 2 \\
        Neurons per Layer 
        & 512 & 256 & 512 & 512 \\
        Mixture Components
        & - & 4 & - & - \\
        Cutpoints Ratio (Min.\footnotemark[1]) 
        & - & - & 0.15 (0) & 0.125 (3) \\
        KL Penalty
        & - & - & - & 0.00162\\
        Roughness Penalty
        & - & - & - & $10^{-5}$ \\
        Mean Penalty
        & - & - & - & $10^{-6}$\\
        Activation Function 
        & \texttt{LeakyReLU} & \texttt{LeakyReLU} & \texttt{LeakyReLU} & \texttt{LeakyReLU}  \\
        Output Activation 
        & \texttt{exponential} & \texttt{Softmax}, \texttt{Softplus} & \texttt{Softmax} & \texttt{Linear}  \\
        \bottomrule
        \bottomrule
        \end{tabular}
        \vspace{0.2em}
    }
\begin{flushleft}
    \footnotemark[1]{
        The minimum number of training observations required in each partitioned interval.
    }
\end{flushleft}
\end{table}

\section{Evaluation Metrics}
\label{appendix: evaluation metrics}

Let $\mathcal{D}=\{(\boldsymbol{x}_i, y_i)\}_{i=1}^{N}$ be the dataset for evaluation and $\boldsymbol{w}$ represent the trained parameters/weights of the model.
The four evaluation metrics are continuous ranked probability score (CRPS), negative log-likelihood (NLL), root mean squared error (RMSE), and $\alpha$ quantile loss ($\alpha$ QL), respectively.

\begin{enumerate}
    \item Continuous ranked probability score (CRPS):
    \begin{align}
        \text{CRPS}(\mathcal{D}, \boldsymbol{w}) = \frac{1}{|\mathcal{D}|} \sum_{(\boldsymbol{x}_i, y_i)\in \mathcal{D}} \int_{y}\big(F_{Y|\boldsymbol{X}}(y|\boldsymbol{x}_i; \boldsymbol{w}) - \mathds{1}_{\{y>y_{i}\}}\big)^2 \mathrm{d}y.
    \end{align}
    A lower CRPS score indicates better model performance.
    \item Negative log-likelihood (NLL):
    \begin{align}
        \text{NLL}(\mathcal{D}, \boldsymbol{w}) = -\frac{1}{|\mathcal{D}|} \sum_{(\boldsymbol{x}_i, y_i)\in \mathcal{D}}  \ln f_{Y|\boldsymbol{X}}(y_i|\boldsymbol{x}_i; \boldsymbol{w}).
    \end{align}
   A lower NLL score indicates better model performance.
    \item Root mean squared error (RMSE):
    \begin{align}
        \text{RMSE}(\mathcal{D}, \boldsymbol{w}) = \sqrt{\frac{1}{|\mathcal{D}|}\sum_{(\boldsymbol{x}_i, y_i)\in \mathcal{D}}  \big(\mathbb{E}_{Y|\boldsymbol{X}}[Y|\boldsymbol{x}_i; \boldsymbol{w}]-y_i\big)^2}.
    \end{align}
    A lower RMSE score indicates a better model performance.
    \item $\alpha$ quantile loss ($\alpha$ QL):
    \begin{align}
       \text{QL}_{\alpha}(\mathcal{D}, \boldsymbol{w}) &= \frac{1}{|\mathcal{D}|} \sum_{(\boldsymbol{x}_i, y_i)\in \mathcal{D}}
              (y_i-Q_{Y|\boldsymbol{X}}(\alpha| \boldsymbol{x}_i;\boldsymbol{w}))
    \big(\alpha - \mathds{1}_{\{y_i\le Q_{Y|\boldsymbol{X}}(\alpha|\boldsymbol{x}_i;\boldsymbol{w})\}}\big).
    \end{align}
    A lower QL indicates a better model performance.
\end{enumerate}

\section{Kernel SHAP}
\label{appendix: Kernel SHAP}

The Shapley value of a feature $X_j$ represents the average marginal contribution of $X_j$ to the model's prediction across all possible coalitions of features.
Specifically, the Shapley value for the $j$th feature of instance $\boldsymbol{x}$ is given by
\begin{align} \label{SHAPvalue}
    \phi_j(v_{\mathcal{M}},\boldsymbol{x}) = \sum_{S \subseteq D \backslash \{j\}} \frac{|S|!(p - |S| - 1)!}{p!} (v_{\mathcal{M}}(S \cup \{j\}, \boldsymbol{x}) - v_{\mathcal{M}}(S, \boldsymbol{x})), \quad j = 1, \ldots, p.
\end{align}
where $p$ is the number of features and $S\subseteq D=\{1, \ldots, p\}$ represents a subset of the feature component indices and $v_{\mathcal{M}}(S, \boldsymbol{x})$ is a value function.

\citet{lundberg2017unified} proposed the Kernel SHAP that solves the following constrained optimisation problem
\begin{align} \label{TrueSHAPOptimisation}
     &\boldsymbol{\phi}(v_{\mathcal{M}},\boldsymbol{x}) = \underset{\boldsymbol{\beta}}{\text{arg min}} \sum_{S\subseteq D} (v_{\mathcal{M}}(S, \boldsymbol{x}) -\langle \boldsymbol{z}'(S), \boldsymbol{\beta} \rangle)^2 \cdot  \frac{p-1}{\binom{p}{|S|}}|S|\cdot(p - |S|)\\
    &  \text{s.t.
} \beta_0 = v_{\mathcal{M}}(\emptyset, \mathcal{D})\equiv\mathbb{E}_{\boldsymbol{X}}[\mathcal{M}(\boldsymbol{X})] \text{, }  \langle \boldsymbol{1}, \boldsymbol{\beta} \rangle = v_{\mathcal{M}}(D, \boldsymbol{x})\equiv \mathcal{M}(\boldsymbol{x}),
\end{align}
where $\boldsymbol{\phi}(v_{\mathcal{M}}, \boldsymbol{x})= (\phi_0(v_{\mathcal{M}}, \mathcal{D}), \phi_1(v_{\mathcal{M}}, \boldsymbol{x}), \ldots, \phi_{p}(v_{\mathcal{M}}, \boldsymbol{x}))^{\top}\in\mathbb{R}^{p+1}$, $\boldsymbol{\beta}=(\beta_0, \beta_1, \ldots, \beta_p)^{\top}\in \mathbb{R}^{p+1}$ and $\boldsymbol{z}'(S)= (1, \mathds{1}_{\{1\in S\}}, \ldots, \mathds{1}_{\{p\in S\}})^{\top} \in \{0, 1\}^{p+1}$.
Intuitively, the Kernel SHAP method finds $\boldsymbol{\phi}(v_{\mathcal{M}}, \boldsymbol{x})$ to explain the prediction for the instance $\boldsymbol{x}$.
To compute the value function $v_{\mathcal{M}}(S, \boldsymbol{x})$, Kernel SHAP leverages the conditional Shapley value that is the expected value of the regression output from the model $\mathcal{M}$ conditional on $\boldsymbol{X}_{S}=\boldsymbol{x}_{S}$.
Mathematically,
\begin{align} \label{TrueSHAPValueFunction}
    v_{\mathcal{M}}(S, \boldsymbol{x}) &= \mathbb{E}_{\boldsymbol{X}_{\bar{S}}|\boldsymbol{X}_{S}}[\mathcal{M}(\boldsymbol{X})|\boldsymbol{X}_{S} = \boldsymbol{x}_{S}]\\
    & = \mathbb{E}_{\boldsymbol{X}_{\bar{S}}|\boldsymbol{X}_{S}}[\mathcal{M}(\boldsymbol{X}_{S}, \boldsymbol{X}_{\bar{S}})|\boldsymbol{X}_{S} = \boldsymbol{x}_{S}] \nonumber \\
    &= \int_{\boldsymbol{x}_{\bar{S}}} \mathcal{M}(\boldsymbol{x}_{S}, \boldsymbol{x}_{\bar{S}}) \cdot f_{\boldsymbol{X}_{\bar{S}}|\boldsymbol{X}_{S}}(\boldsymbol{x}_{\bar{S}}|\boldsymbol{x}_{S}) \ \mathrm{d}\boldsymbol{x}_{\bar{S}}
    \approx \frac{1}{M}\sum_{m=1}^{M} \mathcal{M}\Big(\boldsymbol{x}_{S}, \boldsymbol{x}_{\bar{S}}^{(m)}\Big), \label{KernelSHAPValueApproximation}
\end{align}
where $\bar{S}$ denotes the complement of $S$ and $\boldsymbol{X}_{S}$ represents the components of features $\boldsymbol{X}$ in the feature subset $S$.
The Kernel SHAP employs the Monte Carlo approximation to compute the value function since the conditional density of the missing features $\boldsymbol{X}_{\bar{S}}=\boldsymbol{x}_{\bar{S}}$ given the observed features $\boldsymbol{x}_{S}$ is rarely known.
Therefore, it uses the marginal distribution of $\boldsymbol{X}_{\bar{S}}$ obtained from a subset of the training samples, where $\boldsymbol{x}_{\bar{S}}^{(m)}$ for $m \in \{1, \ldots, M\}$ are samples from the training data independent of $\boldsymbol{x}_{S}$.

Kernel SHAP also facilitates a smooth transition from local to global interpretability.
After calculating SHAP values for each observation, $\{(\boldsymbol{x}_{i}, \boldsymbol{\phi}(v_{\mathcal{M}}, \boldsymbol{x}_i))\}_{i=1}^{n}$, the SHAP feature importance is computed by
\begin{align}
   I_j =\frac{1}{n} \sum_{i=1}^{n}|\phi_{j}(v_{\mathcal{M}}, \boldsymbol{x}_i)|
\end{align}
for all $j\in\{1, \ldots, p\}$.
Theoretically, a larger $I_j$ value indicates greater importance for feature $X_j$.
Two types of visualisations can be generated: the SHAP beeswarm plot and the SHAP dependence plot.
The beeswarm plot presents the distribution of SHAP values for all features across various instances.
This visualisation aids in identifying key features and their impact on model predictions.
The SHAP dependence plot demonstrates the variation of SHAP values for a particular feature across all data points.
The overall shape of the plot allows for the inference of a feature's value on the model's predictions.
Further, it permits investigation of interaction effects between features, with points $\{(x_{i,j}, \phi_{j}(v_{\mathcal{M}}, \boldsymbol{x}_i))\}_{i=1}^{n}$ colour-coded with the values of another feature $X_k, k\neq j$.
Illustrative examples of these plots are provided in Sections~\ref{SyntheticModelIntepretability} and \ref{Real Dataset Global Interpretability}.

\section{Definition of Calibration}
\label{appendix: calibration}

\begin{definition}[Probabilistically Calibrated {\normalfont - \cite{gneiting2007probabilistic}}] \label{def: probcalibration}
Let $(F_t)_{t=1,2,\ldots}$ and $(G_t)_{t=1,2,\ldots}$ denote sequences of continuous and strictly increasing CDFs, possibly depending on stochastic parameters.
We think of $(G_t)_{t=1,2,\ldots}$  as the true data generating process and of $(F_t)_{t=1,2,\ldots}$ as the associated sequence of probabilistic forecasts.
The sequence $(F_t)_{t=1,2,\ldots}$ is probabilistically calibrated relative to the sequence $(G_t)_{t=1,2,\ldots}$ if
\begin{align} \label{Def1}
    \frac{1}{T} \sum_{t=1}^{T} G_t \circ F_{t}^{-1}(p) \to p \quad \text{for all} \quad p\in(0, 1),
\end{align}
as $T\to \infty$.
\end{definition}

\end{document}

%% file: images/tikz/DRN_Overview.tex
\begin{tikzpicture}
    \coordinate (A) at (-5.6,-8.5-1.5);
    \coordinate (B) at (-5.6,-13.5-1.5);
    \coordinate (C) at (-0.1,-13.5-1.5);
    \coordinate (D) at (-0.1,-8.5-1.5);
    \draw [gray, dashed] (A) -- (B) -- (C) -- (D) -- cycle;
    
    \coordinate (A) at (0.,-8.5-1.5);
    \coordinate (B) at (0.,-16.5-1.5);
    \coordinate (C) at (8.1,-16.5-1.5);
    \coordinate (D) at (8.1,-8.5-1.5);
     \node at (4.8,-16.8-1.5)
        {\textcolor{gray}{\Large \textcolor{blue!70}{Distributional Refinement Network}}};

    \draw [blue, dashed] (A) -- (B) -- (C) -- (D) -- cycle;

    \node(Baseline-1) at (-2.7,-13.5) {};
       \node at (-2.7,-17) {\Large Features};
        \node (Features-1) at (-2.7,-16.) {
        \begin{tabular}{|c|c|c|c|}
            \hline
            \textbf{Policy} & \textbf{Age} & \textbf{Gender} & \ldots \\
            \hline
            A & 29 & Male  & \ldots \\
            \hline
        \end{tabular}
    };
    \draw[->, shorten >=1pt, line width=9pt, color = gray!25] (Features-1) -- (Baseline-1);     
    
        \node (Baseline-0) at (-2.9,-13.0-1.5)
        {\textcolor{gray}{\Large Parametric Baseline Model}};

    \node (Baseline-2) at (0.,-10.35-1.5) {};
    \node (DRN-0) at (2.0,-10.35-1.5) {}; 
    \draw[->, shorten >=1pt, line width=9pt, color = teal!30] (Baseline-2) -- (DRN-0);
        \node at (1.,-11.2-1.5){\textcolor{teal!65}{\Large Input}};

    \node (Features-2) at (0.,-15.35-0.5) {};
    \node (DRN-1) at (2.0,-15.35-0.5) {}; 
    \draw[->, shorten >=1pt, line width=9pt, color = teal!30] (Features-2) -- (DRN-1);
        \node at (1.,-16.2-0.5){\textcolor{teal!65}{\Large Input}};

        \node at (-2.8,-9.5-1.5){\Large \textcolor{gray!70}{Baseline Density of Claim Size}};
    
    \begin{axis}[
        at={(-5.3cm,-13.7cm)}, 
        anchor=south west, 
        xmin=0, xmax=6.5,
        ymin=0, ymax=1,
        axis lines=middle,
        axis y line=none, 
        xlabel=, 
        ylabel=\Large \textcolor{gray!70}{Density of Claim Size},
        ticks=none,
        width=6.6cm,
        height=6cm 
    ]

    \addplot+[domain=0:6.5, samples=10, smooth, line width=1.75pt, color=gray!70, no marks] {gammadist(1.8,1)};
    \end{axis}

    \foreach \i in {2,...,\hiddennumone}
    {   \node[teal!30,
        font=\Huge,
        yshift=(\hiddennumtwo-\inputnum)*16.6 mm
        ]
        (Dotted1-\i) at (2.6,-8.7-1.5-1.5*\i) {$\dots$};
    }
    \foreach \i in {2, 4, 5}
    {
        \node[circle, 
            minimum size = 7.5mm,
            fill=teal!30,
            yshift=(\hiddennumone-\inputnum)*7.2 mm
        ] (Hidden1-\i) at (3.9777,-7-1.5-2.0*\i) {};
    }
    \path (Hidden1-2) -- (Hidden1-4) node [teal!30, font=\Huge, midway, sloped,
                                          xshift=(\hiddennumtwo-\inputnum)*-0.5 mm
                                          ]{$\dots$};

    \foreach \i in {2, 4, 5}
    {
        \node[circle, 
            minimum size = 7.5mm,
            fill=teal!30,
            yshift=(\hiddennumone-\inputnum)*7.2 mm
        ] (Hidden2-\i) at (5.5,-7-1.5-2.0*\i) {};
    }
    \path (Hidden2-2) -- (Hidden2-4) node [teal!30, font=\Huge, midway, sloped,
                                          xshift=(\hiddennumtwo-\inputnum)*-0.5 mm
                                          ]{$\dots$};
    
    \foreach \i in {2,...,\hiddennumone}
    {   \node[teal!30,
        font=\Huge,
        yshift=(\hiddennumtwo-\inputnum)*16.6 mm
        ]
        (Dotted2-\i) at (7.2,-8.7-1.5-1.5*\i) {$\dots$};
    }

    \foreach \i in {2,3,4,5}
    {
        \foreach \j in {2,4,5}
        {
            \draw[->, shorten >=1pt, color = blue!50] (Dotted1-\i) -- (Hidden1-\j);
        }
    }
    \foreach \i in {2,4,5}
    {
        \foreach \j in {2,4,5}
        {
            \draw[->, shorten >=1pt, color = blue!50] (Hidden1-\i) -- (Hidden2-\j);
        }
    }
    \foreach \i in {2,4,5}
    {
        \foreach \j in {2,3,4,5}
        {
            \draw[->, shorten >=1pt, color = blue!50] (Hidden2-\i) -- (Dotted2-\j);
        }
    }

    \node (DRN-2) at (7.6,-12.3-1.5) {};
    \node[circle, 
            minimum size = 1mm,
            yshift= 0 mm,
            fill=white!50] (Output-1) at (9.4,-12.3-1.5) {}; 
    \draw[->, shorten >=1pt, line width=9pt, color = purple!50] (DRN-2) -- (Output-1);
        \node at (8.3,-13.2-1.5)
        {\textcolor{purple!70}{\Large Output}};

    \node at (10.5,-11.5-1.5){\Large \textcolor{blue!45}{Refined Density of Claim Size}};
    \begin{axis}[
        at={(8.6cm,-15cm)}, 
        anchor=south west, 
        xmin=0, xmax=6.5,
        ymin=0, ymax=1,
        axis lines=middle,
        axis y line=none, 
        xlabel=,
        ylabel=\Large \textcolor{blue!45}{Density of Claim Size},
        ticks=none,
        width=6.6cm, 
        height=6cm 
    ]

    \addplot+[domain=0:6.5, samples=10, smooth, line width=1.75pt, color=blue!50, no marks] {gammadist(3.0,1.0)};
    \end{axis}

    \end{tikzpicture}

%% file: images/tikz/P1_Partition_Method_Default.tex
\begin{tikzpicture}
\pgfplotsset{
width= 290,
height= 240,
}
\begin{axis}[%
    axis lines = center,
    xlabel = {$y$},
    ylabel = {$\hat{f}_{Y|\boldsymbol{X}}$},
    label style = {anchor=north east},
    xmin = 0, xmax=10,     ymin=0, ymax=0.3,
    ticks = none,
    enlargelimits=false,
    clip=false,
    domain = 0 : 10,
    samples = 50,
    no marks
            ]
\addplot [domain=0:1.2,thick] {0.12};
\addplot [domain=1.2:2.4,thick] {0.24};
\addplot [domain=2.4:3.6,thick] {0.19};
\addplot [domain=3.6:4.8,thick] {0.17};
\addplot [domain=4.8:6.0,thick] {0.12};
\addplot [domain=6.0:7.2,thick] {0.06};
    \node[circle, inner sep=1pt, label={268:{\footnotesize \textcolor{blue}{$ $}}}] (b0) at (0, 120) {};
    \node[circle, inner sep=1pt, label={268:{\footnotesize \textcolor{blue}{$ $}}}] (b1) at (12, 120) {};
     \node[circle, inner sep=1pt, label={268:{\footnotesize \textcolor{blue}{$ $}}}] (c0) at (12, 240) {};
    \node[circle, inner sep=1pt, label={268:{\footnotesize \textcolor{blue}{$ $}}}] (c1) at (24, 240) {};
    \node[circle, inner sep=1pt, label={268:{\footnotesize \textcolor{blue}{$ $}}}] (d0) at (24, 190) {};
    \node[circle, inner sep=1pt, label={268:{\footnotesize \textcolor{blue}{$ $}}}] (d1) at (36, 190) {};
    \node[circle, inner sep=1pt, label={268:{\footnotesize \textcolor{blue}{$ $}}}] (e0) at (36, 170) {};
    \node[circle, inner sep=1pt, label={268:{\footnotesize \textcolor{blue}{$ $}}}] (e1) at (48, 170) {};
    \node[circle, inner sep=1pt, label={268:{\footnotesize \textcolor{blue}{$ $}}}] (f0) at (48, 120) {};
     \node[circle, inner sep=1pt, label={268:{\footnotesize \textcolor{blue}{$ $}}}] (f1) at (60, 120) {};
    \node[circle, inner sep=1pt, label={268:{\footnotesize \textcolor{blue}{$ $}}}] (g0) at (60, 60) {};
      \node[circle, inner sep=1pt, label={268:{\footnotesize \textcolor{blue}{$ $}}}] (g1) at (72, 60) {};
        \node[circle, inner sep=1pt, label={268:{\footnotesize \textcolor{blue}{$ $}}}] (h0) at (72, 60) {};
        \node[circle, inner sep=1pt, label={268:{\footnotesize \textcolor{blue}{$ $}}}] (h1) at (72, 60) {};

    \node[circle, fill, inner sep=1pt, label={268:{\footnotesize \textcolor{blue}{$c_0=l$}}}] (a) at (0, 0) {};
     \node[circle, fill, inner sep=1pt, label={270:{\footnotesize\textcolor{blue}{$c_1$}}}] (b) at (12, 0) {};
    \node[circle, fill, inner sep=1pt, label={268:{\footnotesize\textcolor{blue}{$c_2$}}}] (c) at (24,0) {};
    \node[circle, fill, inner sep=1pt, label={270:{\footnotesize\textcolor{blue}{$ $}}}] (d) at (36,0) {};
    \node[circle, fill, inner sep=1pt, label={270:{\footnotesize\textcolor{blue}{$ $}}}] (e) at (48,0) {};
    \node[circle, fill, inner sep=1pt, label={270:{\footnotesize\textcolor{blue}{$ $}}}] (f) at (60,0) {};
    \node[circle, fill, inner sep=1pt, label={270:{\footnotesize\textcolor{blue}{$c_{K-1}$}}}] (gnew) at (60, 0) {};
    \node[circle, fill, inner sep=1pt, label={270:{\footnotesize\textcolor{blue}{$ $}}}] (g) at (72, 0) {};
    \node[circle, fill, inner sep=1pt, label={272:{\footnotesize\textcolor{blue}{$c_{K}=u$}}}] (h) at (72,0) {};
    \node[black][circle, fill, inner sep=0.0000001pt, label={270:{\textcolor{blue}{$...$}}}] (mid) at (48,0) {};

    \draw[densely dashed] (c1) -- (c);
    \draw[densely dashed] (d1) -- (d);
    \draw[densely dashed] (e1) -- (e);
    \draw[densely dashed] (f1) -- (f);
    \draw[densely dashed] (h1) -- (h);

    \draw[densely dashed] (c0) -- (b);
    \draw[densely dashed] (d0) -- (c);
    \draw[densely dashed] (g0) -- (f);

    \draw [pen colour={red!95}, decorate,
    decoration = {calligraphic brace}] (b1) --  (b) node[pos=.5, right] {\footnotesize\textcolor{red!95}{${\pi}_1(\boldsymbol{X}; \boldsymbol{w})/|T_1|$}};
    
    \draw [pen colour={red!95}, decorate,
    decoration = {calligraphic brace}] (c1) --  (c) node[pos=.45, right] {\footnotesize\textcolor{red!95}{${\pi}_2(\boldsymbol{X}; \boldsymbol{w})/|T_2|$}};

    \draw [pen colour={red!95}, decorate,
    decoration = {calligraphic brace}] (h1) --  (h) node[pos=.45, right] {\footnotesize\textcolor{red!95}{${\pi}_{K}(\boldsymbol{X}; \boldsymbol{w})/|T_K|$}};

    \draw [pen colour={blue!50}, decorate,
    decoration = {calligraphic brace}] (a) --  (b) node[pos=.45, above] {\footnotesize\textcolor{blue!50}{$T_1$}};
    \draw [pen colour={blue!50}, decorate,
    decoration = {calligraphic brace}] (b) --  (c) node[pos=.45, above] {\footnotesize\textcolor{blue!50}{$T_2$}};
    \draw [pen colour={blue!50}, decorate,
    decoration = {calligraphic brace}] (gnew) --  (h) node[pos=.45, above] {\footnotesize\textcolor{blue!50}{$T_{K}$}};

\end{axis}
\end{tikzpicture}

%% file: images/tikz/P1_DRN_New.tex
\begin{tikzpicture}

    \foreach \i in {1, 3}
    {
        \node[circle, 
            minimum size = 12mm,
            yshift=-7 mm,
            fill=black!30] (Input-\i) at 
            (0,-1.5*\i) {};
    }
      
    \path (Input-1) -- (Input-3) node [black!30, font=\Huge, midway, sloped,
                                          xshift=(\hiddennumtwo-\inputnum)*-0.5 mm
                                          ]{$\dots$};
    \foreach \i in {5, 7}
    {
        \node[circle, 
            minimum size = 12mm,
            yshift=-7 mm,
            fill=black!30] (Input-\i) at (0,-1.5*\i) {};
    }

    \foreach \i in {6}
    {
        \node[circle, 
            minimum size = 1mm,
            yshift=-7 mm,
            fill=white!30] (Input-\i) at (0,-1.5*\i) {};
    }

    \path (Input-5) -- (Input-7) node [black!30, font=\Huge, midway, sloped,
                                          xshift=(\hiddennumtwo-\inputnum)*-0.5 mm
                                          ]{$\dots$};
    
    \foreach \i in {2}
    {
        \node[circle, 
            minimum size = 12mm,
            fill=teal!30,
            yshift=(\hiddennumone-\inputnum)*7.2 mm
        ] (Hidden1-\i) at (2,-1.5*\i) {};
    }
    \foreach \i in {4,5}
    {   \node[circle, 
            minimum size = 12mm,
            fill=teal!30,
            yshift=(\hiddennumtwo-\inputnum)*10.6 mm
        ] (Hidden1-\i) at (2,-1.5*\i) {};
    }

    \path (Hidden1-2) -- (Hidden1-4) node [teal!30, font=\Huge, midway, sloped,
                                          xshift=(\hiddennumtwo-\inputnum)*-0.5 mm
                                          ]{$\dots$};
    \foreach \i in {2}
    {   \node[teal!30,
        font=\Huge,
        yshift=(\hiddennumtwo-\inputnum)*16.6 mm
        ]
        (Dotted-\i) at (4,-1.6*\i) {$\dots$};
    }
    \foreach \i in {3,...,\hiddennumone}
    {   \node[teal!30,
        font=\Huge,
        yshift=(\hiddennumtwo-\inputnum)*16.6 mm
        ]
        (Dotted-\i) at (4,-1.63*\i) {$\dots$};
    }
    
    \foreach \i in {2}
    {
        \node[circle, 
            minimum size = 12mm,
            fill=teal!30,
            yshift=(\hiddennumone-\inputnum)*7.2 mm
        ] (Hidden2-\i) at (6,-1.5*\i) {};
    }
    \foreach \i in {4,5}
    {   \node[circle, 
            minimum size = 12mm,
            fill=teal!30,
            yshift=(\hiddennumtwo-\inputnum)*10.6 mm
        ] (Hidden2-\i) at (6,-1.5*\i) {};
    }

    \path (Hidden2-2) -- (Hidden2-4) node [teal!30, font=\Huge, midway, sloped,
                                          xshift=(\hiddennumtwo-\inputnum)*-0.5 mm
                                          ]{$\dots$};
    

    \foreach \i in {1, 2}
    {
        \node[circle, 
            minimum size = 12mm,
            fill=olive!50,
            yshift=-7mm
        ] (Output-\i) at (8.5,0.7-1.8*\i) {};
    }
    
    \foreach \i in {3}
    {
        \node[circle, 
            minimum size = 12mm,
            fill=olive!50,
            yshift=-7mm
        ] (Output-\i) at (8.5,-1.8*\i) {};
    }

    \path (Output-2) -- (Output-3) node [olive!20, font=\Large, midway, sloped,
                                          xshift=(\hiddennumtwo-\inputnum)*-0.5 mm
                                          ]{$\dots$};

    \foreach \i in {4, 5}
    {
        \node[circle, 
            minimum size = 12mm,
            fill=brown!50,
            yshift=-7mm
        ] (Output-\i) at (8.5,0.1-1.8*\i) {};
    }
    

    \foreach \i in {6}
    {
        \node[circle, 
            minimum size = 12mm,
            fill=brown!50,
            yshift=-7mm
        ] (Output-\i) at (8.5,-0.6-1.8*\i) {};
    }

    \path (Output-5) -- (Output-6) node [brown!30, font=\Huge, midway, sloped,
                                          xshift=(\hiddennumtwo-\inputnum)*-0.5 mm
                                          ]{$\dots$};
                      
        \foreach \i in {2,3}
    {
        \node[circle, 
            minimum size = 12mm,
            fill=purple!50,
            yshift=-7mm
        ] (Output2-\i) at (11,0.2-2.0*\i) {};
    }
        
        \foreach \i in {4}
    {
        \node[circle, 
            minimum size = 12mm,
            fill=purple!50,
            yshift=-7mm
        ] (Output2-\i) at (11,-0.5-2.0*\i) {};
    }
        \path (Output2-3) -- (Output2-4) node [purple!30, font=\Huge, midway, sloped,
                                          xshift=(\hiddennumtwo-\inputnum)*-0.5 mm
                                          ]{$\dots$};

    \foreach \i in {3, 1}
    {
        \foreach \j in {2,4,5}
        {
            \draw[->, shorten >=1pt, color = blue] (Input-\i) -- (Hidden1-\j);   
        }
    }
    
    \foreach \i in {2,4,5}
    {
        \foreach \j in {2,4,5}
        {
            \draw[->, shorten >=1pt, color = blue] (Hidden1-\i) -- (Dotted-\j);   
        }
    }
    
    \foreach \i in {2,4,5}
    {
        \foreach \j in {2,4,5}
        {
            \draw[->, shorten >=1pt, color = blue] (Dotted-\i) -- (Hidden2-\j);   
        }
    }
    \foreach \i in {2,4,5}
    {
        \foreach \j in {1, 2, 3}
        {
            \draw[->, shorten >=1pt, color = blue] (Hidden2-\i) -- (Output-\j);
        }
    }

    \foreach \i in {1,2,3,4,5,6}
    {
        \foreach \j in {2,3,4}
        {
            \draw[->, shorten >=1pt, color = purple] (Output-\i) -- (Output2-\j);
        }
    }


    \foreach \i in {1}
    {
        \node[circle, 
            minimum size = 3mm,
            yshift= 0 mm,
            fill=white] (Baseline-\i) at (3.5,-9.75) {};
    }
    

    \foreach \i in {2}
    {
        \node[circle, 
            minimum size = 3mm,
            yshift= 0 mm,
            fill=white] (Baseline-\i) at (6.5,-9.75) {};
    }



    \coordinate (A) at (-1.0,-7.5);
    \coordinate (B) at (-1.0,-12);
    \coordinate (C) at (7.8,-12);
    \coordinate (D) at (7.8,-7.5);
    
    \draw [black, dashed] (A) -- (B) -- (C) -- (D) -- cycle;

        \node (features-baseline) at (0.5, -9.6) {};
        \node (baseline-left) at (2.9, -9.6) {};
        \draw[->, shorten >=1pt, line width=7pt, color = gray!75] (features-baseline) -- (baseline-left);    
         \node at (1.7, -10.1) {\large Baseline Density};

        \node (baseline-right) at (5.8, -9.6) {};
        \node (bhats-left) at (8.1, -9.6) {};
        \draw[->, shorten >=1pt, line width=7pt, color = gray!75] (baseline-right) -- (bhats-left);    
         \node at (6.7, -10.1) {\large Transformation};

    \node at (3.4, -12.3) {\Large Section~\ref{Incorporating Baseline}};

    \begin{axis}[
        at={(3.4cm,-11.5cm)}, 
        anchor=south west, 
        xmin=0, xmax=4,
        ymin=0, ymax=1,
        axis lines=middle,
        xlabel=$Y$,
        ylabel=$f_{Y|\boldsymbol{X}}(y|\boldsymbol{x};\boldsymbol{\beta})$,
        ticks=none,
        width=5cm, 
        height=5cm 
    ]

    \addplot+[domain=0:3.5, samples=10, smooth, line width=0.5pt, color=black, no marks] {gammadist(1.8,1)};
    
    \end{axis}

    \coordinate (A) at (9.2,-3.5);
    \coordinate (B) at (9.2,-10.5);
    \coordinate (C) at (12,-10.5);
    \coordinate (D) at (12,-3.5);
    
    \draw[purple, dashed] (A) -- (B) -- (C) -- (D) -- cycle;
    
    \node at (10.6, -3.1) {\large \textcolor{purple!50}{Equation~\eqref{eq: adjustment factor}}};

    \node[] at (Input-1) {\large ${x}_1$};
    \node[] at (Input-3) {\large ${x}_{p}$};
    \node[] at (Input-5) {\large ${x}_1$};
    \node[] at (Input-7) {\large ${x}_{p}$};
    \node[] at (Output-1) { $\hat{l}_{1}(\boldsymbol{x})$};
    \node[] at (Output-2) { $\hat{l}_{2}(\boldsymbol{x})$};
    \node[] at (Output-3) { $\hat{l}_{K}(\boldsymbol{x})$};
    \node[] at (Output-4) { $\hat{b}_{1}(\boldsymbol{x})$};
    \node[] at (Output-5) { $\hat{b}_{2}(\boldsymbol{x})$};
    \node[] at (Output-6) { $\hat{b}_{K}(\boldsymbol{x})$};
    \node[] at (Output2-2) { $\hat{a}_{1}(\boldsymbol{x})$};
    \node[] at (Output2-3) { $\hat{a}_{2}(\boldsymbol{x})$};
    \node[] at (Output2-4) { $\hat{a}_{K}(\boldsymbol{x})$};
        
    \end{tikzpicture}